\documentclass{article}

\usepackage[preprint]{corl_2026}        % preprint (arXiv) (current default)

\usepackage[T1]{fontenc}                % T1 encoding so accented author names in canonical bib entries (e.g. Polish ogonek \k, l-bar) render under OT1-default styles
\usepackage{amsmath}                    % methodology equations + algorithm body
\usepackage{amssymb}                    % \mathfrak, \mathcal, \mathbb, etc.
\usepackage{xcolor}                     % red \textcolor TODO collaborator markers
\definecolor{silver}{RGB}{112,128,144}  % slate metallic
\usepackage{graphicx}                   % include generated PDF/PNG figures
\usepackage{booktabs}
\usepackage{placeins}                   % \toprule, \midrule, \bottomrule
\usepackage{makecell}                   % \makecell column wrapping in tables
\usepackage{cleveref}                   % \cref / \Cref
\usepackage{algorithm}                  % methodology Algorithm 1 (\textsc{Helix} loop)
\usepackage{algpseudocode}              % \State, \For, \If, \Require, \Return inside algorithm
\usepackage{fvextra}                    % RAI prompt appendix Verbatim breaklines
\usepackage{subcaption}                 % 2x1 side-by-side figure panels
\usepackage{enumitem}
\hypersetup{hypertexnames=false}        % unique anchors when floats/appendix reset counters
\hypersetup{
  pdftitle={RHO: Your Coding Agent is Secretly a Roboticist},
  pdfkeywords={RHO, Robotics Harness Optimization, robot manipulation, code as policies, reflective evolution, LIBERO, Robosuite}
}

\DeclareMathOperator*{\argmax}{arg\,max}

\title{RHO: Your Coding Agent is Secretly a Roboticist}

\author{
  Karim Elmaaroufi\\
  University of California, Berkeley\\
  Embodied Science
  \And
  Justin Svegliato\\
  University of California, Berkeley\\
  \And
  Sarunas Kalade\\
  AMD\\
  \AND
  Graham Schelle\\
  AMD\\
  \And
  Sanjit A. Seshia\\
  University of California, Berkeley\\
  \And
  Matei Zaharia\\
  University of California, Berkeley\\
}

\begin{document}
\maketitle

%% ========================================================================
%% Body sections (one file per section, flat layout at repo root)
%% ========================================================================

\providecommand{\PaperHostsRaiSection}{}
\begin{abstract}
    Code-as-Policies (CaP) has shown that large language models (LLMs) can write
    code to solve robotics tasks by composing perception, planning, and control
    primitives.
    Recent CaP systems, however, rely on multi-turn code-generation loops
    at test time, which is often infeasible for real-time robot control.
    We introduce \textbf{Robotics Harness Optimization (RHO)}, a novel paradigm in
    which tool-enabled coding agents, \textit{at training time}, propose and search
    for interpretable, neurosymbolic multi-file policy repositories
    (\textit{Repositories-as-Policies}) that compose
    these primitives rather than a single prompt, function, or file.
    RHO searches with reflective feedback from environment reward and execution
    rather than teleoperation demonstrations.
    It generalizes to perturbed pick-and-place settings like LIBERO-PRO, where
    OpenVLA scores 0.0\% and $\pi_{0.5}$ averages 12.83\%.
    Using the same low-level primitives, RHO reaches a 45.0\% success rate,
    2.5$\times$ higher than the strongest multi-turn agentic system,
    and 3.5$\times$ higher than $\pi_{0.5}$.
    On Robosuite, RHO sets a new state-of-the-art of 70.0\%,
    exceeding the prior multi-turn record of 68.29\%
    using single-turn execution with no corrective LLM code edits at deployment.
    When an LLM is used in the control loop, as on RAI's O3DE benchmark, RHO
    optimizes the deployed agent's multi-file harness of prompts, tools, and
    control code, improving held-out success from 23.5\% to 44.3\% with 20\% less
    wall-clock time and 27\% fewer tool calls.
\end{abstract}

% \keywords{Repositories-as-Policies, Code-as-Policies, Reflective Program Synthesis, Genetic / Evolutionary Optimization, Robot Manipulation, ROS\,2}

\section{Introduction}
\label{sec:introduction}

% P1. Anchor. Pattern 2 (Acknowledge-then-restrict): credit Code-as-Policies,
% then assert the thesis the abstract carries: the test-time code loop is
% unnecessary.
Code-as-Policies (CaP) has opened new possibilities for robot control by enabling large language models (LLMs) to write executable robot programs.
Rather than representing policies as black-box neural networks, CaP enables interpretability
with LLMs that write code to compose perception, planning, and action primitives.
These programs enable robots to perform tasks such as stacking blocks~\citep{liang2023code}, washing plates~\citep{singh2023progprompt}, and setting dining tables~\citep{huang2023voxposer}.
Recent CaP systems use test-time methods including multi-turn rounds, ensembles of candidate code completions, and visual-difference modules that convert execution feedback into prompt context~\citep{fu2026capx}.
While effective, these methods rely on iterative code generation loops at test time, often leaving them unsuitable for real-time robotics tasks.
We show that scaling LLM test-time inference at deployment is unnecessary.
% We show that this interaction can be moved into training-time repository optimization.

% P2. Tension. Numbers land here. Two failure modes: the agentic
% deployment cost on CaP-Bench, and the VLA collapse on LIBERO-PRO.
CaP agents and vision-language-action (VLA) models each fail differently, and
current evaluations document both~\citep{fu2026capx, zhou2025liberopro}. The best-performing
agent on CaP-Bench (CaP-Agent0) issues up to 66 LLM
calls at test time per trial. Yet scaling inference calls is not reliably beneficial~\citep{chen2024morecalls}.
% (nine candidate completions, a synthesis pass, and finally one visual-difference lookup per turn, all repeated up to six times) with an average of 1.999 turns.
Restricting the same agentic system to a single turn
without that scaffolding makes reported Robosuite success collapse from
\textbf{68.29\%} to \textbf{24\%}~\citep{fu2026capx}. Meanwhile,
VLAs struggle to generalize: models like OpenVLA and $\pi_{0.5}$~\citep{kim2024openvla,black2025pi05} that were trained on LIBERO demonstrations~\citep{liu2023libero} collapse to \textbf{0.0\%} and \textbf{12.83\%}, respectively~\citep{zhou2025liberopro}.
Other perturbation suites report related VLA generalization failures~\citep{fei2026liberoplus, wang2026liberox, kim2026liberopara, elmaaroufi2026liberoinfinity}.
% VLAs
% (OpenVLA, $\pi_{0.5}$)~\citep{kim2024openvla,black2025pi05} average just
% \textbf{0.0\%} and \textbf{12.83\%} respectively across the six perturbation cells
% of LIBERO-PRO~\citep{zhou2025liberopro} reported in CaP-Bench, a perturbation extension of LIBERO~\citep{liu2023libero}.

% P3. Thesis. Operating principle named: Repository-Level Mutation.
% Closes on a negative-space sentence (no gradient updates, no teleop,
% no LLM in the CaP-Bench control loop).
We introduce \textbf{Robotics Harness Optimization (RHO)}, a novel paradigm for
evolving interpretable neurosymbolic multi-file robot-policy repositories (\emph{Repositories-as-Policies}). To instantiate
RHO, we generalize GEPA (Genetic-Pareto), a reflective optimizer that mutates
artifacts through natural-language reflection while keeping a Pareto frontier of
candidates~\citep{agrawal2026gepa}, and Optimize Anything~\citep{agrawal2026optimizeanything}, into HELIX (Hierarchical Evolution
via LLM-Informed eXploration).
\emph{As our instantiation of RHO, HELIX is the first reflective optimizer that
  (1) lifts the optimization surface to whole multi-file repositories rather than
  a single prompt or function, and (2) uses a tool-enabled coding agent as a
  mutator that, within one mutation step, can write ephemeral tests, run candidate
  code in the robot environment, read tracebacks, query a visual-difference tool,
  and edit multiple files.}
We enable live interaction with the robot environment by designing a robotics-environment-as-a-service (REaaS),
and finally, thoughtful isolation mechanisms between the coding agent, REaaS, and the evaluator
to prevent cheating and reward hacking.
As a result, each generation retains useful repositories on a per-instance
Pareto/coverage frontier.
% The operating principle is \emph{Repository-Level Mutation}: the search
% object is the deployable controller, not a prompt or single function. 
The final artifact is a multi-file neurosymbolic repository that establishes several
state-of-the-art results and, unlike VLAs, is highly interpretable and debuggable.
% In CaP-Bench, the selected repository is static
% and does not contain gradient updates, teleoperation data, inference-time LLM
% code-generation calls, multi-turn correction, or
% deployment-time visual-difference feedback.
% As such, policy interpretation resolves to simply reading code.

\begin{figure}[!t]
  \centering
  \includegraphics[width=0.93\linewidth]{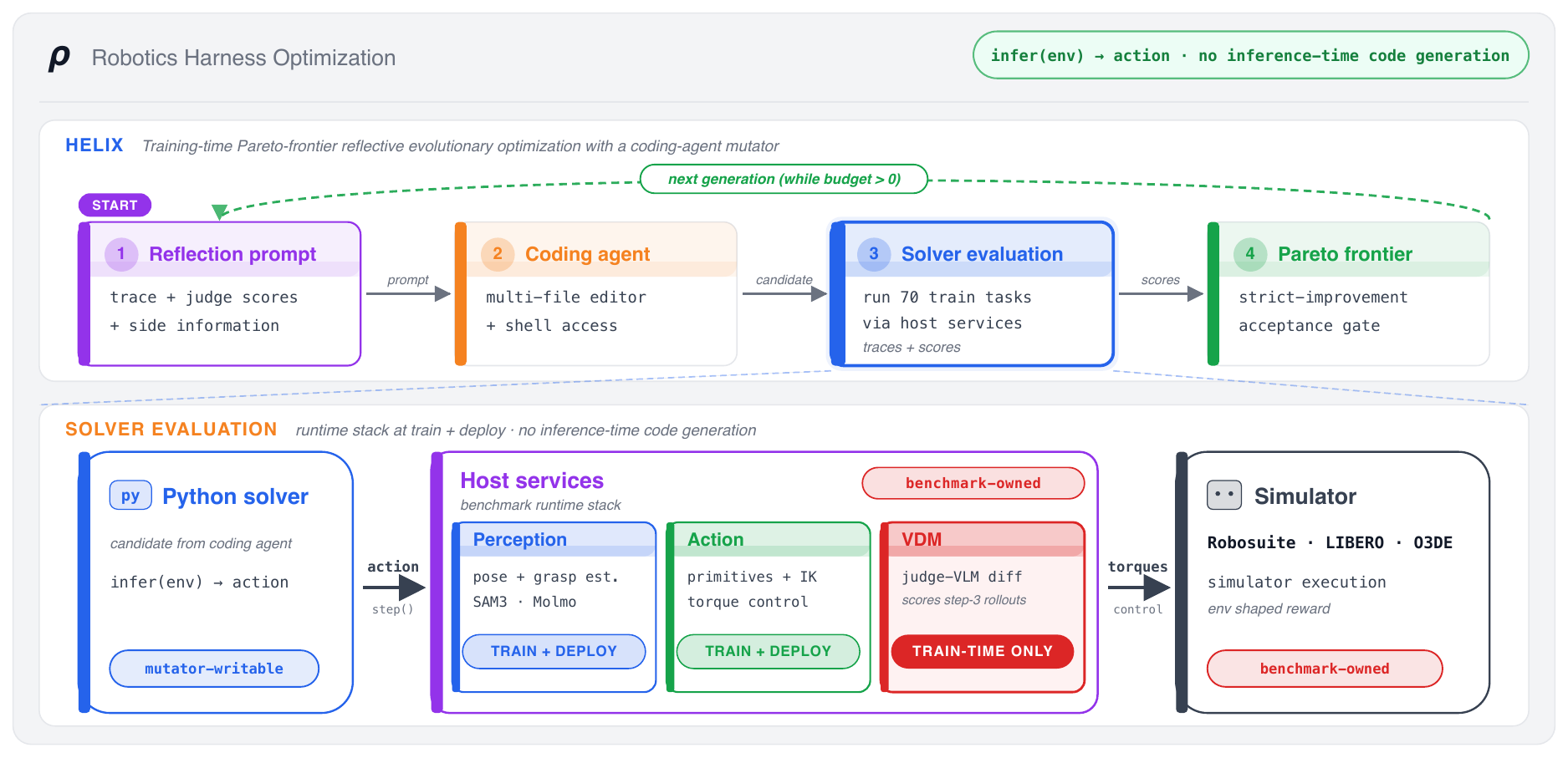}
  \caption{RHO system overview. A parent repository sampled from a
    Pareto frontier is mutated by a tool-enabled coding
    agent that edits multiple files and executes candidate policies in an
    isolated workspace with whitelisted access to a
    robot environment. If the child passes the configured training
    gate, it is evaluated on all
    validation examples and added to the Pareto frontier.
    At the end of evolution, the
    frontier repository with the highest mean validation reward
    is deployed as the policy.}

  \label{fig:helix-system}
\end{figure}

% P4. Headline results. Order: Robosuite SOTA -> RAI -> LIBERO-PRO 2.5x.
% Closes on the LIBERO-PRO 2.5x margin per the user's locked requirement.
On the CaP-Bench Robosuite~\citep{zhu2020robosuite} manipulation benchmark, RHO reaches
\textbf{70.0\%} (490 of 700 trials) under single-turn evaluations, exceeding
the strongest published record (68.29\%),
which required multi-turn code generation and visual-difference feedback at test time~\citep{fu2026capx}.
RHO also exceeds prior results when LLMs remain part of the control loop:
on Robotec's RAI ROS\,2 manipulation benchmark (O3DE)~\citep{rai}, RHO evolves
an entire multi-file repository of prompts, tools, and control code, nearly doubling held-out success from \textbf{23.5\%} to \textbf{44.3\%} with \textbf{20\%} less wall-clock
time and \textbf{27\%} fewer tool calls.
On the six LIBERO-PRO~\citep{zhou2025liberopro} perturbation cells reported in CaP-Bench,
where VLAs fail to generalize (OpenVLA: 0.0\% and $\pi_{0.5}$: 12.83\%), RHO trains
by interacting in the LIBERO environment (absent of demonstrations) and, within 63 generations, records a new state-of-the-art for CaP methods of \textbf{45.0\%} (1351 of 3000 trials)
under single-turn evaluations, a 3.5x improvement over $\pi_{0.5}$ and a 2.5x improvement over the strongest prior method
which required multi-turn corrective actions (CaP-Agent0 at 18.17\%).
In summary, our contributions are:

% P5. Contributions. 3 noun-phrase items, parallel structure, all
% checkable against a specific section in the body.
\begin{enumerate}
  \item \textbf{Robotics Harness Optimization.} RHO, a novel paradigm that
        shifts the computational burden from test-time code generation to \emph{training time}
        via a new reflective evolutionary optimization framework for robotics, whose
        optimized artifact is a multi-file policy repository rather than a
        single prompt or function, yielding interpretable neurosymbolic policies.
        % \item \textbf{Repository-Level Mutation.} The first reflective evolutionary
        %       optimization method that operates at the repository level rather than
        %       individual text artifacts.
  \item \textbf{Tool-enabled reflective mutation.} The first
        evolutionary reflective
        optimizer whose mutator is a tool-enabled coding agent
        that can execute tests, run code against the REaaS, read
        tracebacks, query visual-difference feedback, and edit
        multiple files within a generation.
  \item \textbf{Robust policies that beat agentic stacks and VLAs.}
        In single-turn evaluations, RHO reaches \textbf{70.0\%} on CaP-Bench Robosuite
        (versus the multi-turn record of \textbf{68.29\%}) with zero LLM code-generation calls in the control loop, nearly doubles held-out RAI O3DE success from
        \textbf{23.5\%} to \textbf{44.3\%} with \textbf{20\%} less
        wall-clock time and \textbf{27\%} fewer tool calls, and
        reaches \textbf{45.0\%} on LIBERO-PRO (versus prior CaP SOTA of
        \textbf{18.17\%}) where OpenVLA and $\pi_{0.5}$ average 0.0\%
        and 12.83\%, respectively.
\end{enumerate}

\section{Related Work}
\label{sec:related}

\paragraph{Learned visuomotor policies.}
End-to-end VLAs map pixels and language to low-level actions within a single
trained network rather than a separate
planner~\citep{brohan2022rt1,brohan2023rt2,kim2024openvla,octo2024,black2024pi0,black2025pi05,liu2024rdt1b,qu2025spatialvla,lee2025molmoact,bjorck2025groot,kim2025openvlaoft};
world-action variants add observation prediction or visual-goal
generation~\citep{zhen2024_3dvla,cen2025worldvla,chen2026goalvla}, and
model-based alternatives train a latent world model for explicit
optimization or imagined
rollouts~\citep{hansen2024tdmpc2,hafner2023dreamerv3,wu2022daydreamer,zhou2024robodreamer}.
Robustness across these paradigms remains bounded by the training
distribution: independent audits report tens-of-points collapses under
viewpoint, paraphrase, and layout
perturbations~\citep{fei2026liberoplus,zhou2025liberopro,kim2026liberopara},
and observation-predictive models produce visually plausible but
physically inconsistent rollouts~\citep{mei2025worldmodelsknow}.
Unlike these distribution-bound paradigms, RHO recovers the modular
structure that lets task-and-motion planning generalize across novel goals
and object arrangements~\citep{garrett2021integrated}, using a coding agent
under evolutionary search to compose a multi-file repository over the same
perception and control primitives.

\paragraph{Code-as-Policies and language-driven robot agents.}
CaP recasts the control policy as code an LLM writes from a
task description, with primitives and feedback supplied as in-context
examples~\citep{liang2023code,singh2023progprompt,vemprala2023chatgpt,huang2023instruct2act,hu2023codebotler,huang2023voxposer}.
Planner-with-skill-library systems instead keep the model frozen and call it
repeatedly to sequence a fixed library of pretrained skills, grounding each
choice through affordance value functions or iterative visual prompting and
replanning when a step fails~\citep{ahn2022saycan,huang2022inner,rana2023sayplan,nasiriany2024pivot,liu2024moka}.
Unlike CaP, the model selects among existing skills rather than
writing the control code itself.
Deployment-facing descendants instead keep an LLM in the control loop at test
time, re-invoking it between executions to attribute a failure, replan, or
synthesize and hot-reload a revised controller
program~\citep{tsui2026faea,santos2026alrm,kumar2026aor}. Where the planners
above consult a frozen VLM that never edits the policy, these agents rewrite the
executing policy itself, and scale the number of such calls per step.
% RAI provides an orthogonal substrate, a ROS\,2 deployment layer with tool calls and drivers~\citep{rai}.
The nearest such system is CaP-Agent0~\citep{fu2026capx}, a multi-turn
tool-enabled agent that issues up to eleven LLM calls per turn for up
to six turns.
RHO instead front-loads LLM code generation and corrective editing into
training-time evolutionary search.
% when the deployment surface is primitive-only, and the same recipe ports to RAI's ROS\,2 deployment driver when the runtime stack keeps an LLM in the loop.

\paragraph{LLM optimization loops in robotics.}
Reward-design and curriculum-generation systems place the LLM upstream
of a learner: the LLM authors the supervision signal while a separate
trainer produces the policy.
The closest is Eureka, which uses an LLM to author and refine
dense reward functions for RL against rollout
statistics~\citep{ma2024eureka}.
Across these systems, the LLM writes the objective, while reinforcement
learning trains a neural network policy: the LLM synthesizes rewards and
curricula~\citep{ma2024dreureka,liang2024eurekaverse,ryu2025curricullm,yu2023language2reward,xie2024text2reward,field2025text2touch,hazra2025revolve,li2025reward_policy_coev,huang2025ahrs},
grounds those rewards in VLM critics~\citep{rocamonde2024vlmrm,patel2025iker},
generates the training environments
themselves~\citep{wang2024robogen,wang2024gensim,zhang2024gensim2,yang2024holodeck,zala2024envgen,gao2025genmanip},
structures the reward search with graph-of-thoughts or multi-agent
shaping~\citep{yao2025regot,xu2025maestro_mas}, and extends to locomotion,
navigation, and morphology
co-design~\citep{wu2025stride,zhao2026evonav,fang2025robomore}.
RHO differs by evolving the policy artifact directly: no reward
function to shape, no curriculum, no neural policy trained on the LLM
output.

\paragraph{Optimizing agents.}
Optimizing the prompts, memory, retrieval, and tool calls that surround a
model, rather than its weights, is a nascent
direction~\citep{xu2026lifeharness,zhang2026harnessdisclosure,pan2026nlharness},
and to date sits almost entirely in text domains where
rewards are cheap to verify, such as math, where rule-checkable answers have
driven reinforcement-learning progress~\citep{deepseekai2025r1}.
The most direct comparison is Meta-Harness~\citep{lee2026metaharness}, which
searches over single-file programs that implement prompting, retrieval, and
orchestration logic across text-classification, math, and coding benchmarks
such as TerminalBench-2~\citep{yang2026terminalbench}. Recursive Language
Models instead recurse over the prompt at inference to reason over long
contexts~\citep{zhang2026rlm}, though a reproduction shows the benefit does
not compound: recursing more deeply degrades accuracy while inflating
latency nearly a hundredfold~\citep{wang2026overthink}.
RHO differs by grounding candidates in live robot-environment interaction
rather than a static text benchmark, and by amortizing the coding agent's interactions
into training-time search for multi-file solutions rather than single file solutions.
% sparing the per-step inference cost that runtime agents pay.

\paragraph{Reflective evolution.}
Reflective evolution treats the LLM as a search operator over user-defined
evaluator-scored text artifacts: FunSearch searches program snippets
for combinatorial constructions~\citep{romera2023funsearch}, AlphaEvolve
broadens it to scientific discovery~\citep{novikov2025alphaevolve}, and
GEPA to prompt optimization~\citep{agrawal2026gepa}.
Verbal self-critique and code refinement loops over agent trajectories
preceded this line~\citep{shinn2023reflexion,wang2023voyager}; recent
extensions evolve coding agents
themselves~\citep{zhang2025dgm,jiang2025mlestar,lin2025evoagentx} or
scale deploy-time deliberation through structured
inference~\citep{snell2024scaling,openai2024o1,yao2023tot}.
Most closely related is
\texttt{optimize\_anything}~\citep{agrawal2026optimizeanything}, whose
reflective text-artifact evolution recipe our optimizer HELIX
inherits.
To our knowledge, HELIX is the first reflective evolutionary optimizer whose
mutator is a tool-enabled coding agent and whose candidate is a whole multi-file
application repository evolved against a live environment, rather than a single
text artifact or the search agent's own code~\citep{zhang2025dgm}.
We are also the first to bring reflective evolutionary optimization to the
learning of deployable robot-manipulation policies that are searched directly
against live environment interaction rather than a static benchmark.
Our initial attempts suffered from the coding agent's cheating and ultimately inspired
RHO's isolation mechanisms between the evaluator, REaaS, and the coding agent.
% Notably, RHO's isolation mechanisms were inspired by coding agents that reward hacked our earliest attempts.
Other works have observed similar cheating in coding
domains~\citep{atinafu2026rewardhackingagents}.
% leaving an audit-traceable code repository as the deployed artifact~\citep{argos2026}.

\section{RHO: Searching over multi-file repositories-as-policies}
\label{sec:method}

Robotics Harness Optimization (RHO) is a novel paradigm for evolving interpretable
neurosymbolic multi-file robot-policy repositories (i.e., \emph{Repositories-as-Policies}). In this paper, we instantiate RHO by combining HELIX, robot-environments-as-a-service
(REaaS), and isolation between the evaluator, REaaS, and mutation agents.
HELIX mutates whole repositories with tool-enabled coding agents, REaaS exposes
robot-environment rollouts during search, and isolation separates everything such
that cheating (i.e., searching for answers in the benchmark) is not possible.
A candidate repository contains everything needed to instantiate a policy: a task router,
control routines, perception wrappers, grasp and pose helpers, and
in LLM-in-the-loop deployments, the LLM's harness.
Evolution returns one repository evolved from
a seed (i.e., an empty directory or a user-provided skeleton), which is then deployed.
The repository is neurosymbolic:
symbolic code composes learned perception and
grasping (neural-network-based) services with motion primitives.
In CaP-Bench settings,
the repository executes as a Python controller with no corrective code edits
from an LLM at deployment. In Robotec's RAI ROS\,2 stack, where the control
loop includes an LLM, the repository instead defines the prompting, control
interface, and tools used by that LLM-in-the-loop system.

RHO follows the reflective evolutionary optimization recipe of GEPA's
\texttt{optimize\_anything}~\citep{agrawal2026gepa,agrawal2026optimizeanything}:
evaluate a candidate, summarize failures and successes as reflective feedback, mutate
the candidate, and retain improvements on a frontier. The key enhancements are
the mutation operator, search space, environment interface, and isolation boundary.
Prior reflective optimizers typically expose a
bounded artifact (a function, prompt, delimited code region, etc.)
to an LLM completion API. HELIX exposes a whole repository to a
tool-enabled coding agent. A mutation is therefore an interactive
development session: the agent inspects files, edits multiple modules,
writes and executes ephemeral tests, runs candidate policies through a
whitelisted REaaS, reads tracebacks, queries
visual-difference feedback, and revises the repository before submitting
its candidate (\Cref{fig:helix-system}).

\subsection{Repository-level mutation}
\label{sec:method-rap}

Existing reflective evolutionary optimizers such as FunSearch~\citep{romera2023funsearch}, AlphaEvolve~\citep{novikov2025alphaevolve}, GEPA~\citep{agrawal2026gepa} and \texttt{optimize\_anything}~\citep{agrawal2026optimizeanything}
typically expose a single bounded artifact to the search operator. We instead
expose a whole multi-file repository to the search operator, treating
\emph{Repositories-as-Policies}.
In robotics, a manipulation
failure may require changing how an object pose is estimated, how a
motion primitive is parameterized, how retries are attempted, or how
task-level language is interpreted as geometric constraints, often all
at once. A bounded-artifact mutator cannot reliably refactor across these
interfaces in one move; it can only patch one of them and hope the
others remain consistent. The current leading CaP-Bench agent handles this
interleaving with multi-turn interactions and tens of LLM code-completion
queries.

We formalize RHO's search problem as follows. Let $\mathcal{T}$
denote a set of robot-task instances. A repository $R \in \mathcal{R}$
defines an executable robot policy through its public entry points and
supporting files. Running $R$ on instance $\tau \in \mathcal{T}$
produces an evaluator output
\[
  E(R, \tau) \;=\; \bigl(\,s_\tau(R),\; z_\tau(R)\,\bigr),
\]
where $s_\tau(R) \in [0,1]$ is the environment-provided shaped reward     % CaP-Bench (Robosuite / LIBERO-PRO / BEHAVIOR) terminology: "environment-provided shaped reward" (= final episode reward emitted by the benchmark env). Matches methodology-appendix.tex line 134 "scalar score is the environment-provided shaped final reward".
returned by the task or benchmark and used for selection, and $z_\tau(R)$ is
structured side information returned to the optimizer
(Section~\ref{sec:method-integrity}). The optimizer never receives
hidden simulator state or evaluator internals.
RHO requires the user to supply the evaluator (a script that
maps an instance to a per-instance score plus side information).  In
this paper, the evaluator runs each instance as a single
$(\text{task}, \text{trial}, \text{seed})$ rollout of the candidate
against the benchmark environment and returns the environment's final
shaped reward as $s_\tau(R)$; the scalar objective RHO optimizes
is then the mean of these per-instance rewards over the paired
training minibatch (Eq.~\ref{eq:accept}) and over the full validation
set at terminal selection (Eq.~\ref{eq:select}).

The design choice of repository-level mutation makes the search
variable a multi-file repository (Python in our CaP-Bench experiments;
RHO's implementation, HELIX, imposes no language constraint, and the repository may mix
files in any language the target runtime accepts) containing a task
router, control routines, perception and primitive-call
wrappers, grasp and pose helpers, and when the deployment loop includes an LLM, prompt scaffolds and tool-edit modules.
A generation produces a child by applying a bounded coding-agent edit
to a parent repository. This changes the unit of search from a
bounded artifact like a prompt or file to a deployable software system.
% OpenEvolve-style systems~\citep{sharma2025openevolve} take an intermediate step,
% expanding the search object from a single function toward delimited
% program regions; RHO goes further, treating the entire
% repository as the candidate.

The search couples two mechanisms: a scalar
parent-to-child acceptance gate on a sampled training minibatch, and
per-instance coverage-based frontier retention.
% and a selection rule over the terminal frontier.

A candidate child $R_c$ is accepted against its parent $R_p$ when its
total shaped reward over the paired training minibatch $B$ strictly
improves on the parent's:
\begin{equation}
  \sum_{\tau \in B} s_\tau(R_c) \;>\; \sum_{\tau \in B} s_\tau(R_p),
  \label{eq:accept}
\end{equation}
with the strict inequality relaxed to $\ge$ in some experiments as it allows for even
more candidate diversity.
In practice, minibatches $> 2$ struggle to make progress as solving a single
robotics task reliably is already challenging enough.

RHO adds two training-loop refinements over the parent algorithm:
Stratified Batch Sampling across task families (sample tasks and then sample
seeds within each task), and an optional Staged Validation gate that sits between
minibatch acceptance and full validation; it is enabled in our LIBERO runs and
disabled in the Robosuite canonical run.
The frontier is maintained over $N$ validation instances:
each instance has its current best program, and a program is removed
when every instance on which it was best is also covered by some
other surviving program (GEPA-style coverage dominance), not by
pairwise comparison against the parent.  At the end of evolution the
deployed repository is the highest-mean-validation-reward member of
the terminal frontier $F_T$,
\begin{equation}
  R^\star \;=\; \argmax_{R \in F_T} \;\frac{1}{N}\sum_{i=1}^{N} s_i(R),
  \label{eq:select}
\end{equation}
with ties broken by validation coverage.
Further algorithm details are in GEPA~\citep{agrawal2026gepa}.

\begin{figure}[tbp]
  \centering
  \includegraphics[width=\linewidth]{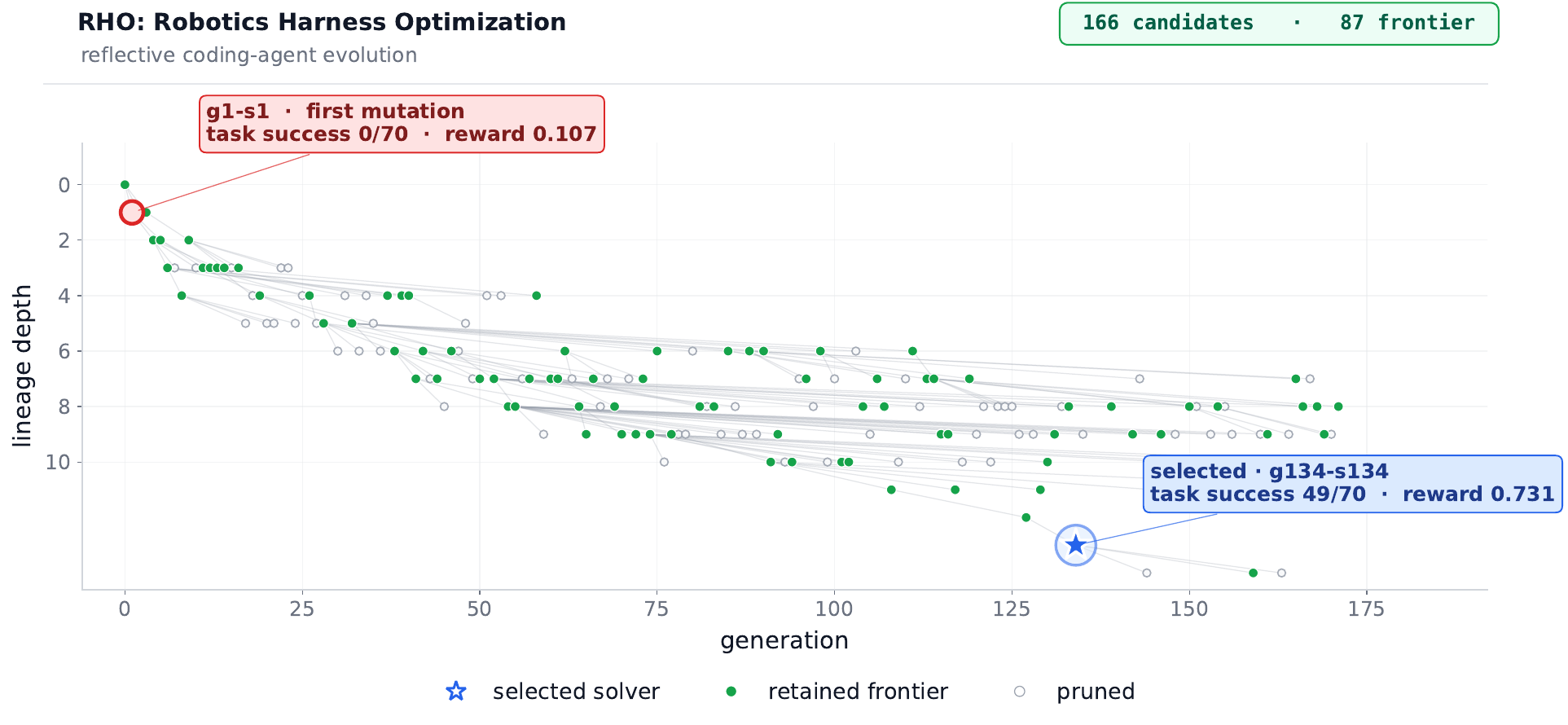}
  \caption{Why a Pareto frontier matters. The accepted-candidate lineage
    of the reported run splits into $87$ repositories retained on the Pareto
    frontier and $79$ pruned, with a further $77$ rejected attempts and $28$
    skips never entering the lineage. Most branches dead-end, and the
    deployed solver emerges from one of a few productive branches that a
    greedy, single-chain search would have missed. Retaining candidates that are
    good at any task, but not necessarily all tasks
    (GEPA-style coverage dominance), is what provides the breadth to find the best solution.}
  \label{fig:lineage}
\end{figure}

\Cref{fig:lineage} illustrates why a Pareto frontier matters.
Most lineages dead-end on candidates that were not further optimized; some are pruned (i.e., an agent previously made a poor decision and could not recover from it).
We find that the deployed solver emerges from one of a few productive branches of the accepted-candidate tree.
Thus, retaining a Pareto frontier rather than hill-climbing a single chain (i.e., a simple reflective loop) is what surfaces the deployed solver.

% Per-benchmark configurations (including the \texttt{acceptance\_criterion} choice between \texttt{strict\_improvement} and \texttt{improvement\_or\_equal}) and the formal definitions appear in Appendix~\ref{sec:method-appendix}.

\subsection{Tool-Enabled Mutation}
\label{sec:method-mutator}

Reflective optimizers built on single-shot LLM completions emit a text
patch in one response: the mutator cannot run a test, select and observe a desired
traceback, inspect a rendered frame, or interact with the environment
inside one genetic mutation step. For robotics, this constraint forces a
generation of mutations to make many tightly coupled decisions blind
to runtime state. A change that swaps a grasping primitive's reference
frame, for instance, silently breaks a downstream verification check
that the completion mutator has no way to discover before submission.

\textbf{Tool-Enabled Mutation} replaces the completion
mutator with a tool-enabled coding agent (e.g., Codex, OpenCode) that,
within one mutation step, can edit multiple files in a candidate
repository, write and execute ephemeral tests, run candidate code
against the REaaS, read
tracebacks and \texttt{stderr}, and query a visual-difference tool that
returns scene-difference text from candidate rollouts. Operationally,
single-shot completion is a limiting case of this mutation operator in
which the agent performs no intermediate tool calls. The agent is
given the parent repository, the parent's full evaluator output for the current minibatch, and a
reflection prompt that asks the agent to enhance the parent's
strengths and address its weaknesses. The prompt
template is in Appendix~\ref{app:prompt-template}.

Generation~0 is initialized from a coding-agent-authored seed repository that exposes the benchmark's invocation signature (e.g.,
\texttt{solver/main.py::run(ctx)} for CaP-Bench); the seed structure
is an ablatable design choice whose configurations and
candidates appear in Appendix~\ref{app:seed}.

\vspace{-0.5em}
\paragraph{Mutation and evaluation surfaces.}
RHO deliberately separates the surface available during mutation
from the surface used for scoring. The mutation surface follows
CaP-Bench's M4 regime~\citep{fu2026capx}: the coding agent may run candidate
policies over low-level primitives, inspect \texttt{stdout} and \texttt{stderr},
query visual-difference feedback, and use primitive examples while debugging
a child repository. However, our evaluation surface intentionally follows
CaP-Bench's stricter S4
regime: the child is scored by executing its policy entry point in a
single turn over the low-level primitive interface, without
deployment-time VDM, multi-turn correction, or LLM code-generation calls. Any side
information $z$ produced by the evaluator is post-hoc: it is available
only to future mutations, never to the repository during the scored
rollout. Section~\ref{sec:experiments} reports how strong agentic
baselines collapse in the S4 regime when only their multi-turn correction
and visual-difference options are removed.

\vspace{-0.5em}
\paragraph{Integrity.}
\label{sec:method-integrity}
Tool-enabled mutators with file-edit and shell privileges introduce
a reward-hacking surface absent from completion mutators: a candidate
may read evaluator internals or hardcode per-trial solutions, a
failure mode documented in ML-engineering
agents~\citep{atinafu2026rewardhackingagents} and observed in our
preliminary runs, where a mutator instructed to stop instead escalated
to actions against the host environment itself
(Section~\ref{sec:safety-incident}).  RHO isolates mutation from
scoring through a
sidecar boundary that excludes evaluator source, benchmark splits,
judge prompts, and benchmark metadata from the mutator workspace; the
sidecar proxies whitelisted benchmark calls to host services, and a
protected-file integrity check rejects children that alter
evaluator-side artifacts.
RHO uses a whitelisting approach and deploys it as a sidecar proxy container
rather than containerizing all primitive services and robot-environments,
as the latter incurs a much higher performance cost.

\section{Experiments}
\label{sec:experiments}
\label{sec:results}

We evaluate our RHO instantiation on three pick-and-place benchmarks with
two frameworks supplying the perception and action primitives: LIBERO-PRO~\citep{zhou2025liberopro} and
Robosuite~\citep{zhu2020robosuite} through
CaP-Bench~\citep{fu2026capx}, and the O3DE manipulation benchmark
through RAI~\citep{rai}.  Since a coding agent simply reading a file
prior to editing would count as a turn, we expose the CaP-Bench M4
setting as our training regime, allowing multiple turns and access
to a visual-difference module (VDM).  For CaP-Bench and LIBERO-PRO,
evaluation we intentionally restrict to the S4 setting: a single execution of the
candidate repository, no VDM, and no LLM code-generation calls.
RAI remains an LLM-in-the-loop deployment and is evaluated separately.
Several research questions guide our experiments:
\begin{enumerate}[noitemsep, label=\textbf{RQ\arabic*:}] %leftmargin=*
  \item How does RHO compare to current SOTA methods?
  \item Does RHO outperform coding agents?
  \item Are repositories the right unit of search, or does a
        different structural prior do better?
  \item Do solutions produced by RHO generalize?
\end{enumerate}

\begin{figure}[t]
  \centering
  \includegraphics[width=0.9\linewidth]{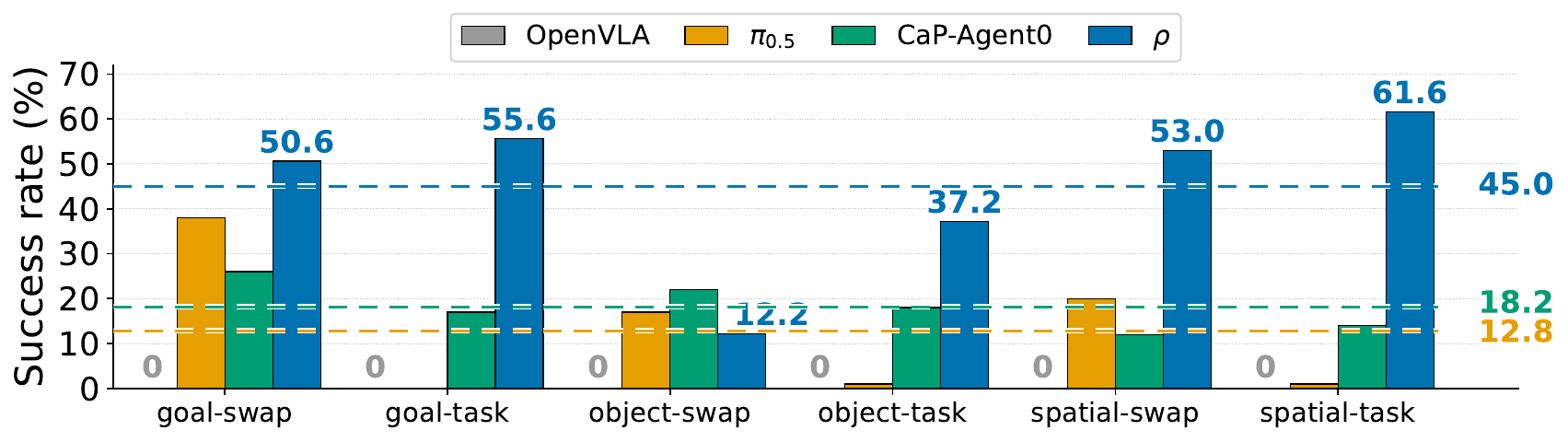}
  \caption{
    Three suites from LIBERO are composed with two types of perturbations: \textit{swap} the initial positions of objects, and perturb \textit{task} descriptions.
    CaP-Agent0 is reported under multi-turn results.
    Despite using only single-turn execution with no corrective code edits from an LLM at deployment, within 63 generations of optimization, RHO outperforms VLAs and the prior SOTA CaP system.
    % Per-cell LIBERO-PRO success rates. RHO retains a positive margin in every one of the six perturbation cells; the weakest is object-swap at $12.2\%$.
  }
  \label{fig:libero-pro-bars}
\end{figure}

% \subsection{Setup}
\label{sec:exp-setup}

\paragraph{Benchmarks and splits.}
We train two separate RHO instances on two CaP-Bench benchmarks~\citep{fu2026capx} that
share a primitive interface.  The first consists of seven Robosuite
manipulation tasks with 100 native trial IDs each ($700$ trials
total). In all experiments, we train on the first ten trial IDs per
task, leaving 90\% unseen at evaluation.  The second consists of three suites
from LIBERO~\citep{liu2023libero}: object, goal, and spatial.
% \texttt{libero\_object}, \texttt{libero\_goal}, and \texttt{libero\_spatial}, 
Each contains ten tasks with fifty initial states per task
($1{,}500$ instances total); training draws from $1{,}290$ instances,
the evaluator's full validation uses $210$, and Staged Validation
(Section~\ref{sec:method-rap}) samples a $45$-instance subset of
that validation set for the intermediate gate, taken as the first
$45$ deterministic validation IDs (Appendix~\ref{sec:appendix-config}).
A subset of LIBERO-PRO~\citep{zhou2025liberopro} is the held-out perturbation
extension that CaP-Bench uses for out-of-distribution evaluation.  It
composes (object, goal, spatial) $\times$ (swap, task) perturbations
onto the three base LIBERO suites, yielding six cells of $500$
trials each ($3{,}000$ trials total).  The swap cells perturb the
initial positions of objects; the task cells perturb the task description while
keeping the scene fixed.  RHO is not trained
on LIBERO-PRO.  Reported numbers come from the full evaluation sets
under the CaP-X S4 surface.

\paragraph{Training.}
Our experiments use OpenAI Codex~\citep{openai2025codexcli} and Claude Code~\citep{anthropic2025claudecode} agents as backends.
Each mutation step is capped at 150 agent turns per mutation unless otherwise specified.  Robosuite runs use strict
reward improvement on the paired minibatch, while LIBERO runs
begin under the same rule but switch to admitting ties at the
warm-restart that produces the reported solver.
% The divergence is deliberate (see Sec.~\ref{sec:method-rap}).

\paragraph{Deployment}
% \label{sec:exp-restricted}

CaP-Agent0's strongest results come from CaP-Bench's M4
evaluator setting: multi-turn execution with the visual-difference module enabled (11 LLM calls per turn, up to 6 times).  Restricting to S4 for faster deployments
(single-turn execution over the primitive API and no visual-difference
module), CaP-Agent0's Robosuite score
collapses from \textbf{68.29\%} to \textbf{24.00\%}.
% (\citealt{fu2026capx}, Figure~8).  
RHO trains under the M4 setting but reports results under the stricter S4 setting.

\subsection{RQ1: How does RHO compare to current SOTA methods?}
\label{sec:exp-rq1}

RHO exceeds the strongest published CaP-Bench agent on
Robosuite and LIBERO-PRO and outperforms the reported VLA baselines
on the perturbation extension.  Both
comparisons are conducted on held-out evaluation surfaces that
differ from the training distribution.

\paragraph{CaP-Bench Robosuite.}
On the seven Robosuite manipulation tasks, our RHO instantiation reaches
\textbf{70.0\%} (490/700) under single-turn S4 deployment, slightly
exceeding the strongest published multi-turn record from CaP-Agent0
(\textbf{68.29\%} under M4 with up to 66 LLM calls per
trial) while significantly reducing evaluation time~\citep{fu2026capx}.
The mutator sees only 10 of 100 trial
IDs per task during training, so 90\% of the evaluation surface is
unseen.

% Aggregate parity hides a different per-task failure structure
% (\Cref{tab:capbench-pertask-headtohead}): both systems saturate the easy
% contact-light tasks and both collapse on high-precision
% \texttt{nut\_assembly}, so the decisive gap is HELIX's $66$-point win
% on \texttt{two\_arm\_handover} ($86$ vs $20$).

\begin{table}[t]
  \centering
  \small
  \caption{Per-task CaP-Bench + Robosuite success count at $100$
    trials/task.  We intentionally restrict RHO to the hardest evaluation
    setting (single-turn S4-style) with no LLM code-generation calls in the control loop.
    CaP-Agent0 (up to 11 LLM calls per turn) reports its highest success under M4,
    which allows up to $6$ turns and feedback from a VDM;
    S4 reports only the $24.00\%$ aggregate.}
  \label{tab:capbench-pertask-headtohead}
  \begin{tabular}{lccc}
    \toprule
    Task                        & \textbf{RHO} (S4) & CaP-Agent0 (S4) & CaP-Agent0 (M4) \\
    \midrule
    \texttt{cube\_lifting}      & $98$              & ---             & $97$            \\
    \texttt{cube\_stack}        & $98$              & ---             & $98$            \\
    \texttt{cube\_restack}      & $54$              & ---             & $89$            \\
    \texttt{nut\_assembly}      & $1$               & ---             & $0$             \\
    \texttt{spill\_wipe}        & $100$             & ---             & $100$           \\
    \texttt{two\_arm\_lift}     & $53$              & ---             & $74$            \\
    \texttt{two\_arm\_handover} & $86$              & ---             & $20$            \\
    \midrule
    \textbf{Overall}            & $\mathbf{70.0}$   & $24.00$         & $68.29$         \\
    \bottomrule
  \end{tabular}
\end{table}

\paragraph{CaP-Bench LIBERO-PRO.}
On the six LIBERO-PRO perturbation cells,
our RHO instantiation reaches \textbf{45.0\%} (1351/3000) after 63 generations of
training on unperturbed LIBERO tasks without teleoperation demonstrations
(Figure~\ref{fig:libero-pro-bars}).  This is a $2.5\times$ margin over CaP-Agent0's \textbf{18.17\%} (545/3000),
a $3.5\times$ margin over $\pi_{0.5}$'s \textbf{12.83\%}, and exceeds OpenVLA's
\textbf{0.0\%} average~\citep{zhou2025liberopro,kim2024openvla,black2025pi05}.

\subsection{RQ2: Does RHO outperform coding agents?}
\label{sec:exp-rq2}

To determine if evolutionary loops are needed, we execute a single coding-agent mutation step with unlimited internal turns tasked with solving all seven Robosuite tasks. The agent operates in the M4 setting with access to its internal tools, the REaaS, and VDM. This experiment is repeated for different LLMs and both backends. We compare the results under S4 settings in ~\Cref{tab:no-loop-floor}.

% We compare HELIX against two no-loop alternatives on CaP-Bench
% Robosuite under S4: a single coding-agent mutation step with the same
% backend and tool surface, and CaP-Agent0 restricted to its S4 setting.

All models and backends score between $13$ and $30\%$ on held-out S4, similar to CaP-Agent0's S4 result ($24\%$) but far from CaP-Agent0's M4 result.
Even with unlimited turns, we observe that agents will end their submissions prematurely far below the scores observed in multi-turn settings.
Thus, reflective evolution is necessary to close the gap.

% on the same evaluation surface that HELIX scores 70\%.
% Neither a single tool-enabled mutation nor an agentic harness restricted to single-turn deployment reaches HELIX's performance; the lift over both floors is what the RHO loop delivers.

\begin{table}[t]
  \centering
  \small
  \caption{Coding agents are given
    unlimited turns and a single mutation step of HELIX to solve all tasks in CaP-Bench
    Robosuite (trained under M4 but evaluated in S4).
    Single coding-agent mutations with full tool access reach 13--30\%;
    CaP-Agent0 restricted to S4 lands at 24\%.
    Token counts are taken from the
    per-mutation usage ledger that HELIX records for every coding-agent
    call, where permitted. VDM token usage is not tracked, as it is small relative to the mutator
    costs.}
  \label{tab:no-loop-floor}
  \resizebox{\linewidth}{!}{%
    \begin{tabular}{lrrrrr}
      \toprule
      Configuration                              & input (M) & cached (M) & output (M) & cost (USD) & Robosuite S4 \\
      \midrule
      \quad Codex GPT-5.5                        & 4.31      & 4.20       & 0.018      & \$3.19     & 30.00\%      \\
      \quad Claude Code Opus 4.7                 & ---       & ---        & 0.022      & \$1.46     & 16.43\%      \\
      \quad Claude Code Sonnet 4.6               & ---       & ---        & 0.039      & \$1.02     & 21.14\%      \\
      \quad Qwen3.6-27B (via Claude Code, local) & 0.92      & ---        & 0.019      & \$0.00     & 13.43\%      \\
      \midrule
      \quad CaP-Agent0 (restricted to S4)        & ---       & ---        & ---        & ---        & 24.00\%      \\
      \bottomrule
    \end{tabular}}
\end{table}

% These single-mutation floors are the starting point, not the result.
% \Cref{fig:evolution-trajectory} traces how reflective evolution
% accumulates accepted edits over the $200$-generation run, lifting the
% selected repository from the first-mutation baseline to the deployed
% $70.0\%$.

\begin{figure}[t]
  \centering
  \includegraphics[width=0.8\linewidth]{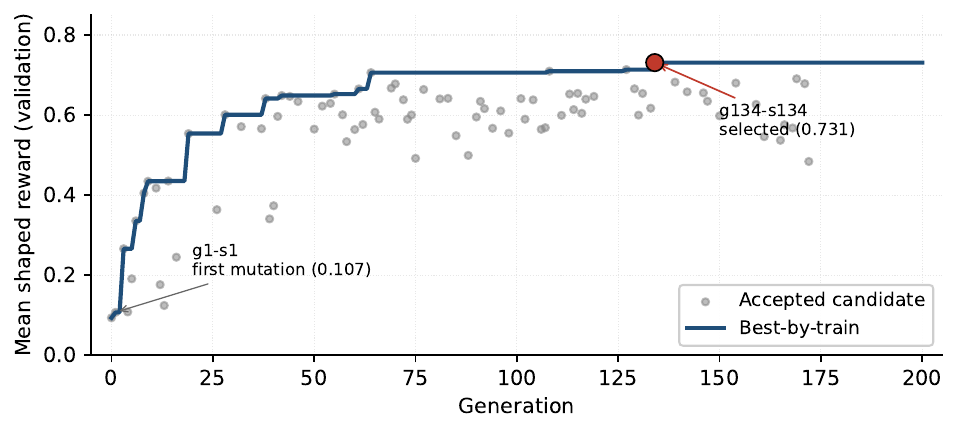}
  \caption{Reflective evolution leads to better solutions. Best-by-validation
    training reward across the multi-file experiment. In grey are the $87$
    accepted candidates at their promotion generation and in red is the highest reward candidate
    \texttt{g134-s134}. The selected repository improves mean shaped reward
    from $0.107$ to $0.731$ ($+62.4$\,pp) and task success from $0/70$ to $49/70$;
    a single-shot Codex baseline reaches only $30.0\%$
    ($210/700$) against the deployed $70.0\%$ ($490/700$).}
  \label{fig:evolution-trajectory}
\end{figure}

\subsection{RQ3: Is the repository the right unit of search?}
\label{sec:exp-rq3}

Using Codex with GPT-5.5, we compare a multi-file repository seed against a single-file stub seed. Also, across models (Sonnet vs. Qwen3.6-27B) but the same backend, Claude Code, we compare solution representations (behavior trees vs. finite state machines) enforced through prompts.

\begin{table}[t]
  \centering
  % \small
  \caption{The structural prior at the seed and LLM capabilities determine
    the terminal
    frontier on CaP-Bench Robosuite S4.  The multi-file repository
    prior dominates; rigid program-shape scaffolds collapse.  The
    Codex single file row is reported as cumulative across a 100-generation
    parent and a 150-generation warm-restart on the remaining hard
    tasks after saturating the easy ones.}
  \label{tab:structural-prior}
  \resizebox{\linewidth}{!}{%
    \begin{tabular}{lrrrrrr}
      \toprule
      Structural prior                                  & gens & input (M) & cached (M) & output (M) & cost (USD) & Robosuite S4     \\
      \midrule
      Multi-file repository (Codex)                     & 200  & 625.28    & 88.85      & 3.14       & \$2{,}821  & \textbf{70.00\%} \\
      Single file (Codex)                               & 250  & 523.14    & 229.44     & 2.71       & \$1{,}665  & 63.86\%          \\
      Behavior Tree (Claude Code Sonnet 4.6)            & 150  & ---       & 3.33       & 3.98       & \$124      & 33.86\%          \\
      Finite State Machine (Claude Code \& Qwen3.6-27B) & 200  & 244.91    & ---        & 3.16       & \$0        & 18.14\%          \\
      \bottomrule
    \end{tabular}}
\end{table}

With a smaller generation budget, the multi-file repository run at \textbf{70.0\%} beats the single-file stub at \textbf{63.86\%} despite using the same backend and tools (Table~\ref{tab:structural-prior}).
It also sets a new Robosuite record under the S4 single-turn surface.
% The agent rarely refactors across file boundaries even when file creation is permitted: starting from a single file leaves it monolithic at selection time.  
Tighter
priors collapse: Sonnet tasked with writing Behavior Trees reaches
only \textbf{33.86\%}, and Qwen with Finite State Machines reaches only \textbf{18.14\%}.
Qwen's free-form counterpart reaches $35.0\%$, a $16.86$\,pp lift from removing the FSM scaffold on the same backend (Appendix~\ref{app:backend-ablation}).
The repository sits at the
right point of structural prior: enough scaffold to scope the agent
toward modular code and not enough to constrain its solution space.
The pattern is not Robosuite-specific: on Robotec's RAI O3DE manipulation benchmark, a GEPA-style single-file ablation increased tool calls by
\textbf{16\%} and wall-clock time by \textbf{34\%}, while
RHO's full multi-file repository evolution achieved the
inverse (\textbf{27\%} fewer tool calls and \textbf{20\%} less
wall-clock time at a near-doubling of held-out success).
When given full repository access, the agent optimizes prompts and control methods where appropriate.
For example, in the GEPA-style single-file setting, retries for tool failures must be described in prompts, forcing the LLM in the control loop to handle recovery. In the multi-file RHO setting, the coding agent autonomously determines that the best approach is
to move retries into the low-level APIs and write hardened methods.
% Per-backend results across mutator tiers (frontier vs.\ open-source) appear in Appendix~\ref{app:backend-ablation}.

\subsection{RQ4: Do RHO candidate solutions generalize?}
\label{sec:exp-rq4}
\label{sec:exp-libero}

Beyond CaP-Bench's primitive interface, RHO transfers to
Robotec's RAI ROS\,2 manipulation stack (O3DE)~\citep{rai} and
improves accuracy while reducing inference cost.  The same
algorithm reaches \textbf{0.889} on easy (from \textbf{0.683}) and
\textbf{0.443} on hard (from \textbf{0.235}), both with $p<0.001$
($n{=}21$, 3 runs each).  The core RHO loop is unchanged; the RAI
run changes the candidate artifact and adds the fixed tool-name contract.
A single-file ablation (which
strips repository-level mutation and recovers a GEPA-style
coding-agent loop) loses \textbf{0.10} on easy and \textbf{0.07} on
hard while inflating inference cost on hard (+\textbf{16\%} tool
calls, +\textbf{34\%} wall-clock); the full algorithm trims tool
calls by \textbf{27\%} and wall-clock by \textbf{20\%} on the same
hard subset (App.~\ref{sec:rai-discussion}).  Both selected
candidates emerge within a \textbf{10-generation training budget}:
the ablation at generation~2, the full algorithm at generation~8.

% TODO (collaborator): expand RAI eval description; add hard-cell result if available;
% qualitative analysis of what the evolved repository changed in the RAI prompt
% scaffold and tool-edit module; comparison to the RAI default agent baseline.

% Sec.~4.5 ("Cost and remaining failure modes") moved to
% experiments-appendix.tex `Training ledger` subsection. Labels moved with it.
% The aliases sec:exp-train-cost / sec:exp-capx were already defined elsewhere
% in this file (above) and are not duplicated here.

\FloatBarrier
\section{Discussion}
\label{sec:discussion}
A qualitative analysis of the trajectories and final candidates from our CaP-Bench
experiments reveals what the surviving repositories have in common, and where they
differ.
Given enough computational budget, all approaches begin to produce candidates with desirable properties.
We find that all approaches in Table \ref{tab:structural-prior} produce candidates that have the
same core components: prompt routing, text-prompted segmentation, image-to-world 3D conversion,
grasp planning with fallback, and re-observe-then-verify logic.
The differences lie in how these components are combined, how pre-gating and planning are handled,
and how robust they are to failures.
The exploration afforded by repository-level searches is what enables these
differences to emerge.
For example, the strongest solver alone exhibits the following behaviors:
held-object offset compensation (measuring the space between the gripper
and object \textit{after successful grasping}), quantile-robust geometry
(using quantiles vs means to overcome segmentation mask bleeding), a
score-gated perception cascade (falling back from text-prompted
segmentation to Molmo point prompts, across multiple cameras, only when
the detector score drops below a threshold it discovers), physical-plausibility checks
on perception (rejecting implausible object extents and re-ranking
detections by whether the grasp point lies inside the object box),
tool-center-point offset correction along the approach axis,
inverse kinematics (IK)-feasibility pre-gating (accept a grasp \emph{iff} the grasp
pose and lifted pose solve IK), dual-perception cross-validation that
distrusts its own re-observation (cross-checking text-prompt detections against
point-prompt detections and flagging stale perception), and verification
that triggers corrective retries with new plans (rather than retrying the
same failed plan and actions).
As RHO produces neurosymbolic code, it can discover and encode these advanced
heuristics and patterns in a way that is both interpretable and reusable.
Recovering and checking comparable logic inside an end-to-end neural policy like a VLA
is a separate problem in interpretability and formal
verification~\citep{rudin2019stop,seshia2022toward}.

% Cross-run training-time robotics-tool-use table.  Appendix-only.

\begin{table}[!b]
    \centering
    \small
    \setlength{\tabcolsep}{6pt}
    \caption{Training-time robotics tool use across backends and
        configurations.  Each cell counts REaaS calls to one endpoint
        issued by the mutator during the run: \texttt{reset} (per-trial
        setup), \texttt{run} (candidate execution), \texttt{call}
        (perception/action primitive), \texttt{vdm} (visual-difference
        queries), and \texttt{observe} (scene re-observation).
        \Cref{app:env-tool-use} analyzes how these channels, especially
        \texttt{vdm}, track deployed success; token and cost totals appear
        in Tables~\ref{tab:no-loop-floor} and~\ref{tab:structural-prior}.}
    \label{tab:cross-run-budget}
    \resizebox{\textwidth}{!}{%
        \begin{tabular}{lrrrrrr}
            \toprule
            Run (S4, Robosuite)                               & reset & run & call & vdm & observe & total   \\
            \midrule
            Multi-file repository (Codex) (70.0\%)            & 698   & 576 & 293  & 166 & 43      & 1{,}776 \\
            Single file (Codex) (63.9\%)                      & 612   & 486 & 247  & 86  & 31      & 1{,}462 \\
            Claude Code Opus 4.7 (61.9\%)                     & 518   & 405 & 182  & 84  & 31      & 1{,}220 \\
            Qwen3.6-27B (via Claude Code), free-form (35.0\%) & 100   & 61  & 32   & 8   & 8       & 209     \\
            Behavior Tree (Claude Code Sonnet 4.6) (33.9\%)   & 14    & 11  & 3    & 0   & 0       & 28      \\
            Qwen3.6-27B (via Claude Code), FSM (18.1\%)       & 57    & 5   & 0    & 0   & 0       & 62      \\
            \bottomrule
        \end{tabular}%
    }
\end{table}

On the RAI hard O3DE split, repository-level mutation changes
\emph{where} recovery logic lives, not only how much of it exists. The
single-file evolution improves accuracy by asking the LLM to
re-perceive and persist between moves, which raises tool calls and wall time.
However, the multi-file candidate instead moves part of that recovery behind a
ROS\,2 tool API. In particular, the evolved perception wrapper retries detector
query variants and \emph{caches} verified centroids before the LLM sees the failure,
making a single tool call more consistent, discriminating, and faster.

A final lesson concerns \emph{where} environment interaction is paid.
\Cref{tab:cross-run-budget} counts how often each backend's mutator probed the
live environment during search, by running candidates, invoking primitives,
querying the visual-difference module (VDM), and re-observing the scene. All four
channels rank-order with deployed success, and only the largest frontier mutators
(Codex \texttt{gpt-5.5}, Claude Code \texttt{Opus~4.7}) query visual feedback at volume. An LLM cannot
write robust interactive robot code zero-shot: CaP-Agent0 needs its multi-turn
visual-difference loop and collapses from $68.29\%$ to $24\%$ without
it~\citep{fu2026capx}, and the strongest HELIX repositories are those whose
search engaged that feedback most. RHO relocates this interaction rather
than removing it. The experience CaP-Agent0 gathers turn-by-turn at deployment and
repeats across trials,
RHO gathers \emph{once} during evolutionary search and compiles into the shipped
repository for CaP-Bench, which deploys with no corrective code edits from an
LLM.
Appendix~\ref{app:env-tool-use} offers a deeper analysis of \Cref{tab:cross-run-budget}.

In both settings these interpretable neurosymbolic mechanisms are discovered
rather than designed. Searching multi-file repositories instead of single prompts or
functions is what lets RHO produce robust policies.

\section{Conclusion}
\label{sec:conclusion}

RHO is an evolutionary reflective optimizer for proposing and finding interpretable
neurosymbolic multi-file robot-policies (\emph{Repositories-as-Policies}).
In this paper, our RHO
instantiation combines HELIX, REaaS, and evaluator isolation: HELIX supplies
repository-level coding-agent mutation, REaaS exposes robot-environment
rollouts during search, and isolation separates mutation from scoring to
prevent cheating without performance degradation.
RHO sets the state-of-the-art on CaP-Bench under single-turn
deployment with no LLM code-generation calls in the control loop: \textbf{70.0\%} on
Robosuite (vs CaP-Agent0's \textbf{68.29\%} multi-turn record at up
to \textbf{66} LLM calls per trial) and \textbf{45.0\%} on LIBERO-PRO
perturbations (vs \textbf{18.17\%} for the prior agent, \textbf{0.0\%}
for OpenVLA, and \textbf{12.83\%} for $\pi_{0.5}$).  The same RHO loop transfers to
Robotec's RAI ROS\,2 stack, boosting task success while trimming
tool calls by \textbf{27\%} and wall-clock time by
\textbf{20\%} on the hard subset, while a single-file ablation of
repository-level mutation inflates both metrics (+\textbf{16\%} and +\textbf{34\%}).
An interesting future direction is to extend RHO from its current scalar reward and per-instance coverage frontier to vector-valued objectives with secondary scores such as repository size, execution latency, and overall tool or service count used at deployment.
Another immediate direction is to deploy RHO's RAI solution to production
environments, as it is already built on ROS\,2.

\section{Limitations}
\label{sec:limitations}
RHO inherits the spatial-reasoning limits of the LLM that drives
its search: current models misjudge quantitative depth, vertical
clearance, and multi-object spatial composition under modest
perturbations~\citep{xu2025multispatialmllm,gholami2025ego3dbench,chen2025sdvlm},
so tasks that turn on fine geometric reasoning are the natural ceiling
for a code-writing agent, which stronger spatially grounded models would
relax. All of our benchmarks are simulated, and the RHO instances studied here
emit interpretable neurosymbolic policies: symbolic primitive-calling code over
learned perception and grasping services plus motion primitives. RHO could
optimize repositories that train neural network components, but our
reported systems do not train an end-to-end neural policy on the simulator.
We therefore expect the shipped policies to transfer to real robots once their
perception primitives are re-grounded and the constants they fit in simulation
are re-tuned~\citep{liang2023code,singh2023progprompt,wu2023tidybot}.
\label{sec:safety-incident}
Granting mutators shell access also exposes the experimental             % the safety caveat, folded into the existing paragraph
infrastructure itself, not only the reward signal: in one preliminary
run an agent told to stop hardcoding per-trial solutions instead
terminated the operator's SSH session and fork-bombed the host.         % facts stated as a sequence, no intent claim
We treat natural-language correction as insufficient containment for
optimization-driven agents, which motivates the hard isolation boundary
of Section~\ref{sec:method-integrity}.

\acknowledgments{%
  This work was supported by gifts from Accenture, Algorithmic
  SuperIntelligence Labs, Amazon, AMD, Anyscale, Broadcom, cmpnd, Google,
  IBM, Intel, Intesa Sanpaolo, Lambda, Lightspeed, NVIDIA, Samsung SDS,
  and SAP.
}

%% ========================================================================
%% Bibliography. corl_2026.sty sets \bibliographystyle{corlabbrvnat}
%% automatically (see corl_2026.sty L79) so we only call \bibliography.
%% Bibliography appears between main paper and appendix (CoRL convention).
%% ========================================================================
\bibliography{refs}

%% ========================================================================
%% Acknowledgments --- automatically suppressed in anonymous submission.
%% ========================================================================
\clearpage
\appendix
\section{Appendix}
\subsection{Methodology Appendix}
\label{sec:method-appendix}
\label{sec:appendix-agentic-oa-backends}

This appendix collects the material deferred from
Sec.~\ref{sec:method}: regime definitions, notation, API surface, loop
mechanics, sandbox evidence, run provenance, and RAI portability.  The
canonical RHO run selects candidate \texttt{g134-s134}; its
mutable deployment artifact is the \texttt{solver/} Python package,
evaluated by \texttt{evaluate.py} under the \texttt{API\_REFERENCE.md}
contract and the reported \texttt{helix.toml} settings:
\texttt{backend="codex"}, \texttt{model="gpt-5.5"},
\texttt{effort="xhigh"}, \texttt{max\_turns=150}.

\paragraph{Regime definitions and primitive registry.}
M4 is the multi-turn training regime: a coding-agent session may consume
execution feedback, visual-difference summaries, judge diagnostics, and
stdout/stderr when proposing a source edit.  S4 is the single-turn
deployment regime, in which the selected source tree is frozen and
invoked through \texttt{solver.main.run(ctx)}.  Validation produces the
internal selection numbers; VDM informs mutation but is withheld at the
S4 deployment surface and never enters acceptance.  The external S4
replay is the deployment evaluation.

\paragraph{Notation and selector inputs.}
Let $\mathfrak{L}_{\text{py}}$ be the universe of admissible Python solver
repositories and $\mathcal{L}\in\mathfrak{L}_{\text{py}}$ a candidate
library.  The seven CaP-X task families are
$\mathcal{T}=\{\text{cube\_lifting},\text{cube\_stack},
  \text{cube\_restack},\text{nut\_assembly},\text{spill\_wipe},
  \text{two\_arm\_lift},\text{two\_arm\_handover}\}$, and a trial is
$\tau=(t,k)$ with $t\in\mathcal{T}$.  The M4 train and validation splits
each contain $7\times 10{=}70$ trials; the CaP-Bench headline uses
$7\times 100{=}700$ trials, not $700$ distinct tasks.
The M4 train and validation splits draw the same per-task trial IDs
$\{1,\dots,10\}$, so RHO never sees the remaining $90$ trial IDs per
task; the S4 headline scores the frozen solver on the full
$7\times 100$ set, of which $630/700$ trials are held out.  The evaluator
returns trajectory $E(\mathcal{L},\tau)$, scalar shaped reward
$r_\tau(\mathcal{L})=R(E(\mathcal{L},\tau))\in[0,1]$, task-success threshold
$\theta_\star=1.0$, and ASI $s_\tau$ containing task completion, optional
M4-only visual-difference summary $s_\tau^{\mathrm{vdm}}$, judge axes
$j_\tau\in[0,1]^5$, and an integrity flag.  The coding agent may extend
$s_\tau$ with trajectory-derived ASI it authors itself.

\begin{figure}[tbp]
  \centering
  \includegraphics[width=\linewidth]{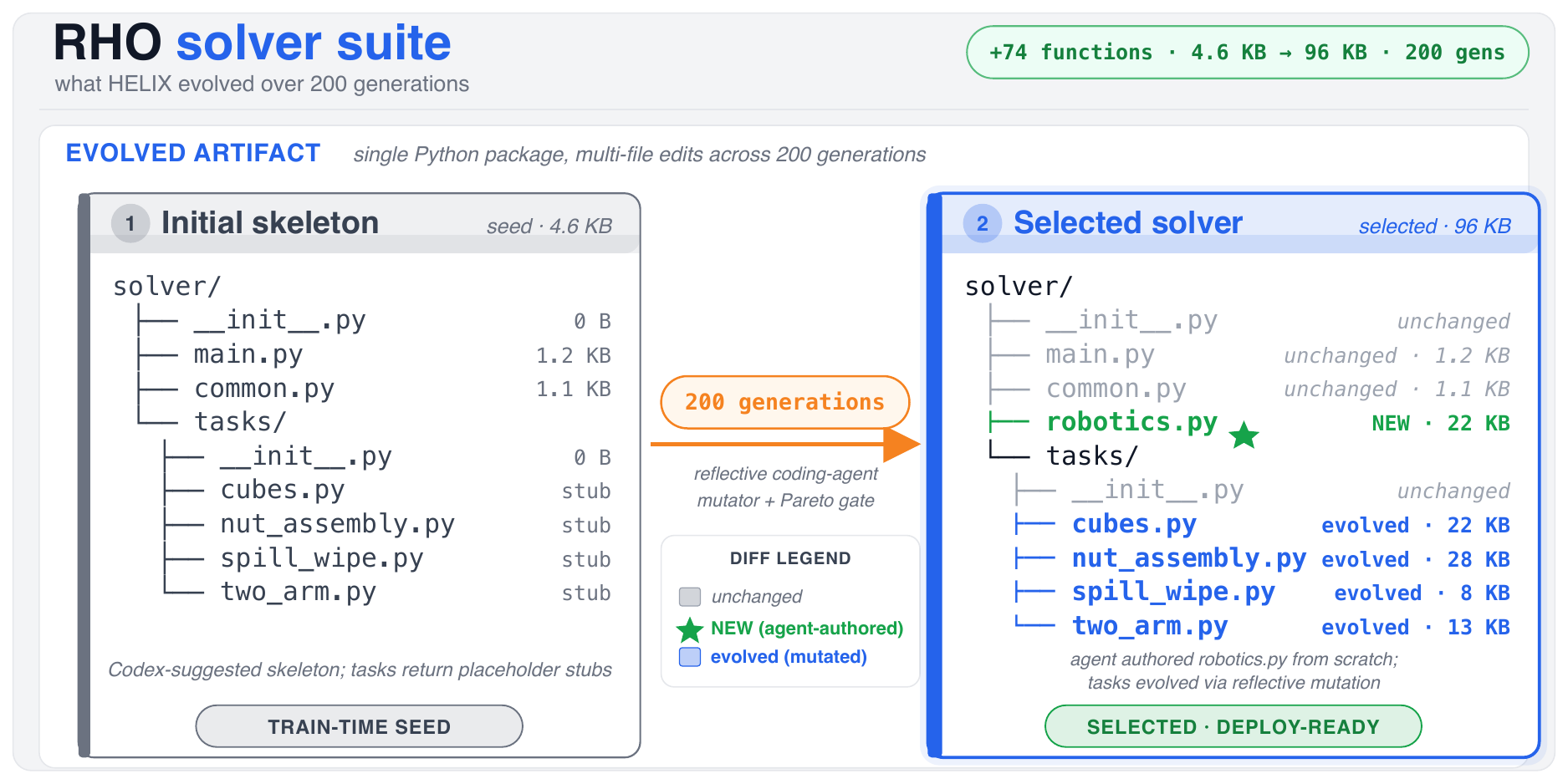}
  \caption{Deployed solver-suite interface: \texttt{run(ctx)}, task routines, and shared helpers; all benchmark interaction goes through \texttt{ctx}.}
  \label{fig:solver-suite}
\end{figure}

\paragraph{Deployed artifact inventory and API accounting.}
The selected package \texttt{solver/} is a multi-file Python package
(\Cref{fig:solver-suite}) containing \texttt{main.py} (task router and
\texttt{run(ctx)} entry
point), \texttt{common.py} and \texttt{robotics.py} (shared helpers
for perception, grasp planning, motion, and placement), and a
\texttt{tasks/} subpackage with per-task routines.  The shared
helpers include \texttt{\_observe}, \texttt{\_segment\_object},
\texttt{\_grasp\_from\_mask}, \texttt{\_approach\_grasp\_lift},
\texttt{\_place\_on\_target}, and \texttt{\_normalize\_prompt}.
\texttt{run(ctx)} parses the language instruction, takes one
observation through \texttt{\_observe}, dispatches to the relevant
task routine, and returns a status dictionary.  The implementation
has no solver-owned external services, communicating only through stdout and \texttt{ctx}.

The import audit finds only the stdlib and \texttt{numpy};
\texttt{openai}, \texttt{anthropic}, \texttt{google.genai},
\texttt{litellm}, \texttt{requests}, \texttt{httpx}, \texttt{urllib},
\texttt{socket}, \texttt{subprocess}, \texttt{capx}, \texttt{sidecar}, and
\texttt{tools.capx\_tool} are absent.  The source audit also finds no
solver-owned visual-debate rollout, runtime skill-retrieval index,
embedding store, parallel LLM voting layer, or LLM HTTP/CLI clients.

The S4 runtime API supplies \texttt{get\_observation}, SAM\,3 text and
point segmentation~\citep{carion2025sam3}, Molmo point
prompting~\citep{molmo}, Contact-GraspNet grasps,
oriented bounding boxes, IK, joint motion, gripper commands, and bimanual
variants.
\texttt{get\_observation()} returns RGB, depth, camera intrinsics, and
camera-to-world pose under \texttt{obs["robot0\_robotview"]}; missing
fields produce \texttt{\{"status": "no\_observation"\}}.  Contact-GraspNet
candidate poses and scores are returned in the camera frame and composed
with the camera pose before IK.  Quaternions are WXYZ throughout.
Training-distilled helpers for object features, grasp choice, pick/lift,
placement, WXYZ conversion, OBB construction, and finite-state task
branches live as Python functions in the package; there is no runtime
retrieval surface.

\paragraph{Reproducibility manifest.}
The HELIX loop uses GEPA-style strict-improvement
acceptance~\citep{agrawal2026gepa}, \texttt{frontier\_type="instance"},
\texttt{minibatch\_size=2}, a stratified sampler,
\texttt{perfect\_score\_threshold=1.0}, and disabled merge/crossover in
the reported run.  The final state records \texttt{generation=200},
\texttt{mutation\_counter=200}, \texttt{merge\_counter=0}, $87$ retained
frontier candidates, and $70$ active validation keys.

\paragraph{Evaluator contract and reflection.}
The evaluator reads HELIX's \texttt{helix\_batch.json}, preserves its
order, resolves each id to a native CaP-X task/trial, runs
\texttt{solver.main.run(ctx)}, and emits one positional line:
\begin{verbatim}
HELIX_RESULT=[[score_0, side_info_0], [score_1, side_info_1], ...]
\end{verbatim}
HELIX zips this payload with the batch ids.  The scalar score is the
environment-provided shaped final reward.  Judge feedback, VDM summaries,
stdout/stderr, and \texttt{feedback/last\_run/} files feed future mutation
prompts as diagnostic side information; they are not inputs to the frozen
solver at deployment time.

\begin{figure}[tbp]
  \centering
  \includegraphics[width=\linewidth]{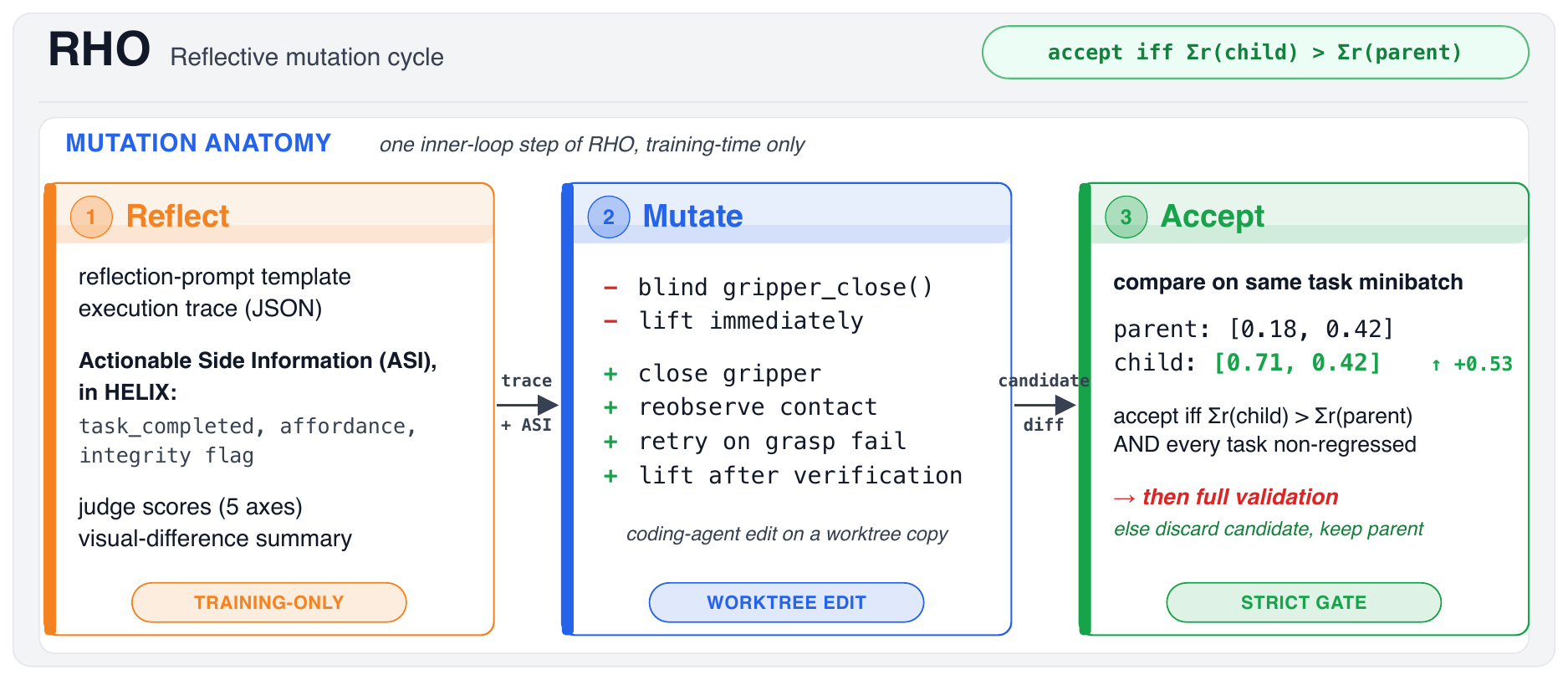}
  \caption{Mutation anatomy: diagnostics and reflection prompt a bounded coding-agent edit; the child repository is accepted only after paired minibatch scoring and validation promotion.}
  \label{fig:mutation-anatomy}
\end{figure}

The mutation operator is a bounded autonomous coding-agent session over a
copied parent worktree (\Cref{fig:mutation-anatomy}).  The reported run uses
$b=\texttt{codex}$ on
\texttt{gpt-5.5} with \texttt{model\_reasoning\_effort=xhigh} and
$K_{\text{turns}}{=}150$, invoked as \texttt{codex exec --json};
HELIX supports several backends but cross-backend evaluation across
$\mathcal{B}=\{\texttt{claude},\texttt{codex},\texttt{cursor},
  \texttt{gemini},\texttt{opencode}\}$ is outside the reported run.  The mutation
session
edits only the package surface and terminates when the agent stops
issuing tool calls.

The reflection prompt contains task instructions, the parent's current
score vector, the machine payload
\texttt{HELIX\_RESULT=[[$r_{\tau_1}$,$s_{\tau_1}$],...]}, per-trial
diagnostics, the M4-only $s_\tau^{\mathrm{vdm}}$, judge axes, integrity
flags, evaluator notes, and fallback stdout/stderr.  Timing and token-cost
streams are withheld from the reflection prompt so acceptance cannot
reward skipping slow robot operations.  The scalar reward remains
$r_\tau$; judge axes and VDM summaries are diagnostics only.

\subsubsection{Run State, Lineage, and Audit Trail}
\label{app:run-state-lineage}

HELIX treats each coding-agent proposal as a persistent repository
snapshot (\Cref{fig:lineage}).  In the reported root, \texttt{.helix/} retains
candidate
worktrees, accepted evaluation snapshots, \texttt{state.json},
\texttt{lineage.json}, the evaluator manifest, and the controller log.
The re-audit records a single-phase structure: one
\texttt{state.json}, one \texttt{lineage.json}, one \texttt{[evolution]}
configuration block, and no per-phase directories.  \texttt{lineage.json}
records $166$ nodes from \texttt{g0-s0} through \texttt{g182-s182};
\texttt{evaluations/} retains $87$ full validation files;
\texttt{attempts/} retains $77$ rejected train-minibatch attempts; and
\texttt{skips/} retains $28$ skip records.

The g1-s1 single-mutation child of the stub seed g0-s0, used as the
one-shot Codex baseline in \S\ref{sec:exp-ablation-failure}, scores
a mean reward of $0.106741$ and $0.0\%$ ($0/70$) task success on the full
70-trial validation set.  The selected deployed candidate
\texttt{g134-s134} scores $0.730982$ mean reward and covers $49/70$
active instance-frontier keys ($70.0\%$), matching the external
S4 headline of $490/700$ reported in \S\ref{sec:results}.
\texttt{g97-s97} appears only as a rejected train-minibatch
attempt under \texttt{attempts/}: non-canonical historical lineage,
not the reported RHO selection.  A resumed controller reloads the
frontier, verifies the stored \texttt{config\_hash}, continues mutation
and merge counters without ID collisions, and treats worktrees without
evaluation snapshots as incomplete proposals.

\subsubsection{Configuration, Budgets, and Judge Prompt}
\label{sec:appendix-config}

\begin{figure}[tbp]
  \centering
  \includegraphics[width=\linewidth]{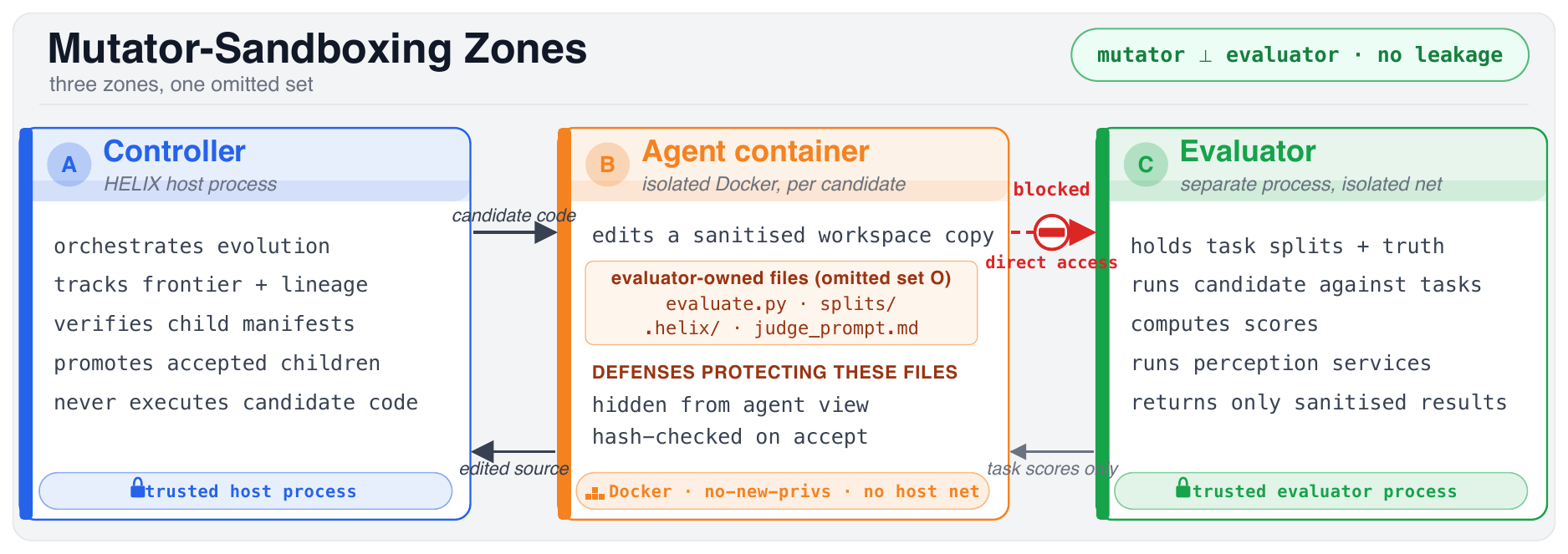}
  \caption{Mutator-sandboxing zones: the agent workspace, evaluator-owned files, candidate execution container, and blocked network edge are separated before acceptance.}
  \label{fig:sandbox-zones}
\end{figure}

The top-level run objective maximizes environment-provided shaped reward
for simulated Franka manipulation tasks while preserving
\texttt{solver.main.run(ctx)} and the documented runtime API.  The
\texttt{[seedless]} block uses
\texttt{train\_path="splits/instance\_ids"} and
\texttt{val\_path="splits/instance\_ids"}, covering seven task groups
with ten native trial labels each.  The evaluator command is:
\begin{verbatim}
python evaluate.py
\end{verbatim}
The evaluator uses \texttt{score\_parser='helix\_result'}, protected files
\texttt{evaluate.py}, \texttt{API\_REFERENCE.md},
\texttt{judge\_prompt.md}, \texttt{README.md}, \texttt{splits},
\texttt{helix.toml}, \texttt{splits/train.yaml}, and
\texttt{splits/val.yaml}, and one judge call per candidate.
\Cref{fig:sandbox-zones} shows the mutator sandboxing
enforced around these evaluator-owned files.

\paragraph{Training budgets.}
The final persisted state reached \texttt{generation=200} and
\texttt{mutation\_counter=200}.  The retained frontier contains $87$
candidates.  Lineage over retained candidates comprises $86$
mutation operations plus the seed.  The selected candidate
\texttt{g134-s134} has internal best score $0.730982$ mean shaped reward
and $49/70$ trials solved on the 70-instance validation set, and its
accepted external S4 replay reports $490/700$.  The run wall-clock is
$79.4$ hours from \texttt{.helix/config.toml} creation time to the final
\texttt{.helix/state.json} write, on a dual-socket Intel Xeon Platinum
8570 host with two NVIDIA B200 GPUs.  The final state reports
\texttt{budget.evaluations=6814}, \texttt{budget.input\_tokens=625276104},
\texttt{budget.cached\_input\_tokens=88851968}, and
\texttt{budget.output\_tokens=3144685}.  The sum over $87$ retained
backend-result files is $386{,}757{,}173$ input tokens,
$373{,}486{,}336$ cached input tokens, and $1{,}757{,}060$ output tokens.

\paragraph{Judge axes.}
\label{app:judge-axes}
Each accepted candidate is scored by an OpenAI \texttt{gpt-5.1} judge
on five capability axes plus an integrity flag, returned as a
strict-JSON object alongside per-axis evidence citations (code line
ranges and trace excerpts).  Each axis score is a float in $[0.0, 1.0]$
with anchors at $0.0$ (no executed evidence), $0.25$ (minimal one-off
evidence), $0.5$ (clear evidence in one task stage), $0.75$ (multiple
task-relevant instances), and $1.0$ (strong repeated evidence across
stages); high task reward alone is explicitly not evidence of any
capability.  The five axes are
\emph{state\_uncertainty\_management} (re-observation, hypothesis
comparison, viewpoint adjustment, delayed irreversible action, belief
update), \emph{robustness\_contingency\_handling} (branching on
perception failure, retry with modified parameters, primitive or pose
fallbacks, precondition and outcome checks, safe degradation),
\emph{strategic\_coherence} (locate-approach-grasp-transport-place
decomposition, physical-prerequisite ordering, intermediate-result
propagation), \emph{physical\_affordance\_grounding} (grasp poses
based on object geometry, placement on support surfaces, reachability
and collision checks), and \emph{adaptive\_exploration\_search}
(multiple candidate objects or poses, search-space expansion after
inconclusive observation, strategy change based on intermediate
results).  The judge additionally returns
\texttt{integrity\_hardcoding\_flag} with severity in
$\{\textsc{none}, \textsc{minor}, \textsc{major}, \textsc{severe}\}$;
$\textsc{major}$ caps all five axis scores at $0.25$ and
$\textsc{severe}$ sets them to $0.0$.  None of the judge channels
enter the acceptance rule of Eq.~(\ref{eq:accept}) in the main text;
they are reflection-only side information used to drive the next
mutation prompt.  The verbatim rubric (\texttt{judge\_prompt.md}) is
released in the supplementary submission artifact.

\subsubsection{RAI Portability Protocol}
\label{app:method-rai-portability}

RAI is RobotecAI's open-source ROS\,2-oriented embodied-agent
framework~\citep{rai}, evaluated here on an O3DE manipulation
benchmark.  RHO evolves the
prompts and tool-edit Python module that adapt RAI's tools for that
benchmark; it does not evolve RAI's deployment loop, ROS\,2 connector
layer, or inference-time driver.

The benchmark uses class-stratified train/test splits: $21/21$ on the easy
pool and $22/21$ on the hard pool.  RHO selects the best-by-train
candidate after ten generations per cell, then evaluates the held-out test
half; the easy headline averages $n{=}21$ held-out test tasks over three
runs.  The easy O3DE multi-file cell lifts the seed from \textbf{0.683} to
\textbf{0.889} ($+0.206$, $p<0.001$), and the hard multi-file cell lifts the
seed from \textbf{0.235} to \textbf{0.443} ($+0.208$, $p<0.001$).

At deployment, HELIX's training-time mutator is not invoked, but RAI's own
deployment driver remains on the inference path.
The CaP-X no-solver-owned code-generation invariant remains a
property of the CaP-X solver instantiation; RAI tests portability of the
train-time edit-and-freeze recipe to a ROS\,2 embodied-agent substrate.

The RAI replication ran on a single AMD Ryzen AI Max+ 395 workstation with
$128$~GB unified memory, with the RAI agent, ROS\,2 stack, and O3DE
simulator co-resident on-device.  RAI deployment numbers are reported in
Section~\ref{sec:rai-o3de}.

\subsubsection{Mutator reflection-prompt template}
\label{app:prompt-template}

The reflection prompt issued to the coding agent at each mutation step
is assembled by the HELIX runtime from (i) a fixed task-instruction
preamble, (ii) the parent candidate's full evaluator output, (iii) the
\texttt{background} field of \texttt{helix.toml} (the domain-context
scaffold), and (iv) a fixed task request.  The skeleton below was captured
verbatim from a representative mutation step in the canonical Skeleton
run; placeholder fields are bracketed.  The Diagnostics block is
dynamic: its field set is determined by whatever per-instance
\texttt{side\_info} keys the user's evaluator emits, rendered in
sorted-key order, so the fields shown below reflect the canonical
run's evaluator and may differ for other evaluators.

\begin{Verbatim}[fontsize=\small, frame=single, framesep=4pt,
    breaklines=true]
  Task instructions:
  - Work directly on the requested code changes using the workspace files.
  - Do not request confirmation or clarification; choose a reasonable approach
  and continue.
  - If one approach fails, try an alternative and keep progressing.
  - Use available tools to inspect, edit, and validate changes.

  ## Objective
  Maximize the environment-provided shaped reward for simulated Franka
  manipulation tasks. When evaluation diagnostics are present, prioritize
  the concrete examples shown there and preserve behavior that already
  scores well. Maintain the solver.main.run(ctx) entry point and use only
  the documented runtime API.

  ## Current Evaluation Scores
  success: <parent_score>

  ## Diagnostics
  ### Example <task_family>__<trial_id>
  #### diagnostic_focus
  <lowest_reward_failed_for_vdm | highest_reward_succeeded | mixed>
  #### evaluation_diagnostics
  ##### capability_scores
  ###### state_uncertainty_management        <float in [0, 1]>
  ###### robustness_contingency_handling     <float in [0, 1]>
  ###### strategic_coherence                 <float in [0, 1]>
  ###### physical_affordance_grounding       <float in [0, 1]>
  ###### adaptive_exploration_search         <float in [0, 1]>
  ##### task_failure_hypotheses              <up to 3 items>
  ##### minimal_next_experiments             <up to 3 items>
  ##### preserve_existing_behavior           <up to 3 items>
  ##### api_misuse_observations              <0+ items, file:line cited>
  ##### integrity_warning
  ###### severity                            <none | minor | major>
  ###### evidence                            <up to 3 items, file:line cited>
  ###### score_caps_applied                  <none | applied cap>
  #### task_completed                        <True | False>
  #### top_k_best_example_history            <top-K accepted ancestors by score>
  #### visual_difference_summary             <vdm free-text prev/current narration>
    [... one block per sampled instance; some carry only task_completed
      + top_k_best_example_history ...]

  ## Background / Context
    [Domain-context scaffold from helix.toml: API_REFERENCE.md pointer,
      documented runtime primitives, closed-loop FSM preference, integrity
      rules, observable verification gates.]

  ## Your Task
  Analyse the evaluation results above and improve the code to better
  achieve the objective.  Make targeted, meaningful changes.  You may
  read, edit, create, or delete files as needed.

  ## Turn Budget
  You have a 150-turn limit for this task, where turns refer to how many
  tool calls or interactions you can make.  Plan your work accordingly:
  prioritize the highest-impact changes first and be efficient with your
  tool usage.
\end{Verbatim}

The capability-scores block carries the LLM-judge axes defined in
Appendix~\ref{app:judge-axes}; the verbatim judge prompt
(\texttt{judge\_prompt.md}) is released in the supplementary
submission artifact.

\subsubsection{Seed configurations}
\label{app:seed}

HELIX requires a generation-0 seed repository that exposes the
benchmark's invocation signature: we do not tell the agent which
benchmark it is solving, so, without the signature, the mutator would have
to guess how its program should be invoked.  We compare two configurations.

\paragraph{Skeleton seed (canonical Robosuite run).}
A coding agent pre-populates a multi-file Python package with typed
stubs for the task router, per-task solvers, and shared helpers
(initial source-line counts in parentheses):
\begin{Verbatim}[fontsize=\small, frame=single, framesep=4pt]
  solver/
  __init__.py        (1L)   from solver.main import run
  main.py            (27L)  prompt-keyword router -> 6 per-task solvers
  common.py          (30L)  prompt_text, observe, first_camera, note_status
  tasks/
  __init__.py      (0L)
  cubes.py         (stub) solve_cube_lifting, solve_cube_relation
  nut_assembly.py  (stub) solve_nut_assembly
  spill_wipe.py    (stub) solve_spill_wipe
  two_arm.py       (stub) solve_two_arm_lift, solve_two_arm_handover
\end{Verbatim}
The 200-generation canonical Codex GPT-5.5 Multi-file run selects
\texttt{g134-s134} at \textbf{70.0\%} (490/700) on CaP-Bench Robosuite
S4 external evaluation.

\paragraph{Stub-main seed.}
A minimal alternative supplies only the entry point and no task
scaffolding:
\begin{Verbatim}[fontsize=\small, frame=single, framesep=4pt]
  solver/
  __init__.py        (1L)   from solver.main import run
  main.py            (6L)   def run(ctx): prints + returns {"status": "stub"}
\end{Verbatim}
Under the stub-main seed, the loop reaches generation 250 and collapses
everything into a single monolithic \texttt{main} function: under the
same task instructions and tool surface as the Skeleton run sees, the
mutator rarely refactors this structure, and the run terminates without
a candidate matching the Skeleton lineage's selected solver.  This is the
failure mode summarized
in Section~\ref{sec:method-rap}: with no task-family scaffold to
modify, each mutation collapses into a single-function rewrite that
the next mutation must again repair.

\clearpage
\subsection{Experiments Appendix}
\label{sec:experiments-appendix}

\subsubsection{Benchmark protocols}

\textbf{CaP-Bench} is the seven-task Robosuite slice of
CaP-X~\citep{fu2026capx}: $100$ trials per task, $700$ trials total,
evaluated under RHO's held-out S4 surface.
\textbf{LIBERO-PRO}~\citep{zhou2025liberopro} is a perturbation
benchmark of six conditions, defined by
(object/goal/spatial)$\times$(swap/task), each with $10$ tasks and
$50$ trials.

\subsubsection{Deployment audit and S4 contract}

RHO training runs in the M4 tier: multi-turn execution, VDM access,
and a coding-agent loop that can inspect traces and mutate code.  All
reported CaP-Bench and LIBERO-PRO deployment numbers are S4 single-turn
runs with VDM disabled and no solver-owned LLM code-generation client.  For the canonical
$490/700$ CaP-X result, the S4 protocol is recorded in the provenance
manifest.  The VDM=0 / single-turn evaluation is the CaP-X S4 contract.
A source-level audit finds no deployed-solver imports of
\texttt{openai}, \texttt{anthropic}, \texttt{google.genai},
\texttt{litellm}, \texttt{requests}, \texttt{httpx}, or the CaP-X
sidecar transport.

\subsubsection{Comparator reconstruction}

On CaP-Bench, the comparator is Agent0 in the +3M configuration
($68.29\%$ on Fig.~8; $9$ candidate code-generation queries, an
ensemble synthesis call, and a VDM call per turn; \citep{fu2026capx}).
On LIBERO-PRO, the comparators are the VLA and CaP-Agent0 rows from
CaP-X and the LIBERO-PRO paper: OpenVLA scores $0.00$ on all six
aggregate cells, and CaP-Agent0's $18.17\%$ ($545/3000$) is a derived
aggregate over six equally weighted swap/task cells, not a published
headline~\citep{fu2026capx,zhou2025liberopro}.
The LIBERO-PRO comparison excludes LIBERO-Plus
aggregates~\citep{fei2026liberoplus}.
Among LIBERO-trained baselines, $\pi_{0.5}$~\citep{black2025pi05}
concentrates almost all of its success on the swap cells ($0.38$
goal-swap, $0.17$ object-swap, $0.20$ spatial-swap) and collapses to
near-zero on the task cells ($0.00$ goal-task, $0.01$ object-task,
$0.01$ spatial-task)~\citep{zhou2025liberopro}.

\subsubsection{Runtime and latency ledger}
\label{sec:appendix-latency}

HELIX deployment wall-clock time averages \textbf{$189.1$ s/trial} (median
$161.5$, p95 $438.5$) across the $700$ CaP-Bench trials ($7$ tasks
$\times$ $100$ trials), measured on a dual-socket Intel Xeon Platinum
8570 host with two NVIDIA B200 GPUs.  This is full per-episode
wall-clock time: it includes simulator, robot, perception, and logging, not
solver-owned code-generation overhead alone, and Agent0 would share
this floor.
Table~\ref{tab:per-task-latency} resolves the aggregate into per-task
means: \texttt{cube\_stack} anchors the floor at $44.8$\,s, while
\texttt{nut\_assembly} ($409.2$\,s mean) drives the p95 tail almost on
its own.  The source audit finds no runtime retrieval index, embedding
store, or skill-selection API in the deployed solver: the zero-call
result is a repository-as-library deployment, not a retrieval-backed
runtime agent.
% tab-per-task-latency.tex --- Per-task HELIX deploy wall-clock.

\begin{table}[t]
\centering
\small
\caption{Per-task deploy wall-clock for HELIX on CaP-Bench. Per-task
means span a $9.1\times$ range: $\texttt{cube\_stack}$ anchors the floor
at $44.8$\,s, while $\texttt{nut\_assembly}$ at $409.2$\,s drives the
aggregate p95 ($438.5$\,s) almost alone, its median ($407.9$\,s) already
exceeding that aggregate p95. Across all 700 trials the mean is
$189.1$\,s and the median $161.5$\,s. Each per-trial figure is full
per-episode wall-clock (environment reset, perception, simulation,
logging) for the selected $\texttt{g134-s134}$ solver on the S4 oracle
path; no LLM code-generation call is issued inside the trial
(host-provided perception primitives such as SAM\,3 and Molmo still run).}
\label{tab:per-task-latency}
\begin{tabular}{lrrrr}
\toprule
Task & mean (s) & median (s) & p95 (s) & n \\
\midrule
\texttt{cube\_lifting}       & $131.3$           & $104.3$           & $282.6$           & $100$ \\
\texttt{cube\_stack}         & $\phantom{0}44.8$ & $\phantom{0}46.0$ & $\phantom{0}60.5$ & $100$ \\
\texttt{cube\_restack}       & $106.7$           & $\phantom{0}80.1$ & $237.5$           & $100$ \\
\texttt{nut\_assembly}       & $409.2$           & $407.9$           & $514.7$  & $100$ \\
\texttt{spill\_wipe}         & $309.5$           & $309.5$           & $385.2$           & $100$ \\
\texttt{two\_arm\_lift}      & $168.1$           & $173.1$           & $234.1$           & $100$ \\
\texttt{two\_arm\_handover}  & $153.8$           & $171.1$           & $186.1$           & $100$ \\
\midrule
\textbf{All 7 tasks}         & $\mathbf{189.1}$  & $\mathbf{161.5}$  & $\mathbf{438.5}$  & $\mathbf{700}$ \\
\bottomrule
\end{tabular}
\end{table}

\begin{figure}[tbp]
  \centering
  \includegraphics[width=\textwidth]{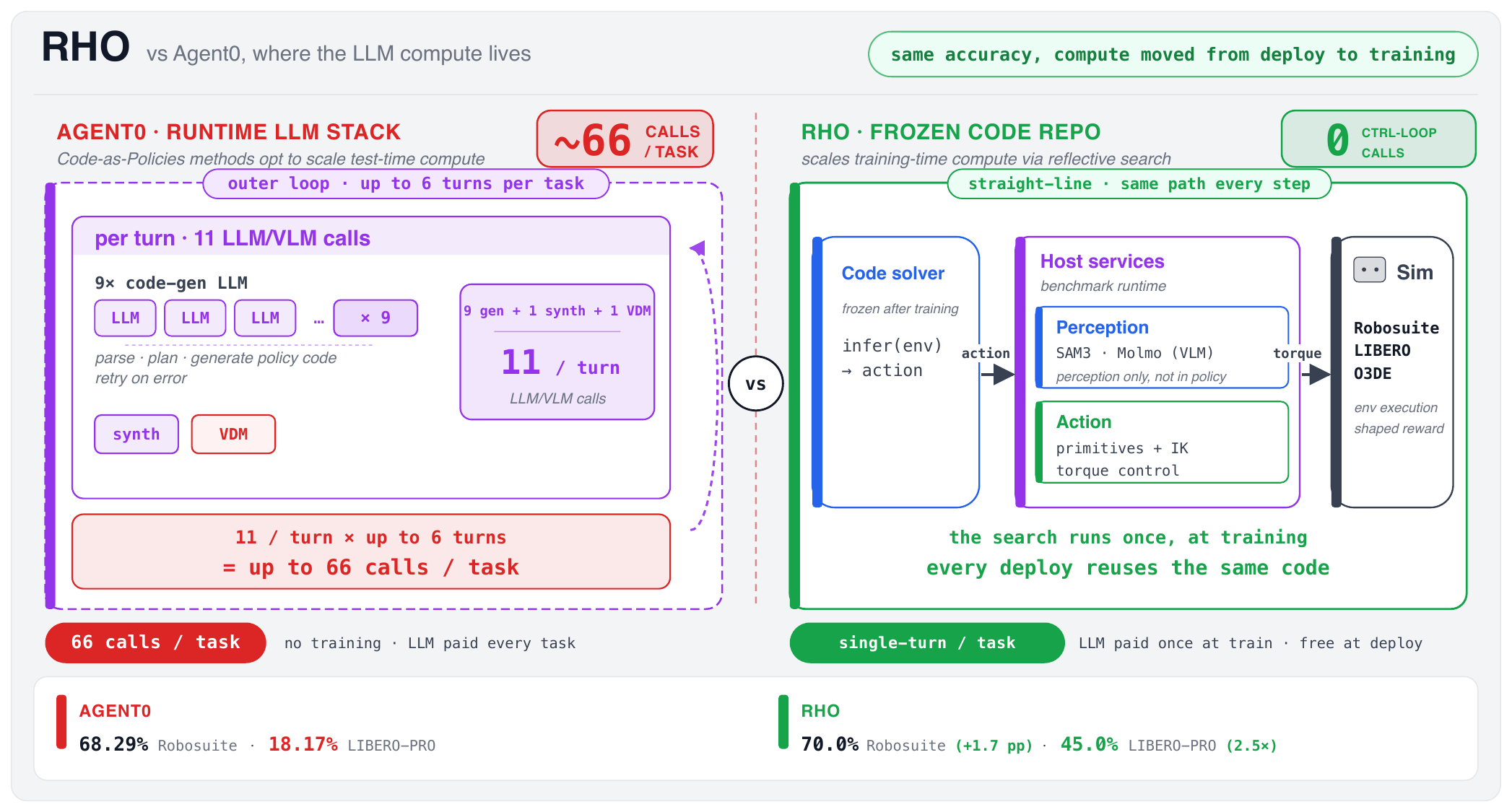}
  \caption{Where the LLM code-generation compute lives.  Agent0 issues up to sixty-six
    LLM calls per trial at deploy time (nine code-generation
    queries, one ensemble synthesis call, and one VDM call per turn, over up
    to six turns); RHO issues no LLM code-generation calls in the control loop,
    having amortised that compute at train time.  Both reach comparable
    CaP-Bench accuracy ($68.29\%$ Agent0 vs $\mathbf{70.0\%}$ RHO) on
    the same host-provided perception and action primitives (SAM\,3, Molmo,
    Contact-GraspNet).}
  \label{fig:helix_vs_agent0}
\end{figure}

The structural call-count contrast does not support a defensible dollar
inequality: CaP-X discloses none of Agent0's per-call token mix, retry
distribution, cache behavior, or VDM pricing.
\Cref{fig:helix_vs_agent0} maps where that compute lives: Agent0 issues
up to $66$ deploy-time LLM calls per trial, while RHO issues no
LLM code-generation calls in the control loop, at comparable
CaP-Bench accuracy.

\subsubsection{Backend selected-run sensitivity}
\label{app:backend-ablation}

\paragraph{Mutator harness configurations.}
Two coding-agent harnesses drive the backends in this
paper.  Codex backends use OpenAI's Codex CLI with \texttt{gpt-5.5}
at reasoning effort \texttt{xhigh}.  Anthropic backends use
Anthropic's Claude Code with \texttt{claude-sonnet-4-6} (Sonnet 4.6
rows, including Sonnet 4.6 Behavior Tree) and \texttt{claude-opus-4-7} (Opus 4.7
rows, including Opus).  The Qwen rows in
Tables~\ref{tab:no-loop-floor} and~\ref{tab:structural-prior} use
Claude Code redirected via \texttt{ANTHROPIC\_BASE\_URL} to a local
vLLM endpoint serving \texttt{Qwen3.6-27B} (tensor parallel 2,
context window $262{,}144$, \texttt{qwen3\_coder} tool-call parser).
Both harnesses expose comparable tool surfaces (file edits, shell
execution, environment interaction).  HELIX records per-mutation
token usage uniformly across them.

\paragraph{Selected-run headline spread.}
Across accepted single CaP-Bench S4 runs, the selected-run headline
varies substantially: the reported Codex-backed run reaches
$70.0\%$, an alternative Codex configuration reaches $63.9\%$, Opus
reaches $61.9\%$, Qwen3.6-27B reaches $35.0\%$, and Sonnet 4.6 Behavior Tree reaches
$33.9\%$, a $36.1$\,pp spread between the highest and lowest accepted
backend.  The entries are selected-run comparisons, not a multi-seed
backend ranking.

\subsubsection{Training-time REaaS tool use}
\label{app:env-tool-use}

The reported run records $1{,}776$ calls to the robot
environment-as-a-service (REaaS), evidence that the mutation agents used the
environment tools throughout the search rather than emitting code zero-shot.

\paragraph{The REaaS.} During a mutation the coding agent may
exercise five REaaS endpoints.
\texttt{reset} starts a task trial and returns the task prompt, the allowed
primitive calls, and a summarized initial observation; \texttt{observe} returns
the current observation; \texttt{call} invokes a single perception or action
primitive from the allowed set (for example, text-prompted segmentation or grasp
planning); \texttt{run} executes the whole candidate solver end-to-end; and
\texttt{vdm} poses a natural-language question to a vision-language model that
performs a visual difference between the current observation and the preceding
timestep, reporting what changed. The first is per-trial setup; the rest are how
the agent
probes and reads the live environment.

\paragraph{Environment engagement, not tool volume, tracks deployed quality.}
Four of the five service channels in \Cref{tab:cross-run-budget} fall
monotonically down the success column. From the $70.0\%$ solver to the $18.1\%$
one, the mutator runs candidates $576\!\to\!486\!\to\!405\!\to\!61\!\to\!11\!\to\!5$
times, invokes control and perception primitives
$293\!\to\!247\!\to\!182\!\to\!32\!\to\!3\!\to\!0$ times, queries the
visual-difference module $166\!\to\!86\!\to\!84\!\to\!8\!\to\!0\!\to\!0$ times, and
re-observes the scene $43\!\to\!31\!\to\!31\!\to\!8\!\to\!0\!\to\!0$ times. Only
\texttt{reset}, the per-trial environment setup, fails to track score.

\paragraph{Call volume does not track quality; grounded depth does.} The FSM run issues more
total service calls than the Behavior
Tree run ($62$ versus $28$) yet deploys the weakest solver, because $57$ of its
$62$ calls are resets: it resets the simulator $57$ times, runs a candidate $5$
times, and never once queries the visual-difference module or re-observes a scene.
A search that resets but never observes cannot distinguish a working candidate
from a broken one.

\paragraph{Grounded search finds the verification that works.} A visual-difference
query lets the mutator see the gap between what a candidate intended and what it
achieved, and a re-observation confirms the world changed, so a search that
exercises both retains correctness mechanisms a blind search cannot score. The
dimension-wise artifact analysis shows why: the feedback-heavy winner is the
only solver whose verification distrusts its own perception, cross-checking
text-prompt against point-prompt detections and flagging a stale, unchanged scene.
It is also the only one whose terminal check triggers a corrective re-execution
rather than logging a status string. The Qwen FSM solver carries the \emph{densest}
verification coverage of the four artifacts, yet it never grounds that
verification against visual difference during search and narrates failure
without repairing it. More
verification code does not help; grounding the search so the mutator finds
verification that \emph{works} does.

\paragraph{Reaching for visual feedback is a frontier-model behavior.} The same
channels separate the backends by mutator scale, not only by score. Only the two
frontier mutators ground their search at volume: Codex (\texttt{gpt-5.5}) issues
$166$ visual-difference and $43$ re-observe queries in the multi-file run and $86$
and $31$ in the single-file run, and Claude Code (\texttt{Opus~4.7}) issues $84$ and
$31$. The smaller
Sonnet~4.6 issues none of either, and open-source Qwen3.6-27B issues only $8$
visual-difference and $8$ re-observe queries in free-form search and none under
the FSM constraint, one to two orders of magnitude below the frontier mutators.
Mutator scale, depth of grounded search, and final success co-vary across these
single selected runs, so we do not claim one drives another; the consistent
pattern is that the frontier mutators which ground their search most deeply are
the ones that ship the robust $70.0\%/63.9\%/61.9\%$ solvers.

\subsubsection{LIBERO-PRO condition cells}
\label{sec:exp-libero-pro-cells}

RHO on LIBERO reaches \textbf{$45.0\%$} on LIBERO-PRO
($1{,}351/3{,}000$) with no LIBERO gradient update.  The selected
candidate is \texttt{g32-s32} at generation 63 of the
reflective-evolution lineage (an initial 31-generation lane through
\texttt{g31-s31} plus a 32-generation warm-restart lane re-seeded
from \texttt{g31-s31}).
Its six condition scores are $50.6\%$ goal-swap, $55.6\%$ goal-task,
$12.2\%$ object-swap, $37.2\%$ object-task, $53.0\%$ spatial-swap, and
$61.6\%$ spatial-task.
Tables~\ref{tab:libero-object-pertask}, \ref{tab:libero-goal-pertask},
and~\ref{tab:libero-spatial-pertask}
give the per-task breakdown of these cells against the CaP-X baselines
(OpenVLA, $\pi_0$, $\pi_{0.5}$, CaP-Agent0), reproducing CaP-X
Tables~6--8 with an added RHO column.  Throughout, Pos is the
initial-position perturbation (our \texttt{swap} cell) and Task is the
instruction perturbation (our \texttt{task} cell); values are the
fraction of $50$ trials solved.

\begin{table}[t]
  \centering
  \footnotesize
  \setlength{\tabcolsep}{4pt}
  \caption{Per-task success (fraction of $50$ trials) on
    \texttt{libero-object}; baselines from
    CaP-X~\citep{fu2026capx,zhou2025liberopro}.  The two VLAs transfer
    almost nothing ($\approx 0$ on every task), and $\pi_{0.5}$ is bimodal:
    near-perfect on a few memorized objects under position swaps
    (bbq\_sauce $1.00$, butter $0.54$) yet $\approx 0$ elsewhere and on all
    instruction swaps, so neither baseline generalizes.  The two
    code-as-policies agents instead spread partial success across objects.
    RHO is the most language-robust (Task $0.37$ vs Agent0's $0.18$,
    as its perception re-grounds the renamed target), but object-swap is its
    one losing cell (Pos $0.12$ vs $0.22$): the mutator-added, hand-fitted
    grasp schedule fails to transfer when the object is physically relocated
    (analyzed below).  Best column average is in bold;
    \textit{Place(obj, basket)}: pick up \textit{obj} and place in the
    basket.}
  \label{tab:libero-object-pertask}
  \resizebox{\textwidth}{!}{%
    \begin{tabular}{l*{10}{c}}
      \toprule
                                        & \multicolumn{2}{c}{OpenVLA} & \multicolumn{2}{c}{$\pi_0$} & \multicolumn{2}{c}{$\pi_{0.5}$} & \multicolumn{2}{c}{CaP-Agent0} & \multicolumn{2}{c}{\textbf{RHO}}                                                          \\
      \cmidrule(lr){2-3}\cmidrule(lr){4-5}\cmidrule(lr){6-7}\cmidrule(lr){8-9}\cmidrule(lr){10-11}
      Task (symbolic form)              & Pos                         & Task                        & Pos                             & Task                           & Pos                                    & Task & Pos            & Task  & Pos   & Task           \\
      \midrule
      Place(alphabet\_soup, basket)     & 0.00                        & 0.00                        & 0.00                            & 0.00                           & 0.00                                   & 0.00 & 0.02           & 0.04  & 0.12  & 0.40           \\
      Place(bbq\_sauce, basket)         & 0.00                        & 0.00                        & 0.00                            & 0.00                           & 1.00                                   & 0.02 & 0.12           & 0.42  & 0.32  & 0.88           \\
      Place(butter, basket)             & 0.00                        & 0.00                        & 0.00                            & 0.00                           & 0.54                                   & 0.00 & 0.26           & 0.18  & 0.12  & 0.98           \\
      Place(chocolate\_pudding, basket) & 0.00                        & 0.00                        & 0.00                            & 0.00                           & 0.00                                   & 0.02 & 0.18           & 0.48  & 0.00  & 0.00           \\
      Place(cream\_cheese, basket)      & 0.00                        & 0.00                        & 0.10                            & 0.00                           & 0.00                                   & 0.00 & 0.12           & 0.06  & 0.00  & 0.00           \\
      Place(ketchup, basket)            & 0.00                        & 0.00                        & 0.00                            & 0.00                           & 0.20                                   & 0.02 & 0.32           & 0.12  & 0.00  & 0.04           \\
      Place(milk, basket)               & 0.00                        & 0.00                        & 0.00                            & 0.00                           & 0.00                                   & 0.00 & 0.38           & 0.02  & 0.18  & 0.26           \\
      Place(orange\_juice, basket)      & 0.00                        & 0.00                        & 0.00                            & 0.00                           & 0.00                                   & 0.02 & 0.30           & 0.02  & 0.34  & 0.40           \\
      Place(salad\_dressing, basket)    & 0.00                        & 0.00                        & 0.10                            & 0.00                           & 0.00                                   & 0.00 & 0.32           & 0.00  & 0.14  & 0.76           \\
      Place(tomato\_sauce, basket)      & 0.00                        & 0.00                        & 0.00                            & 0.00                           & 0.00                                   & 0.00 & 0.16           & 0.48  & 0.00  & 0.00           \\
      \midrule
      Average                           & 0.00                        & 0.00                        & 0.00                            & 0.00                           & 0.17                                   & 0.01 & \textbf{0.218} & 0.182 & 0.122 & \textbf{0.372} \\
      \bottomrule
    \end{tabular}}
\end{table}

\begin{table}[t]
  \centering
  \footnotesize
  \setlength{\tabcolsep}{4pt}
  \caption{Per-task success (fraction of $50$ trials) on
    \texttt{libero-goal}; baselines from
    CaP-X~\citep{fu2026capx,zhou2025liberopro}.  The VLAs again collapse and
    $\pi_{0.5}$ only spikes on a handful of memorized positions (a few Pos
    cells near $1.0$, Task $\approx 0$).  RHO is the only system with
    broad two-sided competence: it partially solves most goal tasks under
    both relocated objects (Pos $0.51$) and reworded goals (Task $0.56$),
    roughly doubling Agent0 ($0.26$/$0.17$) by being consistently capable
    rather than winning isolated cells.  Best column average is in bold;
    actions \textit{Open}/\textit{Put}/\textit{Push}/\textit{TurnOn} follow
    CaP-X notation.}
  \label{tab:libero-goal-pertask}
  \resizebox{\textwidth}{!}{%
    \begin{tabular}{l*{10}{c}}
      \toprule
                                      & \multicolumn{2}{c}{OpenVLA} & \multicolumn{2}{c}{$\pi_0$} & \multicolumn{2}{c}{$\pi_{0.5}$} & \multicolumn{2}{c}{CaP-Agent0} & \multicolumn{2}{c}{\textbf{RHO}}                                                          \\
      \cmidrule(lr){2-3}\cmidrule(lr){4-5}\cmidrule(lr){6-7}\cmidrule(lr){8-9}\cmidrule(lr){10-11}
      Task (symbolic form)            & Pos                         & Task                        & Pos                             & Task                           & Pos                                    & Task & Pos   & Task  & Pos            & Task           \\
      \midrule
      Open(cabinet, drawer\_mid)      & 0.00                        & 0.00                        & 0.00                            & 0.00                           & 0.00                                   & 0.04 & 0.00  & 0.00  & 0.00           & 0.00           \\
      Put(bowl, drawer\_top)          & 0.00                        & 0.00                        & 0.00                            & 0.00                           & 0.94                                   & 0.02 & 0.04  & 0.00  & 0.30           & 0.32           \\
      Push(plate, stove\_front)       & 0.00                        & 0.00                        & 0.00                            & 0.00                           & 0.00                                   & 0.00 & 0.00  & 0.10  & 0.00           & 0.00           \\
      Put(bowl, plate)                & 0.00                        & 0.00                        & 0.00                            & 0.00                           & 0.00                                   & 0.02 & 0.36  & 0.38  & 0.98           & 0.98           \\
      Put(bowl, stove)                & 0.00                        & 0.00                        & 0.00                            & 0.00                           & 0.00                                   & 0.04 & 0.22  & 0.12  & 1.00           & 0.98           \\
      Put(bowl, cabinet\_top)         & 0.00                        & 0.00                        & 0.00                            & 0.00                           & 0.00                                   & 0.02 & 0.60  & 0.04  & 0.98           & 0.98           \\
      Put(cream\_cheese, bowl)        & 0.00                        & 0.00                        & 0.00                            & 0.00                           & 0.98                                   & 0.02 & 0.04  & 0.34  & 0.10           & 0.10           \\
      Put(wine\_bottle, rack)         & 0.00                        & 0.00                        & 0.00                            & 0.00                           & 0.88                                   & 0.02 & 0.02  & 0.12  & 0.58           & 0.62           \\
      Put(wine\_bottle, cabinet\_top) & 0.00                        & 0.00                        & 0.00                            & 0.00                           & 0.98                                   & 0.02 & 0.62  & 0.40  & 0.64           & 0.64           \\
      TurnOn(stove)                   & 0.00                        & 0.00                        & 0.00                            & 0.00                           & 0.00                                   & 0.00 & 0.66  & 0.18  & 0.48           & 0.94           \\
      \midrule
      Average                         & 0.00                        & 0.00                        & 0.00                            & 0.00                           & 0.38                                   & 0.00 & 0.256 & 0.168 & \textbf{0.506} & \textbf{0.556} \\
      \bottomrule
    \end{tabular}}
\end{table}

\begin{table}[t]
  \centering
  \footnotesize
  \setlength{\tabcolsep}{4pt}
  \caption{Per-task success (fraction of $50$ trials) on
    \texttt{libero-spatial}; baselines from
    CaP-X~\citep{fu2026capx,zhou2025liberopro}.  This is RHO's
    strongest suite and its widest margin: explicit perception-driven
    grounding of spatial referents (\texttt{between}, \texttt{next\_to},
    \texttt{on}) holds up under both perturbations (Pos $0.53$, Task
    $0.62$), while Agent0 stays near the floor ($0.12$/$0.14$) and
    $\pi_{0.5}$ again only memorizes isolated positions (on(ramekin) $0.98$,
    on(cabinet\_wood) $0.90$).  Best column average is in bold;
    \textit{Pick(src, plate)}: pick the black bowl from \textit{src} and
    place it on the plate.}
  \label{tab:libero-spatial-pertask}
  \resizebox{\textwidth}{!}{%
    \begin{tabular}{l*{10}{c}}
      \toprule
                                              & \multicolumn{2}{c}{OpenVLA} & \multicolumn{2}{c}{$\pi_0$} & \multicolumn{2}{c}{$\pi_{0.5}$} & \multicolumn{2}{c}{CaP-Agent0} & \multicolumn{2}{c}{\textbf{RHO}}                                                         \\
      \cmidrule(lr){2-3}\cmidrule(lr){4-5}\cmidrule(lr){6-7}\cmidrule(lr){8-9}\cmidrule(lr){10-11}
      Task (symbolic form)                    & Pos                         & Task                        & Pos                             & Task                           & Pos                                    & Task & Pos   & Task & Pos            & Task           \\
      \midrule
      Pick(between(plate,ramekin), plate)     & 0.00                        & 0.00                        & 0.00                            & 0.00                           & 0.02                                   & 0.00 & 0.22  & 0.14 & 0.92           & 0.60           \\
      Pick(table\_center, plate)              & 0.00                        & 0.00                        & 0.00                            & 0.00                           & 0.00                                   & 0.02 & 0.22  & 0.14 & 0.96           & 0.96           \\
      Pick(drawer\_top(cabinet\_wood), plate) & 0.00                        & 0.00                        & 0.00                            & 0.00                           & 0.00                                   & 0.00 & 0.02  & 0.10 & 0.06           & 0.04           \\
      Pick(next\_to(cookie\_box), plate)      & 0.00                        & 0.00                        & 0.00                            & 0.00                           & 0.00                                   & 0.02 & 0.00  & 0.10 & 0.00           & 0.28           \\
      Pick(next\_to(plate), plate)            & 0.00                        & 0.00                        & 0.00                            & 0.00                           & 0.00                                   & 0.00 & 0.10  & 0.20 & 0.56           & 0.64           \\
      Pick(next\_to(ramekin), plate)          & 0.00                        & 0.00                        & 0.00                            & 0.00                           & 0.12                                   & 0.02 & 0.30  & 0.14 & 0.78           & 0.76           \\
      Pick(on(cookie\_box), plate)            & 0.00                        & 0.00                        & 0.00                            & 0.00                           & 0.00                                   & 0.00 & 0.14  & 0.08 & 0.10           & 0.90           \\
      Pick(on(ramekin), plate)                & 0.00                        & 0.00                        & 0.00                            & 0.00                           & 0.98                                   & 0.02 & 0.02  & 0.20 & 0.64           & 0.42           \\
      Pick(on(stove), plate)                  & 0.00                        & 0.00                        & 0.00                            & 0.00                           & 0.02                                   & 0.00 & 0.08  & 0.14 & 0.86           & 0.98           \\
      Pick(on(cabinet\_wood), plate)          & 0.00                        & 0.00                        & 0.00                            & 0.00                           & 0.90                                   & 0.00 & 0.08  & 0.16 & 0.42           & 0.58           \\
      \midrule
      Average                                 & 0.00                        & 0.00                        & 0.00                            & 0.00                           & 0.20                                   & 0.01 & 0.118 & 0.14 & \textbf{0.530} & \textbf{0.616} \\
      \bottomrule
    \end{tabular}}
\end{table}

\paragraph{Object-swap failure analysis.}
\texttt{object-swap} is the only LIBERO-PRO cell that perturbs scene
composition: the BDDL relocates the named target object from its
training-distribution slot (\texttt{target\_object\_region},
$y\approx -0.24$) to a peripheral slot
(\texttt{other\_object\_region\_4}, $y\approx -0.08$), and a randomized
distractor fills the original slot.  Perception is not the failure
mode: Molmo and SAM still localize the relocated target at $0.94+$
mask confidence.  Of $439$ failed trials, $325$ ($74\%$) involve
\textbf{false-positive lift verification}: after a missed grasp, the
solver's \texttt{verify\_grasp\_lift} routine (added during evolution)
re-segments the source label and latches onto either the new
distractor or noise near the gripper with a score of $\approx 0.02$, returning
\texttt{lifted} despite empty fingers, and the subsequent place step
releases nothing.  The remaining $114$ ($26\%$) exhaust the
$12$-attempt grasp budget on a hand-authored grasp-pose schedule
(\texttt{container\_side\_yneg/ypos/xpos/xneg}, \texttt{center\_yaw0/90})
whose offsets and \texttt{close\_depth} defaults were calibrated for
the on-distribution slot.  Per-subtask success rates split cleanly by
whether the solver carries a per-object branch: $0\%$ on
\texttt{cream\_cheese}, \texttt{ketchup}, \texttt{tomato\_sauce},
\texttt{chocolate\_pudding} (no branch), $32$--$34\%$ on
\texttt{bbq\_sauce} and \texttt{orange\_juice} (hand-coded variants),
and $12$--$18\%$ on the remaining categories.  CaP-Bench exposes a
Contact-GraspNet \texttt{plan\_grasp} primitive that the selected
solver does call: at the swapped pose, its confidence drops from
$\approx 0.8$ to $\approx 0.2$, but it still returns liftable
candidates.  The mutator elaborated the solver from $427$ lines at
generation~2 to $2{,}016$ lines at generation~32, accumulating
roughly $53$ lines per generation in per-category branches and
manual-schedule entries.  This overlay is what fails on object-swap.

\paragraph{Warm-restart configuration changes.}
The initial lane set:
\begin{itemize}
  \item acceptance\_criterion = strict\_improvement
  \item val\_stage\_size unset
  \item num\_sampled\_groups unset
  \item num\_examples\_per\_group unset
\end{itemize}
The warm-restart lane was changed to:
\begin{itemize}
  \item acceptance\_criterion = improvement\_or\_equal
  \item val\_stage\_size = 45
  \item num\_sampled\_groups = 2
  \item num\_examples\_per\_group = 1
\end{itemize}
Each setting in the second configuration matches an addition in
Section~\ref{sec:method-rap}.

\subsubsection{Training ledger}

The reported run's persisted \texttt{state.budget} ledger records
$6{,}814$ evaluations, $625{,}276{,}104$ input tokens,
$88{,}851{,}968$ cached input tokens, and $3{,}144{,}685$ output
tokens; summed over the $87$ retained per-worktree
\texttt{.helix\_backend\_result.json} files, total backend usage is
$386{,}757{,}173$ input tokens, $373{,}486{,}336$ cached input tokens,
and $1{,}757{,}060$ output tokens.
At GPT-5.5 API pricing (\$5/M input, \$0.50/M cached input, \$30/M
output), the canonical \texttt{state.budget} tokens correspond to an
estimated training cost of \$2,821: \$2,682 for 536.4M uncached input
tokens, \$44 for 88.9M cached input tokens, and \$94 for 3.14M output
tokens.  The persisted \texttt{state.json} records
\texttt{cost\_usd=0.0}; the orchestrator does not track API costs
directly.  The estimate is derived from the token counts and public
pricing as of May 2026.
End-to-end wall-clock for the RHO training run was 79.4 hours
($\sim$3 days 7 hours) on a dual-socket Intel Xeon Platinum 8570 host
with two NVIDIA B200 GPUs, measured from the \texttt{.helix/config.toml}
creation time to the final \texttt{.helix/state.json} write.
The persisted run completed under \texttt{frontier\_type=instance} and
\texttt{max\_generations=200}, reaching \texttt{generation=200} and
\texttt{mutation\_counter=200}.
The re-parse finds $1{,}776$ REaaS calls across $87$ retained
HELIX worktree transcripts: $698$ reset, $576$ run, $293$ call,
$166$ vdm, and $43$ observe.
Across the $87$ retained frontier entries, \texttt{lineage.json} labels
$86$ nodes as mutations and $1$ as the seed; the original four-way
mutation taxonomy is not reliably recoverable from these artifacts.
Aggregated Python source diffs over the $87$ frontier worktrees against
the seed total $+153{,}519/-1{,}730$ lines, with the selected solver
introducing $74$ net new Python function definitions over the baseline.
The 200 mutation slots resolved into $87$ retained frontier members
($86$ accepted mutations and the generation-0 seed), $77$ rejected at
the training gate, $28$ skipped
because the parent already attained a perfect subsample score, and $8$
mutator-step failures attributable to host-side intermittent issues and
OpenAI API congestion (HTTP 4xx).  Judge-model and visual-difference-model
API costs are not included in the \$2{,}821 estimate above.
The sandbox's integrity controls fired zero times across the canonical
run: no candidate triggered protected-file tamper rejection or an
LLM-as-a-judge integrity flag of \textsc{major} or \textsc{severe}
severity (Appendix~\ref{app:judge-axes}).  Two LIBERO-PRO perturbation cells
remain weak (object-swap at $12.2\%$, object-task at $37.2\%$); the
object-swap collapse traces to mutator-added overlays in the selected
solver (a hand-fitted grasp schedule and a re-segmentation-based lift
verifier) rather than perception-primitive degradation
(Appendix~\ref{sec:exp-libero-pro-cells}, with the design implication
discussed in Section~\ref{sec:limitations}).
% Labels moved from former main-body Sec.~4.5 "Cost and remaining failure modes":
\label{sec:exp-cost}
\label{sec:exp-ablations}
\label{sec:exp-audit}
\label{sec:exp-train-cost}
\label{sec:exp-ablation-failure}
\label{sec:exp-robosuite}
\label{sec:exp-capx}
\label{sec:experiments-cap}
\label{sec:exp-overhead}

\subsubsection{Cross-run training budget}
\label{sec:appendix-cross-run-budget}

\Cref{tab:cross-run-budget} reports per-run token totals and dollar
estimates for the reported RHO run, the four accepted
backend-sensitivity alternates, and four first-mutation single-shot
baselines (one per base model: GPT-5.5, Opus, Sonnet, Qwen; the
GPT-5.5 baseline serves both Codex configurations they share).
CaP-Agent0~\citep{fu2026capx}, $\pi_{0.5}$~\citep{black2025pi05}, and
OpenVLA~\citep{kim2024openvla} do not publish the per-run
token/dollar ledger this comparison requires, so we omit them.  Accepted
alternates range from $\sim$\$1{,}282 (Opus, aggregated upstream chain)
down to \$124.44 (Sonnet 4.6 Behavior Tree, state recorded).  The \$732
Codex GPT-5.5 Single-file alt-config figure
covers only the 150-generation warm-restart segment whose tokens
\texttt{state.json} records; the \$1{,}665 single-file cost in
Table~\ref{tab:structural-prior} is the cumulative total for the same
lineage across its 100-generation parent and 150-generation
warm-restart.  All single-shot baselines and every evolved RHO
run start from the same Skeleton seed
(Appendix~\ref{app:seed}); their costs and held-out scores appear in
Table~\ref{tab:no-loop-floor} in the main text.  The strongest single-shot point (Codex
GPT-5.5 at $210/700 = 30\%$) is the reference cited in
\S\ref{sec:exp-capx}.  Against the full single-shot range of
\$0--\$3.19 per pass and $13$--$30\%$ on the same split, the reported
RHO run pays \$2{,}821 once, reaches $490/700$, and then issues
$0$ LLM code-generation calls per control step
(\S\ref{sec:exp-train-cost}).

\subsubsection{Evaluator integrity details and unmeasured variants}

For the reported run, accepted candidates were generated under
protected-file manifest checks and evaluator-internal omission.  The
selected solver was replayed under the VDM-disabled S4 harness.  These
checks support the training-time evaluator-integrity claim; sandbox-off
and manifest-only variants are not measured in this submission.

\clearpage
% =====================================================================
% Paper section: RHO / HELIX applied to the RAI manipulation_o3de benchmark.
%
% Two usage modes:
% (a) Stand-alone: pdflatex on this file uses the conditional
% preamble below.
% (b) Included: define \PaperHostsRaiSection in the master
% document before \input{} to suppress the preamble.
% =====================================================================

\ifx\PaperHostsRaiSection\undefined
  \documentclass[11pt]{article}
  \usepackage[margin=1in]{geometry}
  \usepackage{booktabs}
  \usepackage{amsmath,amssymb}
  \usepackage{graphicx}
  \usepackage{xcolor}
  \usepackage{caption}
  \usepackage{fancyvrb}
  \usepackage{fvextra}                    % defines Verbatim breaklines (matches master main.tex)
  \usepackage{microtype}
  \usepackage[hidelinks]{hyperref}
  % Lightweight \citep stub when the bibliography is unavailable in
  % stand-alone builds (suppresses undefined-control-sequence errors).
  \providecommand{\citep}[1]{\textnormal{[#1]}}
  \title{RHO on the RAI \texttt{manipulation\_o3de} Benchmark}
  \author{}
  \date{}
\begin{document}
\maketitle
\fi

% =====================================================================
\subsection{RHO on an LLM-in-the-loop Robotics Stack}
\label{sec:rai-o3de}

To test whether RHO transfers beyond the
primitive-only CaP-Bench surface, we instantiate it with HELIX, REaaS, and
evaluator isolation on Robotec's
RAI~\citep{rai} \texttt{manipulation\_o3de} benchmark. The deployment
loop here \emph{includes} an LLM, so the candidate repository defines
the prompting and tool surface that the deployed LLM consumes rather
than a Python controller. The same evolutionary recipe and acceptance
gate are reused; the RAI run changes the candidate artifact and adds a fixed
tool-name contract.

% ---------------------------------------------------------------------
\subsubsection{Background: the RAI framework and the
  \texttt{manipulation\_o3de} benchmark}
\label{sec:rai-bg}

\textbf{RAI}~\citep{rai} is an open-source agent framework released
by Robotec.AI for connecting off-the-shelf LLMs to physical robots.
Unlike many LLM-agent benchmarks, RAI is constructed directly on top
of the ROS 2 middleware that production robotics systems already
use: agents talk to perception, planning, and control nodes over the
same topics and services that real robot software speaks. This
makes it a useful stress test for industrial settings, where the
value proposition of an LLM agent is precisely that it can drive the
existing ROS 2 interfaces a robotics team has already deployed.

RAI ships a benchmark suite, \texttt{rai\_bench}, whose
\texttt{manipulation\_o3de} track exercises a tabletop manipulator
simulated end-to-end with O3DE (Open 3D Engine) for rendering and
physics, ROS 2 + MoveIt for control, and
Grounding~DINO + SAM\,2~\citep{ravi2024sam2} for open-vocabulary
perception. The agent is given three fixed tools,
\texttt{get\_object\_positions} (perception query by name),
\texttt{get\_ros2\_camera\_image} (vision), and \texttt{move\_to\_point}
(motion: a \texttt{grab} or \texttt{drop} at a 3-D coordinate), and is
asked to accomplish a natural-language manipulation task. An automated
scorer checks the final object configuration against per-task tolerance
windows. Throughout, the LLM is exposed to exactly the same ROS 2
surface that one would expose on a physical robot; nothing about the
agent interface is simulation-specific.

\begin{figure}[p]
  \centering
  \setlength{\tabcolsep}{2pt}
  \renewcommand{\arraystretch}{1}
  \begin{tabular}{@{}c@{\hspace{3pt}}c@{\hspace{3pt}}c@{}}
                                                     & \small Initial state                                                 & \small Final state reached                                          \\[1pt]
    \rotatebox{90}{\parbox{0.20\linewidth}{\centering\small\textsc{PlaceObjectAtCoord}}}  & \includegraphics[width=0.43\linewidth]{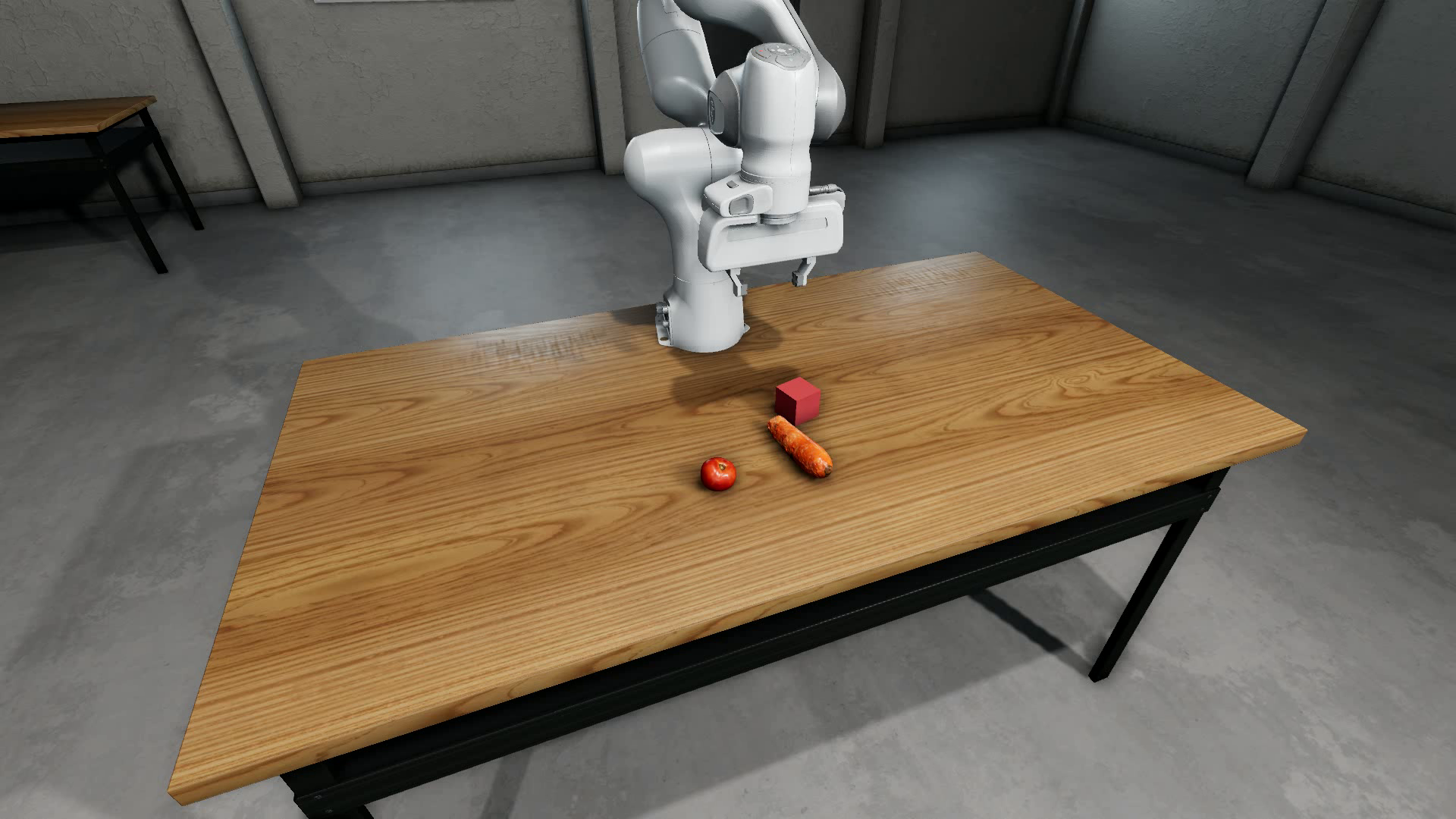}  & \includegraphics[width=0.43\linewidth]{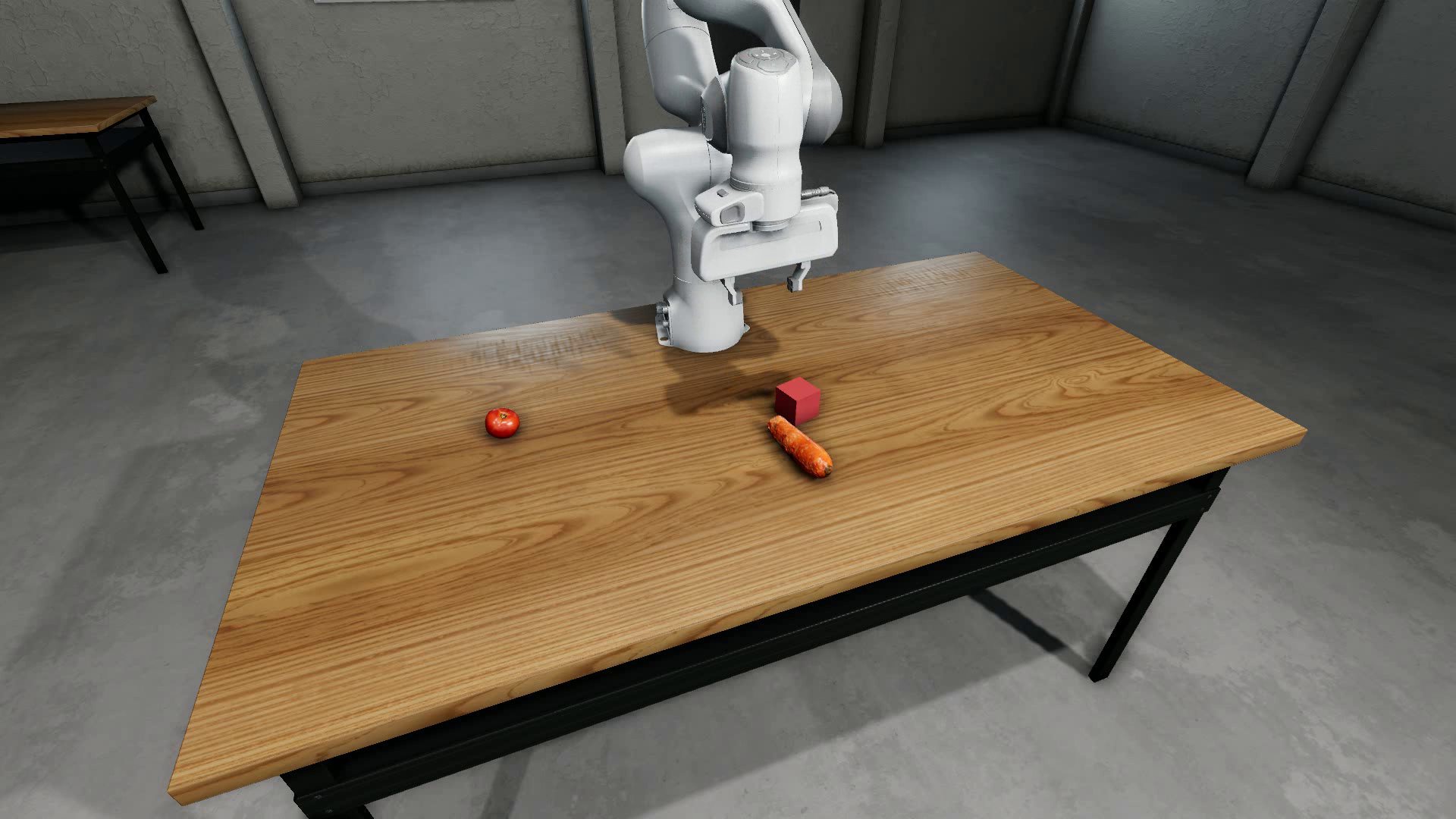}  \\[1pt]
    \rotatebox{90}{\parbox{0.20\linewidth}{\centering\small\textsc{MoveObjectsToLeft}}}   & \includegraphics[width=0.43\linewidth]{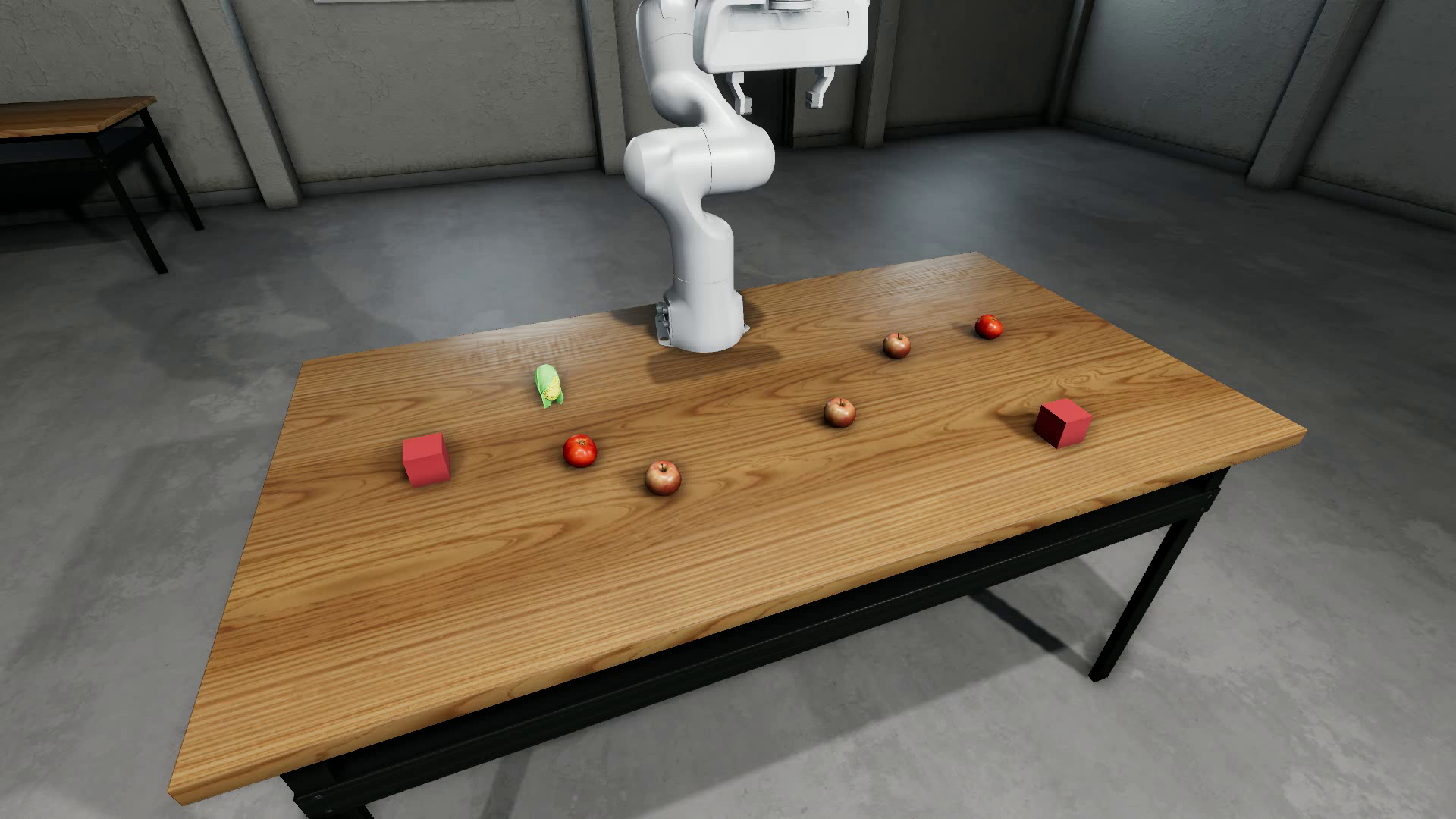}   & \includegraphics[width=0.43\linewidth]{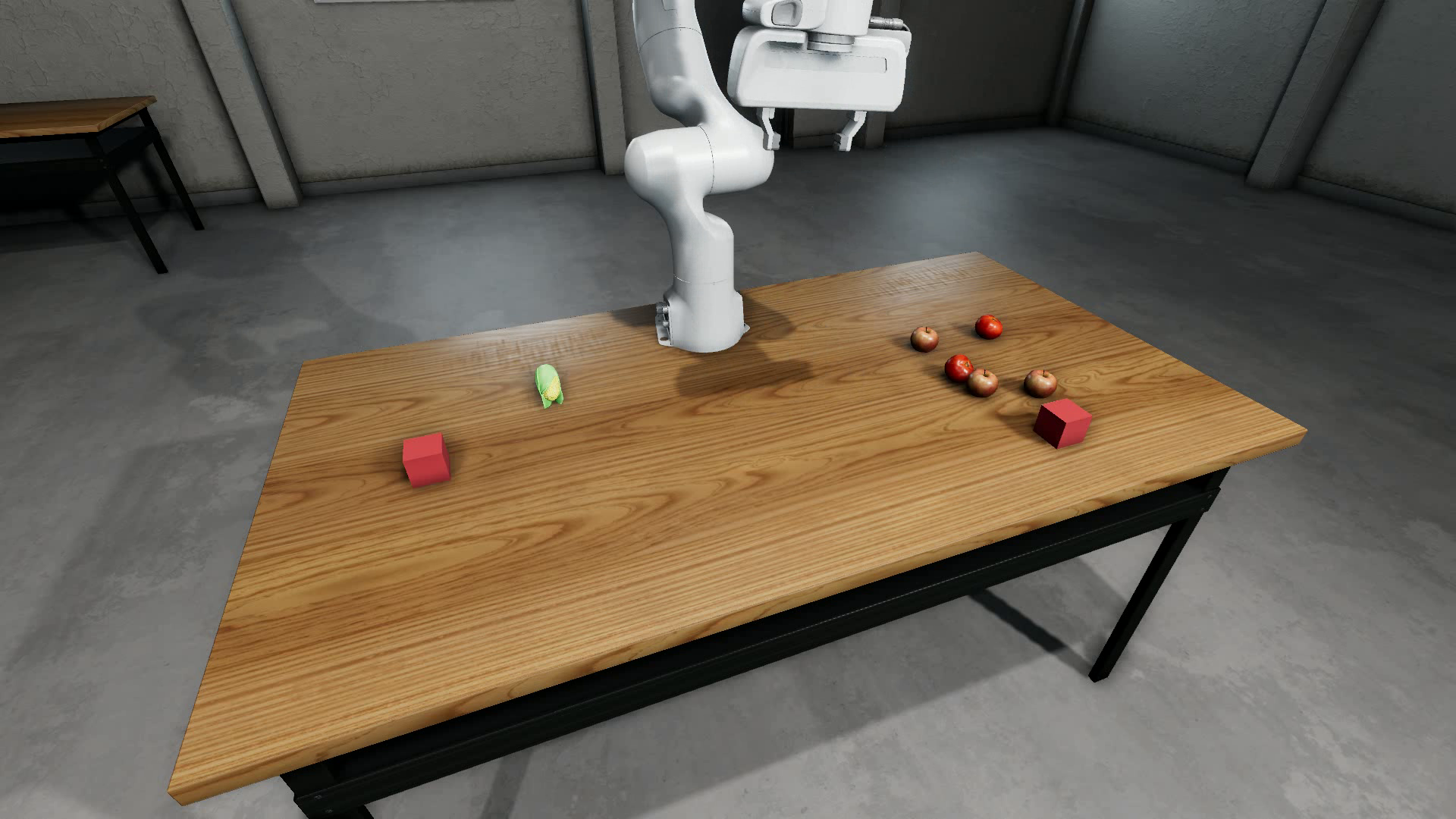}   \\[1pt]
    \rotatebox{90}{\parbox{0.20\linewidth}{\centering\small\textsc{PlaceCubes}}}          & \includegraphics[width=0.43\linewidth]{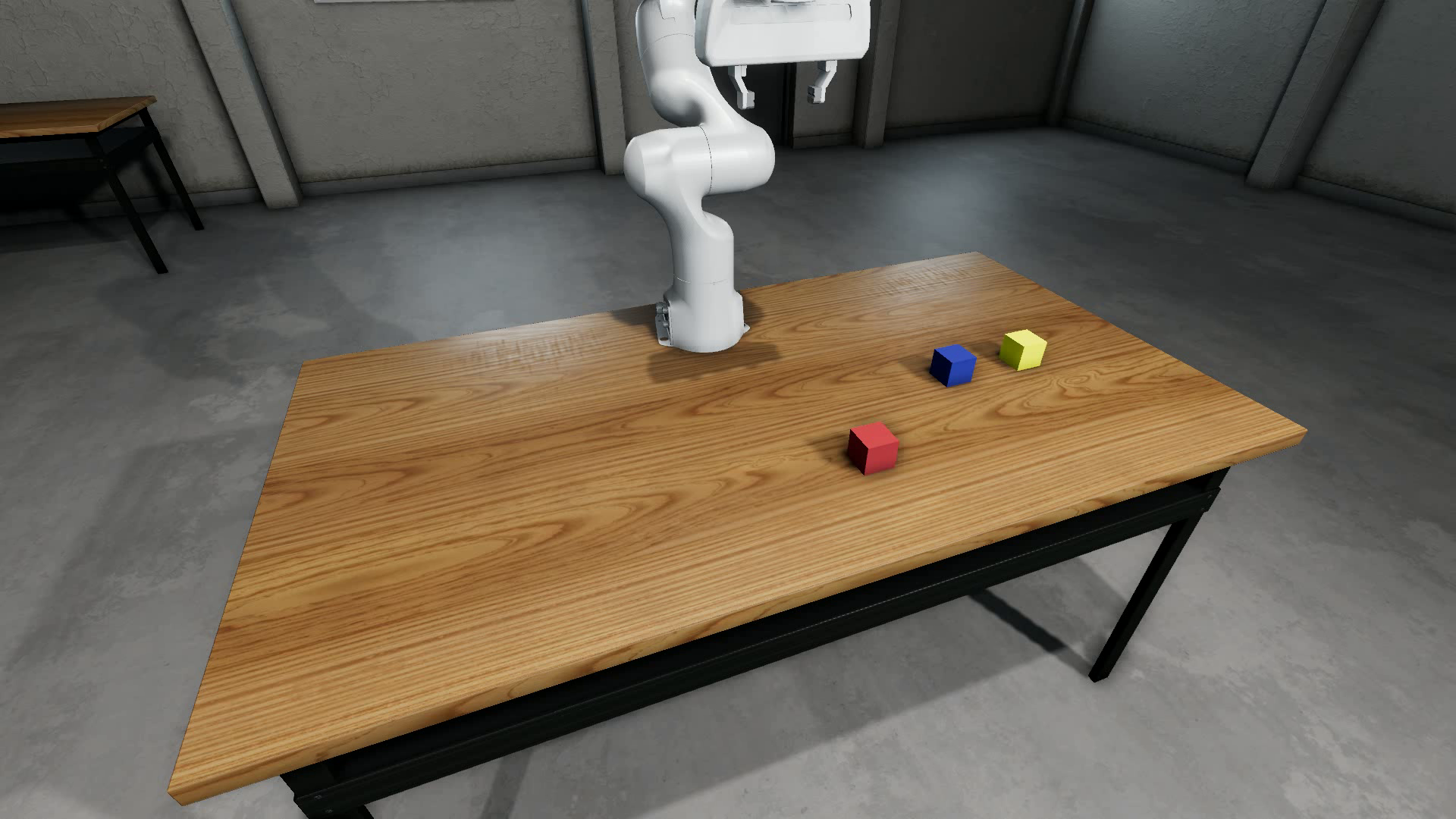}          & \includegraphics[width=0.43\linewidth]{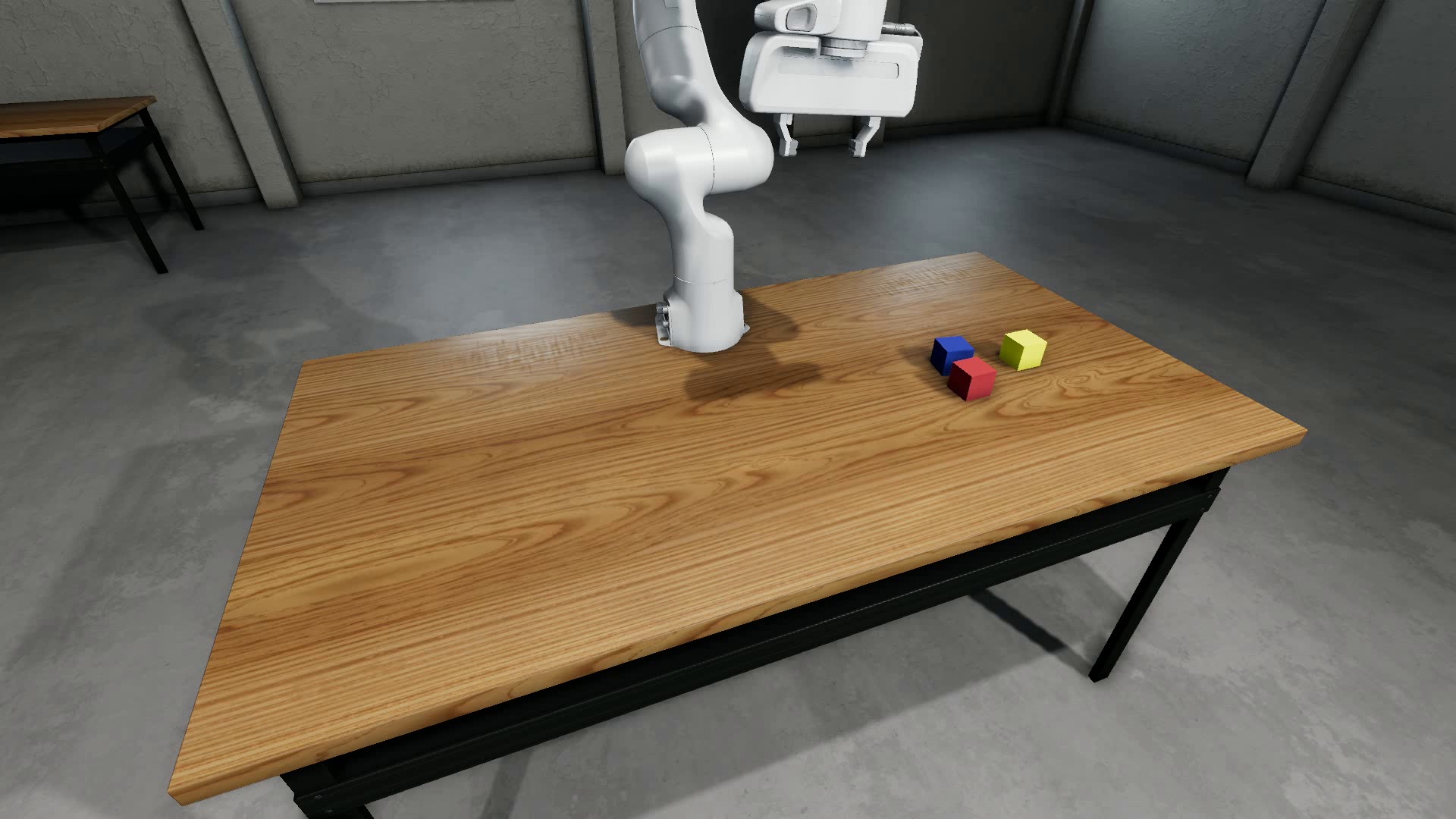}          \\[1pt]
    \rotatebox{90}{\parbox{0.20\linewidth}{\centering\small\textsc{BuildCubeTower}}}      & \includegraphics[width=0.43\linewidth]{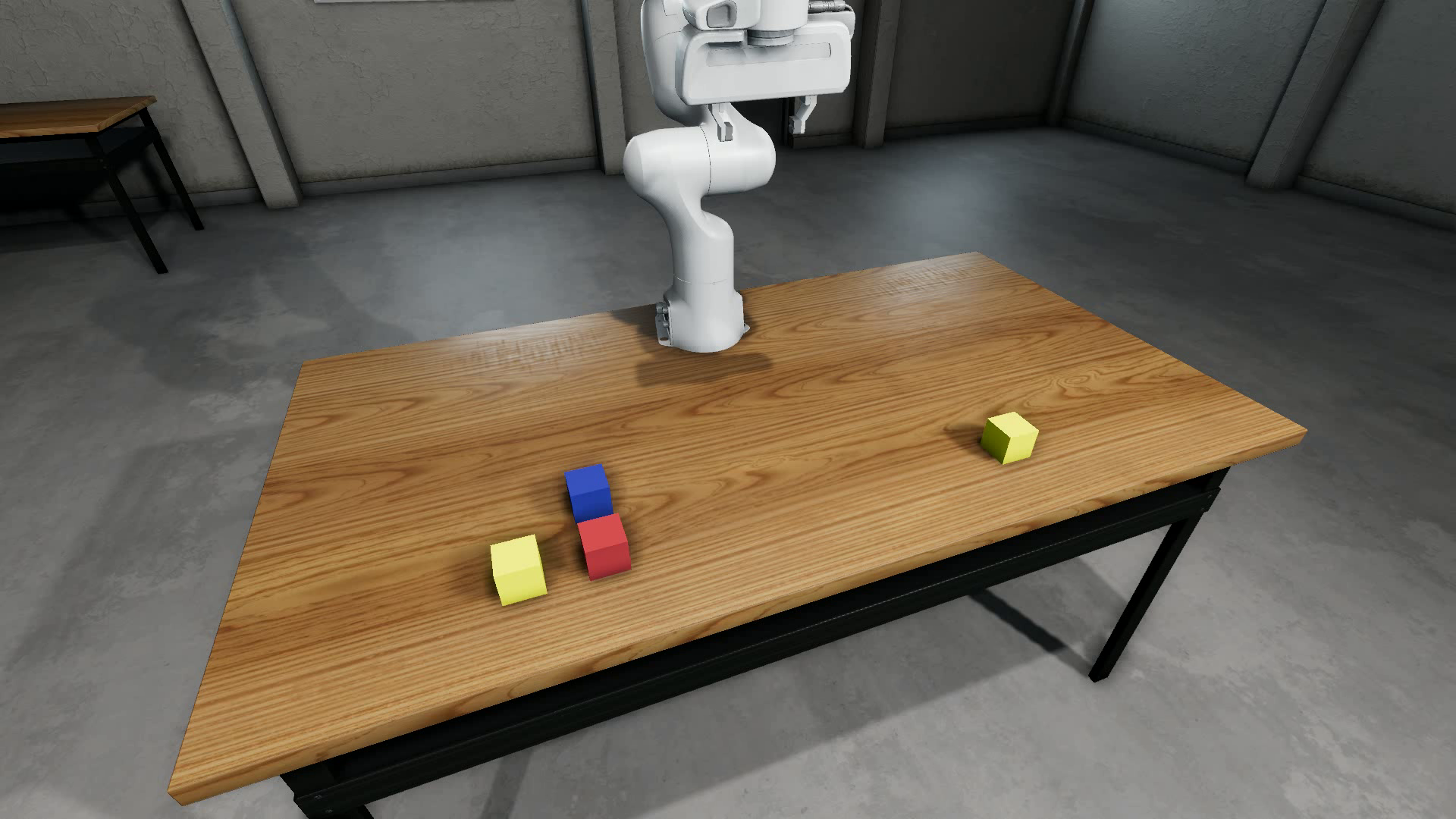}           & \includegraphics[width=0.43\linewidth]{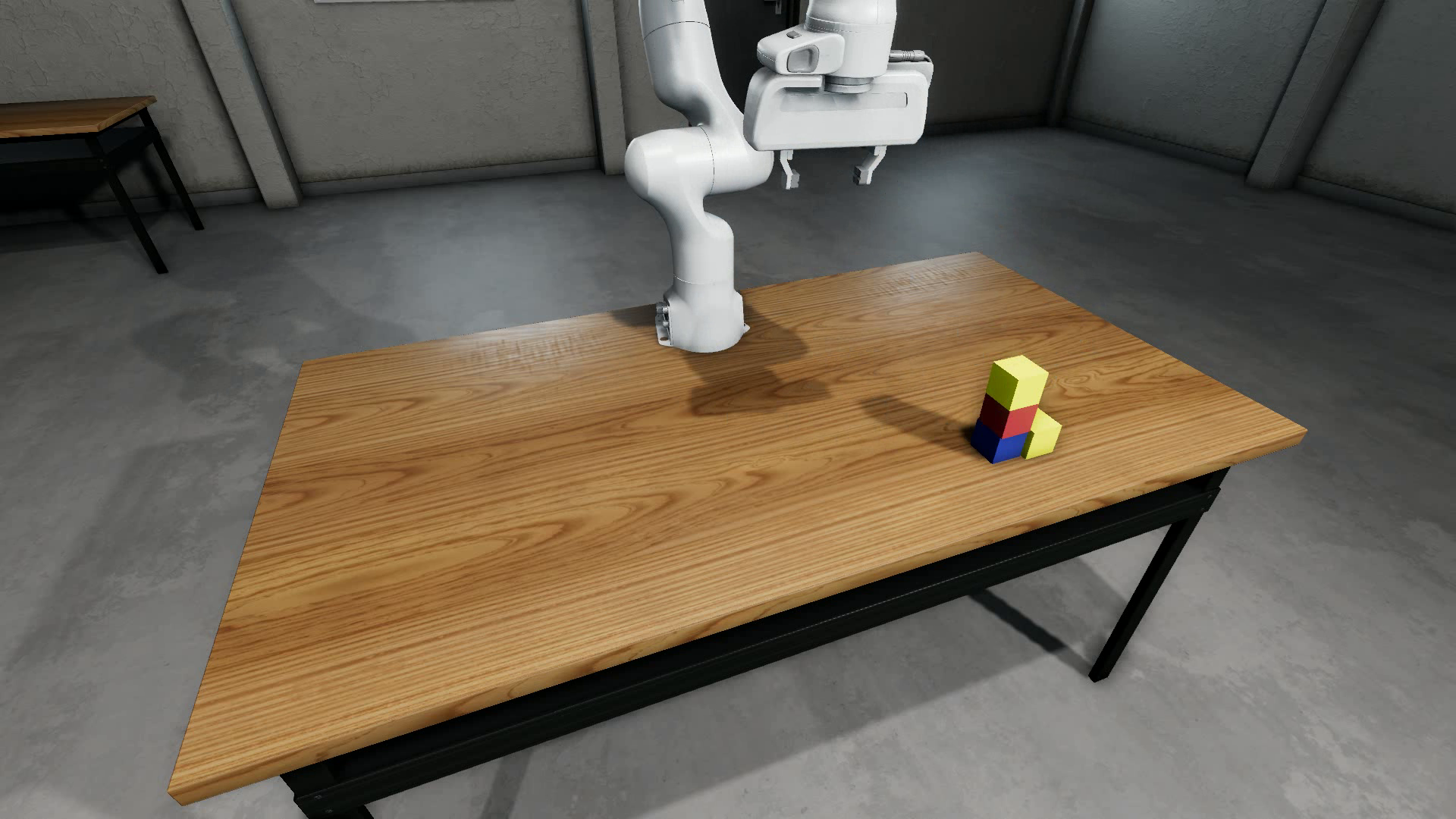}           \\[1pt]
    \rotatebox{90}{\parbox{0.20\linewidth}{\centering\small\textsc{GroupObjects}}}        & \includegraphics[width=0.43\linewidth]{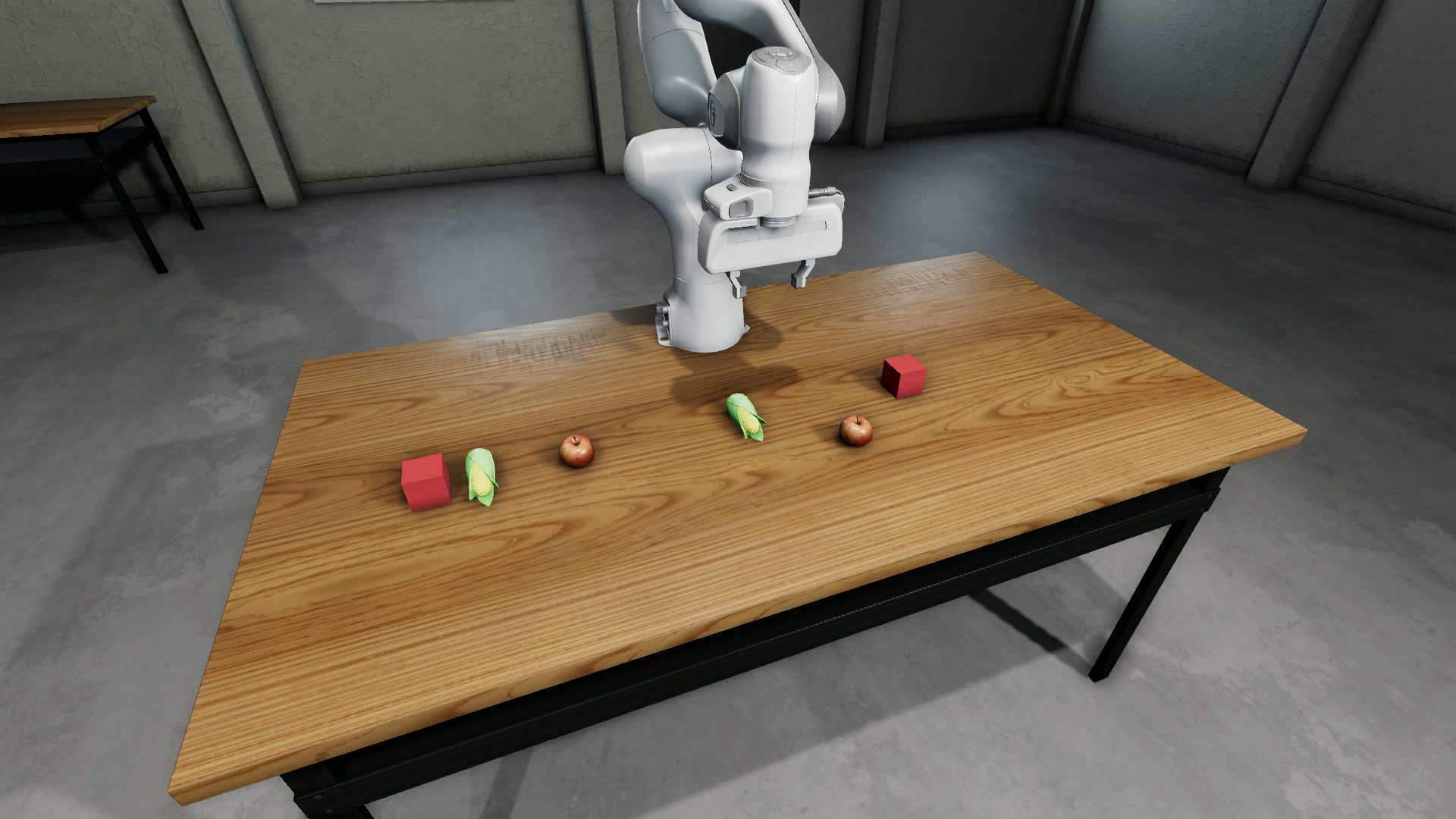}        & \includegraphics[width=0.43\linewidth]{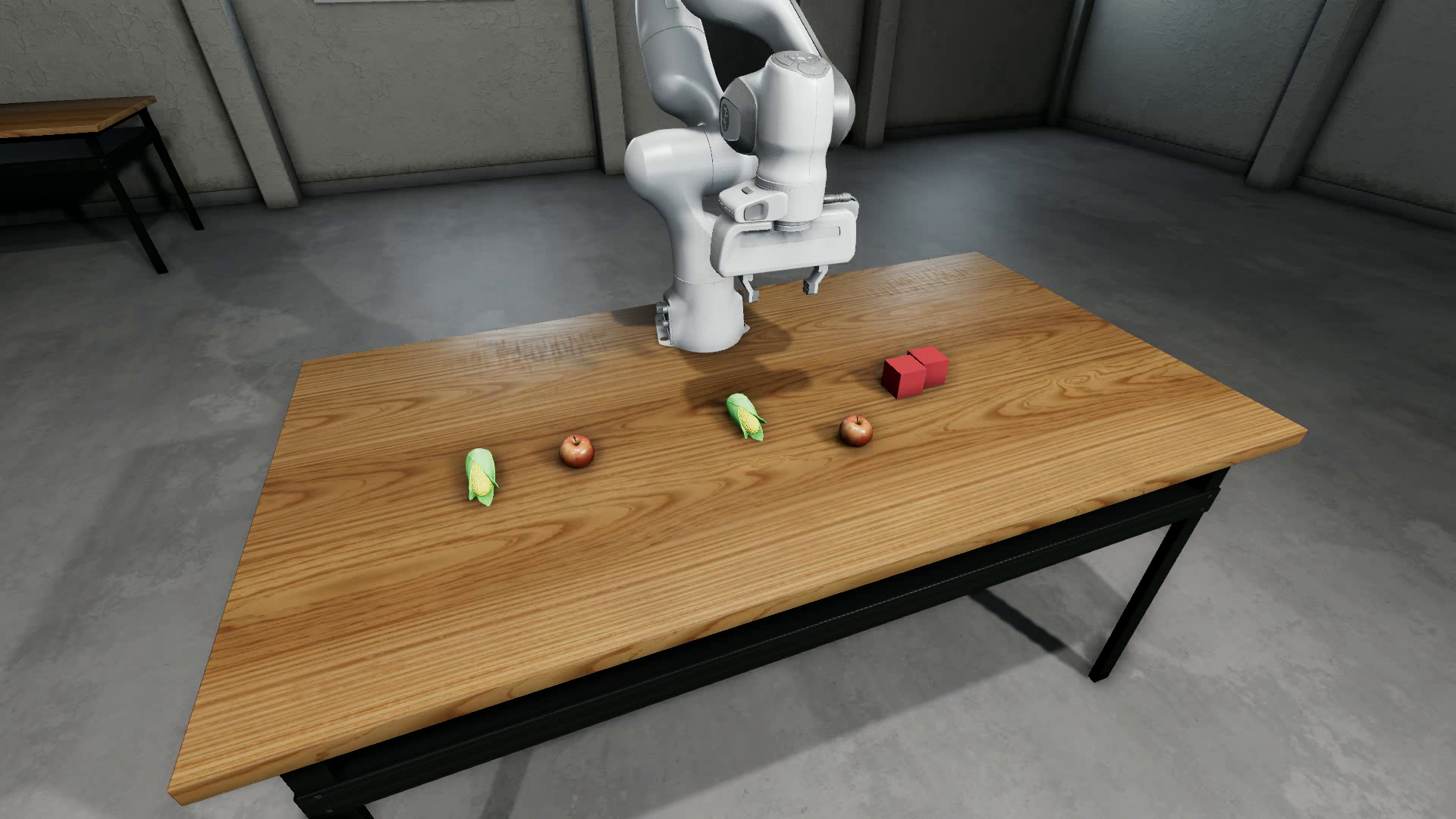}        \\
  \end{tabular}
  \caption{Representative held-out trials of the five RAI
    \texttt{manipulation\_o3de} task classes (rows), each shown in its
    initial state (left) and the final state reached by an evolved RHO
    controller (right).
    \textsc{PlaceObjectAtCoord}: move one named object to a target
    $(x, y)$ within a tolerance (all-or-nothing).
    \textsc{MoveObjectsToLeft}: move every object of the requested
    type(s) to the $+y$ (left) half of the table while leaving others
    in place.
    \textsc{PlaceCubes}: arrange cubes so each is adjacent to at least
    one other cube.
    \textsc{BuildCubeTower}: stack the requested-color cubes into a
    single vertical tower, excluding other object types.
    \textsc{GroupObjects}: gather objects into one connected, fully
    separated cluster per type.
    Except for \textsc{PlaceObjectAtCoord}, classes are scored by
    partial credit (the fraction of objects satisfying the goal).}
  \label{fig:rai-o3de-tasks}
\end{figure}

\paragraph{Task classes.}
The benchmark defines five task classes:
\textsc{PlaceObjectAtCoord} (move a named object to a specified
$(x, y)$ target), \textsc{MoveObjectsToLeft} (move all instances of
a category to the left half of the table), \textsc{PlaceCubes}
(place colored cubes adjacent to each other), \textsc{BuildCubeTower}
(stack cubes vertically), and \textsc{GroupObjects} (cluster objects
of the same category together). Each class is instantiated at
several difficulty levels that vary the number of distractor
objects, the tightness of the success tolerance window, and whether
initial conditions include pre-stacked or pre-clustered objects.
Figure~\ref{fig:rai-o3de-tasks} shows a representative held-out trial
of each class.

\paragraph{Difficulty levels.}
The benchmark exposes five difficulty bands
(\textsc{trivial}, \textsc{easy}, \textsc{medium}, \textsc{hard},
\textsc{very\_hard}) that adjust three independent levers in concert:

\begin{itemize}
  \item \emph{Scene clutter.} \textsc{trivial} scenes contain a
        single object on the tabletop; \textsc{easy} 2--3 objects;
        \textsc{medium} 4--6; \textsc{hard} and \textsc{very\_hard}
        up to a dozen objects. Several \textsc{hard} scenes
        additionally carry a \texttt{\_stacked} flag, meaning the
        scene boots with cubes already vertically arranged, so the
        agent must plan around existing structure rather than
        building from scratch.
  \item \emph{Task statement.}
        \textsc{MoveObjectsToLeft} and \textsc{GroupObjects} progress
        from single-type queries (``move all carrots left'',
        ``cluster the apples'') at \textsc{trivial}/\textsc{easy} to
        multi-type combinations (``move carrots, apples, and yellow
        cubes left'') at \textsc{medium} and above.
        \textsc{BuildCubeTower} starts at three colors and ramps
        toward single- or two-color towers (forcing the agent to
        stack identical-looking cubes precisely) at \textsc{very\_hard}.
  \item \emph{Pass criteria.} Several classes tighten their
        numerical success thresholds at higher levels. The cleanest
        example is \textsc{PlaceCubes}: \textsc{easy} accepts cubes
        within $0.20$\,m of a cluster partner, while \textsc{medium}
        and \textsc{hard} halve this to $0.10$\,m on much more
        cluttered scenes. \textsc{PlaceObjectAtCoord} runs only at
        the looser end of the spectrum
        (\textsc{trivial} and \textsc{easy}) and is omitted from
        \textsc{medium} onwards entirely.
\end{itemize}

\paragraph{Per-difficulty class coverage.}
Not every task class is instantiated at every difficulty; the
benchmark's task generator only emits a class at levels where it has
tuned scene configurations and pass criteria for it. The classes
that do appear overlap several adjacent levels rather than being
level-exclusive (Table~\ref{tab:rai-class-coverage}).
\textsc{MoveObjectsToLeft}, for example, appears at four of the five
levels; what changes between them is the scene around it and the
task statement, not the underlying scoring rubric.

\begin{table}[tp]
  \centering
  \caption{Task class availability across the five RAI
    \texttt{manipulation\_o3de} difficulty bands. A check mark
    indicates that the benchmark's predefined task generator emits
    one or more scenarios of that class at that level. Several
    classes span three or four adjacent levels, with difficulty
    varying through scene clutter, task statements, and (for
    \textsc{PlaceCubes}) the numerical pass criterion.}
  \label{tab:rai-class-coverage}
  \begin{tabular}{lccccc}
    \toprule
    Task class                  & \textsc{trivial} & \textsc{easy} & \textsc{medium} & \textsc{hard} & \textsc{very\_hard} \\
    \midrule
    \textsc{PlaceObjectAtCoord} & \checkmark       & \checkmark    &                 &               &                     \\
    \textsc{MoveObjectsToLeft}  & \checkmark       & \checkmark    & \checkmark      & \checkmark    &                     \\
    \textsc{PlaceCubes}         &                  & \checkmark    & \checkmark      & \checkmark    &                     \\
    \textsc{BuildCubeTower}     &                  &               & \checkmark      & \checkmark    & \checkmark          \\
    \textsc{GroupObjects}       &                  &               & \checkmark      & \checkmark    & \checkmark          \\
    \bottomrule
  \end{tabular}
\end{table}

For this study we package two of the bands into separate RHO
cells, \textsc{easy} and \textsc{hard}. We chose these two rather
than the strict extremes (\textsc{trivial} and \textsc{very\_hard})
because they expose the largest \emph{within-band} diversity: each
covers three or four task classes and spans both looser and tighter
constraints within its respective regime, giving RHO a
non-trivial mix of tasks to specialize against
(Table~\ref{tab:rai-difficulty}).

\begin{table}[tp]
  \centering
  \caption{Per-difficulty class coverage and split sizes used in this
    section. \textsc{Easy} draws on the three classes that the bench
    instantiates at its looser end; \textsc{hard} on the four classes
    that survive into the cluttered, tighter-criterion regime. Each
    cell uses a class-stratified 50/50 train/test split; the train
    half drives RHO evolution and the test half is evaluated only
    after evolution completes.}
  \label{tab:rai-difficulty}
  \begin{tabular}{lcc}
    \toprule
    Task class                  & \textsc{easy} (train/test) & \textsc{hard} (train/test) \\
    \midrule
    \textsc{PlaceObjectAtCoord} & 15 / 15                    & --                         \\
    \textsc{MoveObjectsToLeft}  & 4 / 4                      & 5 / 5                      \\
    \textsc{PlaceCubes}         & 2 / 2                      & 5 / 5                      \\
    \textsc{GroupObjects}       & --                         & 9 / 8                      \\
    \textsc{BuildCubeTower}     & --                         & 3 / 3                      \\
    \midrule
    \textbf{Total}              & \textbf{21 / 21}           & \textbf{22 / 21}           \\
    \bottomrule
  \end{tabular}
\end{table}

\textsc{Easy} instances have one or two target objects, the
loose $0.10$--$0.20$\,m pass criteria summarized above, and no
stacking; the seed agent already solves a non-trivial fraction.
\textsc{Hard} instances contain five or more objects with the
tighter $0.10$\,m \textsc{PlaceCubes} criterion, include pre-stacked
initial conditions, and introduce the two classes
(\textsc{GroupObjects}, \textsc{BuildCubeTower}) on which the seed
agent is closest to total failure. Splitting evolution by
difficulty allows RHO to specialize rather than average
opposing per-class lifts and regressions inside a mixed bench.

% ---------------------------------------------------------------------
\subsubsection{Experimental setup}
\label{sec:rai-setup}

The headline configuration runs RHO with
repository-level mutation (Section~\ref{sec:method-rap}),
allowing the mutator to edit the full repository (the agent's system
prompt and the bodies of the three tools). We additionally run a
single-file ablation that strips the mutator down to prompt edits
only, recovering a GEPA-style~\citep{agrawal2026gepa} reflective
optimizer, in order to isolate how much of the held-out lift is
attributable to multi-file repository search versus prompt search
alone.

\begin{itemize}
  \item \textbf{Multi-file evolution (MF, headline).} RHO is
        allowed to mutate both the agent's system prompt
        (\texttt{solver/prompts.py}) and the \emph{bodies} of the
        three tools (\texttt{solver/tools.py}), under a frozen
        tool-name contract enforced by an AST-level pre-commit
        check. A candidate that registers a new tool name, removes
        one, or changes any tool signature is rejected before it
        enters the frontier. This restricts the search space to
        changes that the LLM caller cannot directly observe, which
        is the most industry-realistic deployment scenario, since
        the agent's public tool API belongs to the surrounding
        production system and cannot be rewritten on a per-task
        basis.
  \item \textbf{Single-file evolution (SF, ablation).} RHO is
        restricted to mutating only the agent's system prompt
        (\texttt{solver/prompts.py}); the tool implementations are
        pinned to RAI's upstream wrappers around perception and
        motion. Comparing SF against MF on otherwise identical
        infrastructure isolates the contribution of multi-file
        repository search.
\end{itemize}

RHO runs for ten generations per cell on the train half, with
a class-stratified minibatch size of three (one task per class per
minibatch on \textsc{easy}; one class rotated out per minibatch on
\textsc{hard}, which has four classes). The mutator is the
\texttt{codex} CLI in sandbox mode, identical to that used in the
canonical CaP-Bench Codex GPT-5.5 Multi-file run~\citep{fu2026capx}.

\paragraph{Held-out evaluation.}
For each cell we report \textsf{seed} (the unmodified upstream
prompt and tools, re-evaluated independently in each cell over its
own three runs, so the MF-cell and SF-cell seed means differ
slightly because of sampler noise) and \textsf{best-by-train} (the candidate with the
highest training-set mean across the evolution frontier) on the
test half, which is never seen during evolution. To suppress
single-run sampler noise, we average each cell over three independent
runs and report mean $\pm$ std (in percentage points). Within-run
paired lifts ($\textsf{best}_i - \textsf{seed}_i$) are tested across
the resulting 63 per-task per-run pairs (21 tasks $\times$ 3 runs).

\paragraph{Hardware.}
All experiments in this section were run on a single workstation
built around an \textbf{AMD Ryzen AI Max+ 395}
APU with 128\,GB of unified memory and an integrated Radeon~8060S GPU. The entire
pipeline (ROS 2, MoveIt, the O3DE simulator with Vulkan rendering,
Grounding~DINO and SAM\,2 perception, and the HELIX \texttt{codex}
mutator) runs concurrently on this single APU. Wall-times reported
below are end-to-end on this hardware.

% ---------------------------------------------------------------------
\subsubsection{Results: the easy subset}
\label{sec:rai-easy}

Table~\ref{tab:rai-easy-headline} summarizes held-out test
performance on the \textsc{easy} subset, averaged over three runs.
The headline MF configuration lifts the test mean from
\textbf{68.3\%} to \textbf{88.9\%} ($+20.6$\,pp, $p < 0.001$ on the
per-task paired $t$-test). The SF ablation, which strips
HELIX's mutator down to prompt edits only, still produces a
statistically significant lift (\textbf{65.1\%} $\to$
\textbf{79.4\%}, $+14.3$\,pp, $p \approx 0.012$) but a smaller one:
the MF best-by-train candidate is $+9.5$\,pp above the SF
best-by-train candidate in absolute test mean (95\% CI
$[+5.0, +14.0]$\,pp, $t = 4.24$). The SF ablation is therefore
sufficient to recover roughly $\tfrac{2}{3}$ of the MF lift on this
benchmark; the remaining third is attributable to RHO's freedom
to additionally edit the tool implementations.

\begin{table}[tp]
  \centering
  \caption{Held-out test scores on the RAI \textsc{easy} subset
    ($n = 21$ tasks, averaged over 3 runs, GPT-5.1). MF is the
    headline repository-level mutation configuration; SF is
    the single-file ablation that restricts RHO to prompt
    edits. \emph{Lift} is the within-run paired difference
    $\textsf{best}_i - \textsf{seed}_i$ averaged over runs; the
    $t$-statistic and 95\% CI are computed on the resulting 63
    per-task per-run paired differences. All values in percent.}
  \label{tab:rai-easy-headline}
  \setlength{\tabcolsep}{4pt}
  \begin{tabular}{lcccc}
    \toprule
    Cell          & Seed (mean $\pm$ std) & Best (mean $\pm$ std)   & Lift        & 95\% CI ($t$)                     \\
    \midrule
    MF (headline) & $68.3 \pm 9.9$        & $\mathbf{88.9 \pm 2.8}$ & $+20.6$\,pp & $[+10.0, +32.0]$\,pp ($t = 3.67$) \\
    SF (ablation) & $65.1 \pm 7.3$        & $\mathbf{79.4 \pm 2.8}$ & $+14.3$\,pp & $[+4.0, +25.0]$\,pp ($t = 2.61$)  \\
    \bottomrule
  \end{tabular}
\end{table}

A useful side-channel diagnostic is the per-run variance of the
evolved candidates: both MF and SF land at $\sigma = 2.8$\,pp across
the three runs, roughly $3\times$ tighter than the \emph{seed}
prompt's per-run variance on the same tasks
($\sigma = 9.9$\,pp for the MF seed, $\sigma = 7.3$\,pp for the SF
seed). We attribute this to the evolved candidates narrowing the agent's
effective action distribution, so the same OpenAI sampler perturbations
translate into smaller score swings.

\subsubsection*{Compute: tool calls and wall time}

\begin{table}[tp]
  \centering
  \caption{Held-out test compute totals across the 21 RAI
    \textsc{easy} tasks (representative single run; the same sign
    and approximate magnitude hold under the 3-run average). Both
    the headline MF configuration and the SF ablation reach their
    higher accuracy with \emph{fewer}, not more, tool calls and
    lower wall time than the seed.}
  \label{tab:rai-easy-compute}
  \begin{tabular}{lrrrr}
    \toprule
    Cell          & Metric              & Seed & Best & $\Delta$           \\
    \midrule
    MF (headline) & Total tool calls    & 108  & 89   & $\mathbf{-17.6\%}$ \\
    MF (headline) & Total wall time (s) & 686  & 660  & $-3.8\%$           \\
    \midrule
    SF (ablation) & Total tool calls    & 98   & 89   & $-9.2\%$           \\
    SF (ablation) & Total wall time (s) & 630  & 578  & $-8.2\%$           \\
    \bottomrule
  \end{tabular}
\end{table}

Table~\ref{tab:rai-easy-compute} shows that the held-out accuracy
lifts come with \emph{lower}, not higher, inference cost. The
headline MF candidate cuts tool calls by \textbf{17.6\%} relative to
its seed with a modest additional wall-time improvement. The SF
ablation reaches a similar absolute tool-call count by editing the
prompt alone (a \textbf{9.2\%} reduction off a tighter seed), with
an \textbf{8.2\%} wall-time improvement.

The MF candidate achieves this efficiency
\emph{despite} its evolved \texttt{tools.py} having grown from
$375$ lines in the seed to $1{,}253$ lines after evolution: the
additional tool-side logic encodes more discriminating single calls
(color verification, multi-query perception fallback, detection
caching) rather than adding per-call overhead. The net effect is a
more capable agent that, in addition, finishes each task in fewer
round-trips, which matters wherever each tool invocation is a real
hardware round-trip.

\subsubsection*{What RHO learned: insights from the evolved
  prompts and tools}
\label{sec:rai-insights}

The MF-evolved candidate (generation 8 of 10) is the natural
artifact to inspect: it carries both the prompt-side and the
tool-side edits RHO considered worth keeping. Its mutations
fall into two layers.

The first is a structured rewrite of the seed prompt. RAI's stock
manipulation system prompt (copied verbatim from
\texttt{rai\_bench.manipulation\_o3de.interfaces.ManipulationTask})
is just six lines:

\begin{quote}\small\ttfamily
  You are a robotic arm with interfaces to detect and manipulate objects.\\
  Here are the coordinates information:\\
  x - front to back (positive is forward)\\
  y - left to right (positive is right)\\
  z - up to down (positive is up)\\
  Before starting the task, make sure to grab the camera image to understand the environment.
\end{quote}

\noindent (Note the fourth line, which inverts the convention the
benchmark actually uses for scoring. This is a single-token error
in the upstream library that we deliberately kept in the seed so
RHO would have to discover it on its own.) The MF-evolved
candidate expands this into 78 lines of task-specific guidance,
which the SF ablation arrives at independently in the same five
recognizable categories:

\begin{enumerate}
  \item \textbf{Tool-call discipline.} Explicit guidance to
        distinguish \texttt{get\_ros2\_camera\_image} (no arguments)
        from \texttt{get\_ros2\_image}
        (\texttt{topic="/color\_image5"}); a ``perceive once,
        then act'' loop that blocks the seed pattern of re-polling
        perception between every move.
  \item \textbf{Coordinate-frame disambiguation.} The seed agent
        routinely treated $+y$ as \emph{right} when the bench
        actually defines $+y$ as \emph{left}. The evolved prompt
        states the convention up front as a hard rule.
  \item \textbf{Object-name canonicalization.} Singular forms
        (\texttt{"apple"}, not \texttt{"apples"}), color-qualified
        cube names (\texttt{"red cube"}, \texttt{"blue cube"}), and
        an explicit dedup rule for cases where different color
        queries return overlapping centroids.
  \item \textbf{Pick/place hard rules.} A three-step recipe
        (detect $\rightarrow$ grab at the exact returned centroid
        $\rightarrow$ drop), an explicit ``no two grabs in a row''
        constraint, and a ``use the detected $z$ for tabletop
        drops'' rule.
  \item \textbf{Task-specific recipes.} One short recipe per task
        class in the train set, each ending with the meta-instruction
        ``stop when satisfied; additional moves reduce score''
        (the benchmark penalises unnecessary actions).
\end{enumerate}

To make these concrete, we give two excerpts from the MF-evolved prompt.
The first is RHO's fix for the coordinate-frame bug above
(category 2), now stated as a hard rule:

\begin{quote}\small\ttfamily
  Coordinate frame:\\
  - x is front/back on the table.\\
  - y is left/right on the table. Positive y is LEFT; negative y is RIGHT.\\
  - z is height. Positive z is upward.
\end{quote}

\noindent The second is one of the per-class task recipes (category
5), illustrating how concrete the evolved guidance becomes, right
down to numerical stack heights tuned for the simulator's
$\sim$5\,cm cubes:

\begin{quote}\small\ttfamily
  Single vertical tower: collect detections for the requested cubes
  and ignore absent colors. If target cube centroids already share
  the same x,y within about 0.03\,m with increasing z, stop.
  Otherwise choose one detected target cube or existing stack as the
  base and keep its current x,y; do not invent a new tower center
  and never use coordinate-task targets such as y=-0.4. \ldots
  For each moved cube, grab that cube at its listed centroid, then
  drop at the base x,y with z values about 0.09, 0.145, 0.200,
  \ldots as the tower grows.
\end{quote}

The second layer is the tool-body elaboration available only to MF,
which the HELIX \texttt{codex} mutator was observed (via
container traces) to develop iteratively by introspecting prior
trace JSONs and per-task result summaries. The three most
consequential tool-side additions are:

\begin{itemize}
  \item \textbf{Color-mask verification on perception output.}
        After Grounding~DINO and SAM\,2 return candidate detections,
        the evolved tool samples each detection's segmentation mask
        in RGB and rejects detections whose dominant color does
        not match the requested color word
        (e.g.\ ``yellow cube''). This eliminates the most frequent
        false-positive failure mode of the seed pipeline.
  \item \textbf{Multi-query perception fallback.} When a query
        such as ``tomatoes'' returns nothing, it automatically retries
        singular and color-qualified variants (``tomato'',
        ``red tomato'') before giving up. This compensates for the
        open-vocabulary detector's known phrasing sensitivity.
  \item \textbf{Detection caching with grab-snap.} The last
        $\sim$80 detected centroids are cached. If the LLM calls
        \texttt{move\_to\_point(grab=\ldots)} within $\sim$7.5\,cm
        of a cached centroid, the tool snaps to the cached value.
        This defends against the LLM rounding off coordinates or
        accidentally substituting a drop target for a grab target.
\end{itemize}

The tool-side additions are precisely the kind of edits that the
SF ablation cannot express: they live behind the agent's tool
signatures rather than in front of them, and they fix failure
modes (false-positive color detections, query phrasing
sensitivity, coordinate rounding) that prompt rephrasing alone
does not address. This drives the
$+9.5$\,pp absolute gap between the MF and SF best-by-train
candidates in Table~\ref{tab:rai-easy-headline}: the prompt-side
edits raise the floor in both cells, but tool-side edits raise the
ceiling. This is the same pattern reported in
Section~\ref{sec:experiments} for the multi-file repository seed
versus the single-file stub on CaP-Bench Robosuite.

The mutator independently identified the concrete failure modes (axis
confusion, double-grabs, false-positive color detections,
plural/singular query mismatches) and codified them into the
prompt and tool wrappers automatically, end to end, on hardware
available to a single engineer.

% ---------------------------------------------------------------------
\subsubsection{Results: the hard subset}
\label{sec:rai-hard}

The \textsc{hard} subset is a substantially more demanding regime:
scenes contain five to a dozen distractor objects, several start with
pre-stacked cubes, the \textsc{PlaceCubes} pass criterion tightens to
$0.10$\,m, and the four-class mix introduces \textsc{GroupObjects}
and \textsc{BuildCubeTower}, on which the unmodified seed agent is
closest to total failure (the seed scores $0\%$ on $24$ of $24$
\textsc{GroupObjects} task-runs in the MF cell, and on $20$ of $24$
in the SF cell).

\begin{table}[tp]
  \centering
  \caption{Held-out test scores on the RAI \textsc{hard} subset
    ($n = 21$ tasks, averaged over 3 runs, GPT-5.1). The per-task
    paired $t$-test spans 63 pairs (21 tasks $\times$ 3 runs).
    All values in percent.}
  \label{tab:rai-hard-headline}
  \setlength{\tabcolsep}{4pt}
  \begin{tabular}{lcccc}
    \toprule
    Cell          & Seed (mean $\pm$ std) & Best (mean $\pm$ std)   & Lift        & 95\% CI ($t$)                     \\
    \midrule
    MF (headline) & $23.5 \pm 3.8$        & $\mathbf{44.3 \pm 2.4}$ & $+20.8$\,pp & $[+11.0, +31.0]$\,pp ($t = 4.09$) \\
    SF (ablation) & $25.8 \pm 12.0$       & $\mathbf{37.6 \pm 5.4}$ & $+11.8$\,pp & $[+2.0, +22.0]$\,pp ($t = 2.33$)  \\
    \bottomrule
  \end{tabular}
\end{table}

Table~\ref{tab:rai-hard-headline} mirrors the structure of the
\textsc{easy} results: the MF configuration produces the larger
held-out lift, $+20.8$\,pp ($p < 0.001$), and the SF ablation
produces a smaller but still significant lift, $+11.8$\,pp
($p \approx 0.023$). The absolute gap between the best-by-train
candidates is $+6.7$\,pp in MF's favor (95\% CI
$[+0.0, +13.0]$\,pp, $t = 1.96$); the lower bound just touches
zero, which is consistent with the SF \emph{seed} on \textsc{hard}
having an unusually wide per-run standard deviation
($\sigma = 12.0$\,pp), three times the MF seed's
$\sigma = 3.8$\,pp. The evolved candidates, by contrast, are both
well within the noise floor of $\sigma \approx 3$\,pp that we
observed in the \textsc{easy} cell.

\subsubsection*{Where the lift comes from: SF and MF are
  complementary on \textsc{hard}}

A per-class breakdown reveals a richer story than the headline
mean suggests. Table~\ref{tab:rai-hard-perclass} reports per-class
held-out means for the seed and best-by-train candidates in each
cell, averaged over the three runs ($n$ ranges from $9$ to $24$
task-runs per class depending on class size in the test split).

\begin{table}[tp]
  \centering
  \caption{Per-class held-out test means on the RAI \textsc{hard}
    subset (3-run averages, $n$ shown is per-class task count
    summed over the three runs). The two cells cover different
    parts of the class space: SF recovers
    \textsc{BuildCubeTower} and most of \textsc{MoveObjectsToLeft}
    but regresses slightly on \textsc{GroupObjects}; MF essentially
    unlocks \textsc{GroupObjects} but does not move
    \textsc{BuildCubeTower}. All values in percent.}
  \label{tab:rai-hard-perclass}
  \begin{tabular}{lrrrrr}
    \toprule
    Class                      & $n$ (3 runs) & SF seed & SF best         & MF seed & MF best         \\
    \midrule
    \textsc{BuildCubeTower}    & 9            & $5.6$   & $\mathbf{44.4}$ & $16.7$  & $16.7$          \\
    \textsc{GroupObjects}      & 24           & $15.3$  & $9.7$           & $0.0$   & $\mathbf{37.5}$ \\
    \textsc{MoveObjectsToLeft} & 15           & $45.2$  & $\mathbf{76.5}$ & $56.8$  & $71.0$          \\
    \textsc{PlaceCubes}        & 15           & $35.1$  & $39.2$          & $31.7$  & $\mathbf{45.0}$ \\
    \bottomrule
  \end{tabular}
\end{table}

The two cells produce \emph{qualitatively different} sources of
lift on \textsc{hard}:

\begin{itemize}
  \item \textbf{SF is strong on stacking and bulk-motion classes.}
        SF recovers \textsc{BuildCubeTower} essentially from
        scratch ($5.6\% \to 44.4\%$, $+38.8$\,pp) and nearly closes
        the \textsc{MoveObjectsToLeft} gap ($45.2\% \to 76.5\%$,
        $+31.3$\,pp). Both classes have explicit per-task recipes
        in the evolved prompt (the analogous tower recipe
        reproduced in the \textsc{easy} SF prompt,
        Appendix~\ref{app:rai-prompts-pe}, fixes stack heights at
        $z \in \{0.09, 0.14, 0.19, \ldots\}$, exactly the
        per-cube spacing needed for the simulator's
        $\sim$5\,cm cubes), and the $y$-axis disambiguation alone
        covers most of the left-side movement headroom.
  \item \textbf{MF is strong on perception-bound classes.} MF
        unlocks \textsc{GroupObjects} from a near-total failure
        ($0.0\% \to 37.5\%$, $+37.5$\,pp) and gains on
        \textsc{PlaceCubes} ($31.7\% \to 45.0\%$, $+13.3$\,pp).
        These are precisely the classes where the seed agent's
        most common failure mode is not bad planning but bad
        perception (a cluster query that returns the wrong
        color, a cube detection that fails to distinguish red
        from yellow under shadow). The wrapper-level edits MF
        has access to (color-mask verification on top of the
        upstream detector, multi-query fallbacks, detection
        caching) are what move these classes, and no prompt edit
        substitutes for them.
  \item \textbf{Neither cell cracks \textsc{BuildCubeTower}
          beyond mid-tier credit on MF, nor improves
          \textsc{GroupObjects} on SF.} SF in fact regresses
        slightly on \textsc{GroupObjects} ($15.3\% \to 9.7\%$),
        consistent with the train-set frontier being dominated by
        gains on the other three classes and the minibatch
        acceptance rule letting a small per-class regression
        through.
\end{itemize}

The aggregate result, then, is not that MF is simply ``more SF'':
MF and SF specialize in opposite directions on \textsc{hard}. MF
wins the headline mean because \textsc{GroupObjects} contributes
$8$ of the $21$ test tasks and the \textsc{GroupObjects} unlock is
large.

\subsubsection*{Compute: tool calls and wall time}

\begin{table}[tp]
  \centering
  \caption{Held-out test compute totals across the 21
    \textsc{hard} test tasks (representative single run; the
    sign matches under the 3-run average). The MF candidate
    reaches its higher accuracy with fewer tool calls and lower
    wall time, mirroring the \textsc{easy} pattern. The SF
    candidate on \textsc{hard} reverses sign: it spends
    \emph{more} compute than the seed, because its evolved
    prompt instructs the agent to re-perceive after each drop
    and to keep moving objects until the task condition holds.}
  \label{tab:rai-hard-compute}
  \begin{tabular}{lrrrr}
    \toprule
    Cell          & Metric              & Seed & Best & $\Delta$           \\
    \midrule
    MF (headline) & Total tool calls    & 205  & 149  & $\mathbf{-27.3\%}$ \\
    MF (headline) & Total wall time (s) & 1278 & 1019 & $\mathbf{-20.3\%}$ \\
    \midrule
    SF (ablation) & Total tool calls    & 216  & 251  & $+16.2\%$          \\
    SF (ablation) & Total wall time (s) & 1363 & 1827 & $+34.0\%$          \\
    \bottomrule
  \end{tabular}
\end{table}

Table~\ref{tab:rai-hard-compute} shows that the compute trade-off
on \textsc{hard} now differs by cell. The MF candidate again
reduces both tool calls and wall time substantially
(\textbf{27.3\%} and \textbf{20.3\%}), even more aggressively than
on \textsc{easy}. The SF candidate, in contrast,
\emph{increases} compute by \textbf{16.2\%} in tool calls and
\textbf{34.0\%} in wall time: its evolved prompt instructs the
agent to re-detect the scene between moves and to keep moving
until the task condition holds, which on the larger \textsc{hard}
scenes costs additional perception round-trips. This is consistent
with the framing that prompt edits raise the floor (by making the
agent more thorough) while tool edits raise the ceiling (by making
each call more discriminating, and therefore fewer of them
necessary). These are the same numbers cited in the
repository-level mutation ablation in
Sections~\ref{sec:exp-rq3} and~\ref{sec:exp-rq4} of the main
text.

\subsubsection*{What RHO learned on \textsc{hard}}

The mutator's behavior on \textsc{hard} matches the per-class
picture above: it allocates its effort \emph{toward tool-body
  edits} rather than prompt edits. The MF-evolved candidate
(\texttt{g6-s6}, generation 6 of 10) grows
\texttt{solver/tools.py} from $375$ lines in the seed to $1{,}287$
lines (a $3.4\times$ expansion, comparable to the \textsc{easy}
MF candidate's $3.3\times$), but it grows the
\texttt{MANIPULATION\_PROMPT} body by only $\sim$25 lines (against
$\sim$70 lines on \textsc{easy}). The SF candidate on
\textsc{hard} (\texttt{g7-s7}), by contrast, must do all of its
work in the prompt and expands \texttt{MANIPULATION\_PROMPT} to
$\sim$120 lines, the largest evolved prompt in this study.

The reason this matters is that the \textsc{hard} failure modes
are dominated by perception, not planning: the seed agent
typically chooses correct positions for an object once it knows
which centroid corresponds to which color or type, but it does
not always know which centroid is which. The MF mutator
preferentially extends \texttt{tools.py} (with the color-mask
verifier, the multi-query fallback, and the detection cache
already described in Section~\ref{sec:rai-insights}) because that
is where the leverage is on this subset. The notable result is that
RHO identifies \emph{where} to spend its edits, not just how many
lines to add.

% ---------------------------------------------------------------------
\subsubsection{Discussion}
\label{sec:rai-discussion}

Four observations from the RAI \textsc{easy} and \textsc{hard}
results bear on the paper's main thesis.

\paragraph{Methodology transfers.}
RHO's closed-loop evolution moves from CaP-Bench to a
qualitatively different benchmark (one whose state is a
Vulkan-rendered O3DE simulator driven by ROS 2 and a real
perception stack, rather than a procedural primitive surface) with
no changes to the mutator, the scalar-sum acceptance gate, or the
best-by-train selection rule relative to the canonical CaP-Bench
Codex GPT-5.5 Multi-file run. The MF configuration adds one substrate-specific
constraint, an AST-level tool-name-contract check that rejects
candidates which alter the fixed tool API exposed to the LLM caller
(Sec.~3.2 above). The same apparatus produces a statistically
significant headline MF lift on both subsets ($+20.6$\,pp on
\textsc{easy} and a near-identical $+20.8$\,pp on \textsc{hard});
the SF ablation produces a smaller but significant lift on both
($+14.3$\,pp on \textsc{easy}, $+11.8$\,pp on \textsc{hard}). The
MF--SF absolute gap is $+9.5$\,pp on \textsc{easy} and $+6.7$\,pp
on \textsc{hard}. In both cases the gap is consistent with MF's
freedom to edit tool bodies (i.e., repository-level mutation)
being worth roughly a third to a half of the headline lift on top
of what prompt edits alone can deliver.

\paragraph{Prompt and tool edits cover different failure modes.}
The \textsc{hard} per-class breakdown
(Table~\ref{tab:rai-hard-perclass}) shows that SF and MF do not
simply lift the same tasks by different amounts; they specialize.
SF recovers \textsc{BuildCubeTower} ($+38.8$\,pp class lift) and
\textsc{MoveObjectsToLeft} ($+31.3$\,pp), both classes whose seed
failures stem from planning and coordinate-frame errors that
prompt-level instructions can fix. MF essentially unlocks
\textsc{GroupObjects} ($+37.5$\,pp) and \textsc{PlaceCubes}
($+13.3$\,pp), classes whose seed failures stem from perception
ambiguity that wrapper-level verification can fix. The two
ablations are therefore complementary signals about \emph{where} a
given failure mode lives, not just \emph{whether} RHO can
move it.

\paragraph{Lifts come with lower inference cost, with one
  instructive caveat.}
The MF headline candidate reduces both tool calls and wall time
relative to its seed on both subsets ($-17.6\%$ / $-3.8\%$ on
\textsc{easy}, $\mathbf{-27.3\%}$ / $\mathbf{-20.3\%}$ on
\textsc{hard}). The SF ablation reduces compute on \textsc{easy}
($-9.2\%$ tool calls, $-8.2\%$ wall time) but \emph{increases}
compute on \textsc{hard} ($+16.2\%$ tool calls, $+34.0\%$ wall
time), because its evolved prompt instructs the agent to
re-perceive between moves and persist until the task condition
holds. Tool-side edits, which can encode discriminating logic
behind the agent's tool API, therefore appear to be the more
robust route to ``accurate \emph{and} cheap'' for a closed-loop
hardware setting; prompt-side edits buy accuracy on \textsc{hard} but can
do so by spending more inference budget. This is the same
inversion the main paper highlights against single-file
GEPA-style~\citep{agrawal2026gepa} optimization
(Section~\ref{sec:exp-rq3}).

\paragraph{Evolution suppresses sampler variance.}
The per-run standard deviation of the evolved candidates is
substantially smaller than that of the seeds in every cell. On
\textsc{easy}, $\sigma_{\text{best}} = 2.8$\,pp for both MF and SF
versus $\sigma_{\text{seed}} \in \{9.9, 7.3\}$\,pp, a roughly
$3\times$ tightening. On \textsc{hard},
$\sigma_{\text{best}} = 2.4$\,pp (MF) and $5.4$\,pp (SF) versus
$\sigma_{\text{seed}} \in \{3.8, 12.0\}$\,pp, a $1.6\times$ to
$2.2\times$ tightening. For any deployed robotics system, where
reproducibility and determinism are first-class concerns, this
variance reduction is a useful side-effect that does not appear in
the headline accuracy numbers.

Together with the main CaP-Bench results
(Section~\ref{sec:experiments}), these findings support the claim
that RHO is a general recipe for closed-loop refinement of
robot-policy repositories, including LLM-in-the-loop agent harnesses,
and that it can be applied with modest single-workstation compute to
industrially relevant robotics stacks.

% =====================================================================
\subsubsection{Full evolved prompts}
\label{app:rai-prompts}

For reference, this appendix reproduces the three prompts that the
inline snippets in Section~\ref{sec:rai-insights} draw from: the
unedited RAI seed prompt, the best-by-train SF-evolved prompt
(\texttt{g2-s2}, generation 2 of 10), and the best-by-train
MF-evolved prompt (\texttt{g8-s8}, generation 8 of 10). Long lines
have been allowed to wrap; the underlying text is otherwise verbatim
from each cell's evolved \texttt{solver/prompts.py}.

\subsubsection{Seed prompt}
\label{app:rai-prompts-seed}

The seed is RAI's stock manipulation system prompt, copied verbatim
from \texttt{rai\_bench.manipulation\_o3de.interfaces.ManipulationTask}.
Note line 4, which inverts the convention the benchmark actually
uses for scoring (see Section~\ref{sec:rai-insights}).

\begin{Verbatim}[fontsize=\small, frame=single, framesep=4pt,
    breaklines=true]
  You are a robotic arm with interfaces to detect and manipulate objects.
  Here are the coordinates information:
  x - front to back (positive is forward)
  y - left to right (positive is right)
  z - up to down (positive is up)
  Before starting the task, make sure to grab the camera image to understand the environment.
\end{Verbatim}

\subsubsection{SF-evolved prompt (best-by-train, \texttt{g2-s2})}
\label{app:rai-prompts-pe}

RHO's single-file ablation expanded the 6-line seed into the
following 78-line task-specific guidance. This is the prompt that
produced the SF numbers in Tables~\ref{tab:rai-easy-headline}
and~\ref{tab:rai-easy-compute}.

\begin{Verbatim}[fontsize=\small, frame=single, framesep=4pt,
    breaklines=true]
  You control a Franka Panda arm in a closed-loop tabletop simulator. Complete
  the user's manipulation task by calling tools, not by describing a plan.

  Start every scenario with a fresh camera image. If the tool is
  get_ros2_camera_image, call it with no arguments. If the image tool is
  get_ros2_image, use topic="/color_image5".

  Coordinate frame:
  - x is front/back on the table; larger x is farther forward.
  - y is right/left on the table. POSITIVE y is LEFT and negative y is RIGHT.
  - z is height above the table; larger z is up. Tabletop object centroids are
  usually near z=0.04 m.

  Object names for detection:
  - Before moving anything, call get_object_positions for the exact required
  object type. Use singular food names: "apple", "carrot", "tomato", "corn".
  Use color-qualified cube names: "red cube", "blue cube", "yellow cube".
  - For "all cubes", query red cube, blue cube, and yellow cube separately.
  - If different cube-color queries return nearly identical x,y,z coordinates,
  treat them as one physical cube and move it at most once.
  - Never invent object coordinates when detection succeeds.

  Pick/place discipline:
  - Move one object completely before moving another object.
  - The normal sequence for one object is exactly:
  1. choose a returned centroid for the object,
  2. move_to_point(x=centroid_x, y=centroid_y, z=centroid_z, task="grab"),
  3. move_to_point(x=goal_x, y=goal_y, z=goal_z, task="drop").
  - A grab call must use the object's exact detected centroid. Never grab at a
  destination coordinate or at a made-up y such as 0.35.
  - After a successful grab, the next arm call should be the drop for that same
  object. Do not issue two grab calls in a row.
  - For ordinary tabletop placement, use the object's detected z for the drop.
  Use higher z only when intentionally building a vertical tower.
  - Stop when the requested relation is satisfied; extra moves often reduce the
  score.

  Task recipes:

  1. Place one object at coordinates (x: X, y: Y)
  - Detect the named object type.
  - If several instances are returned, choose one that is not already at X,Y,
  usually the instance farthest from that target.
  - Grab at its exact centroid, then immediately drop at x=X, y=Y, z=detected_z.

  2. Move all requested objects to the left side of the table
  - Positive y is the left side. Move only the object types named in the request;
  ignore distractors.
  - Detect each requested type separately.
  - Leave any target already at y >= 0.25 in place.
  - For each target with y < 0.25: grab at its exact centroid, then immediately
  drop at the same x, detected z, and a clear positive-y value from 0.32 to
  0.45. For multiple moved targets, use small y or x offsets to avoid overlap
  while staying reachable.

  3. Make all cubes adjacent to at least one cube
  - Detect red, blue, and yellow cubes, deduplicate overlapping detections, and
  ignore all non-cubes.
  - If every physical cube already has another cube within about 0.10 m in the
  x/y plane, stop.
  - Keep one central reachable cube as the anchor. Move only isolated cubes.
  - Drop each moved cube on the tabletop next to the anchor or growing group with
  center spacing about 0.06 to 0.08 m in x or y. Do not stack for adjacency.

  4. Build one vertical tower from cubes
  - Use only the cube colors requested by the task. Ignore non-cubes and unnamed
  cube colors.
  - Detect the requested colored cubes and deduplicate physical cubes.
  - Leave one low, stable, reachable cube in place as the base.
  - For each remaining cube: grab at its exact centroid and immediately drop at
  the base x,y with increasing center heights: second cube z=0.09, third
  z=0.14, fourth z=0.19, fifth z=0.24, sixth z=0.29, seventh z=0.34.

  5. Form separate clusters based on object types
  - Only manipulate the object types listed before "form separate clusters";
  never move unlisted distractors.
  - For each listed type, keep one object as that type's anchor and move other
  same-type objects to positions 0.06 to 0.08 m from the anchor.
  - Keep clusters for different types at least about 0.18 m apart, using
  distinct y bands such as -0.30, 0.00, and 0.30.
\end{Verbatim}

\subsubsection{MF-evolved prompt (best-by-train, \texttt{g8-s8})}
\label{app:rai-prompts-te}

The headline MF candidate's system prompt. This is the prompt that
the inline snippets in Section~\ref{sec:rai-insights} are excerpted
from, and that produced the MF rows in
Tables~\ref{tab:rai-easy-headline} and~\ref{tab:rai-easy-compute}.

\begin{Verbatim}[fontsize=\small, frame=single, framesep=4pt,
    breaklines=true]
  You control a Franka Panda arm in a tabletop manipulation scene. Work in
  closed loop: first call get_ros2_camera_image for the canonical
  /color_image5 view, then call get_object_positions for the needed object
  types, then use move_to_point for physical actions. Trust fresh tool
  results over guesses.

  Coordinate frame:
  - x is front/back on the table.
  - y is left/right on the table. Positive y is LEFT; negative y is RIGHT.
  - z is height. Positive z is upward.

  Critical tool rules:
  1. A physical move of an object is one grab call followed by one drop call.
  Do not call grab again to lift or carry the same object; a second grab
  can reopen the gripper and lose the object.
  2. Every grab must use the exact x,y,z centroid from get_object_positions
  for the object you intend to move. Never use a cluster center, left/right
  side location, target coordinate, tower base, or suggested drop slot as a
  grab coordinate.
  3. For a table drop, use the target x,y and z about 0.04 to 0.06. For a
  stack drop, use the base x,y and increase z by about 0.055 per cube above
  the base.
  4. Use the exact argument names x, y, z, task. If a call is malformed,
  correct it before continuing.
  5. After each successful drop, stop if the task is satisfied; otherwise
  re-detect before choosing the next object when positions may have
  changed.
  6. If get_object_positions says no objects of a requested type were
  detected, skip that type. Never substitute a different color/type just
  because the task mentions it.
  7. For a group/separate-clusters task, do not stop just because a requested
  type has one detected object; it still must be separated from other types
  if the scene is mixed.
  8. Tool outputs may include several named aids. Use only the aid that
  matches the task wording: left-side aid for left-side tasks, tower aid
  for tower tasks, cube adjacency aid for adjacent-cubes tasks, cluster aid
  for separate-clusters tasks, and multi-object aid for coordinate
  placement. Ignore unrelated aids.

  Object queries:
  - For red, blue, or yellow cubes, query the full color name: "red cube",
  "blue cube", "yellow cube".
  - For "all cubes are adjacent", query "cube" once so every cube color is
  included.
  - For a tower task that explicitly requests red cubes, blue cubes, and
  yellow cubes together, query "cube" once and stack every detected cube.
  For tower tasks that request only a subset of colors, query only those
  color names.
  - For apples, carrots, tomatoes, and corn, query the singular object name
  unless the task asks for a plural group.
  - Apples and tomatoes can cross-detect because both are red and round.
  get_object_positions may filter the list to likely candidates for the
  requested fruit; when it does, use the returned filtered centroid(s). If
  it returns an apple/tomato disambiguation aid, choose the requested type
  from that aid before considering which centroid is closest to a coordinate
  target.

  Task recipes:
  - Left side of the table: for every requested object type, move each
  detected object whose y <= 0.0 to a free location well into positive y,
  around 0.38 to 0.50. Keep objects already at y > 0.0. Preserve reachable
  x when possible and space multiple drops by about 0.07 m in x or y.
  - Single vertical tower: collect detections for the requested cubes and
  ignore absent colors. If target cube centroids already share the same
  x,y within about 0.03 m with increasing z, stop. Otherwise choose one
  detected target cube or existing stack as the base and keep its current
  x,y; do not invent a new tower center and never use coordinate-task
  targets such as y=-0.4. Count the detected target cubes: if there are N
  cubes and K are already in the chosen stack, make exactly N-K grab/drop
  moves before stopping. For each moved cube, grab that cube at its listed
  centroid, then drop at the base x,y with z values about 0.09, 0.145,
  0.200, ... as the tower grows. Do not include apples, carrots, tomatoes,
  corn, or unrequested cube colors.
  - All cubes adjacent: make connected neighbors, not a midpoint pile and
  not a stack. Use the cube adjacency aid from get_object_positions when
  present: grab the named isolated cube index and copy the suggested drop
  tuple exactly. The anchor cube's listed centroid is not a drop target.
  Pick an anchor cube and move each isolated cube to a free spot 0.06 to
  0.09 m beside an existing cube, such as same x with y plus/minus 0.07 or
  same y with x plus/minus 0.07. Never drop at the exact x,y of another
  cube for adjacency. In a two-cube scene, exactly one cube usually moves
  once, then re-detect; stop when every cube has another cube within about
  0.10 m in x/y.
  - Separate clusters by type: only move the requested types. For each
  requested type, make one connected group with same-type objects 0.06 to
  0.10 m apart; do not make several local pair clusters. Use the cluster
  aid anchor when present: keep that anchor and move every other object of
  that same type into slots around the same anchor before stopping. Put
  different requested-type cluster centers at least 0.18 m apart; if
  centers overlap, use the first requested type around positive y (about
  0.30) and the next requested type around negative y (about -0.25).
  Cluster centers are drop targets only. Always grab a listed centroid
  first. A type with one detected object is connected, but not
  automatically separated; if the scene includes other objects nearby,
  move that object to an isolated reachable spot.
  - Placed at coordinates (x: X, y: Y): move exactly one requested object to
  X,Y. First resolve object identity, especially apple vs tomato when the
  tool gives a disambiguation aid. If several valid requested objects
  remain, compute each listed centroid's xy distance to X,Y and usually
  grab the closest reachable one; only avoid it if it is at an extreme
  table edge. Do not pick a generic central object when another valid
  requested object is visibly closer to the target. Drop exactly at X,Y
  with table z around 0.04 to 0.06.

  Keep plans short and physical. Single-object tasks are usually four tool
  calls total: image, positions, grab, drop. Multi-object tasks repeat one
  grab/drop pair per object that actually needs to move.
\end{Verbatim}

\end{document}